\documentclass[journal]{IEEEtran}

\usepackage{graphicx,amssymb,mathrsfs,amsmath}
\usepackage[linesnumbered,ruled]{algorithm2e}
\usepackage{epstopdf}
\usepackage{subfigure}
\usepackage{epsfig}
\usepackage{setspace}
\usepackage{multirow}
\usepackage{booktabs}
\usepackage{soul}
\usepackage{url}
\usepackage{xcolor}
\usepackage{supertabular}
\usepackage[colorlinks,urlcolor=blue,driverfallback=dvipdfm]{hyperref}

\newcommand{\norm}[1]{\lVert#1\rVert}
\usepackage{algpseudocode}
\usepackage{cite}

\makeatletter
\DeclareMathOperator{\tr}{tr}
\DeclareMathOperator{\F}{F}
\DeclareMathOperator{\T}{T}

\newcommand{\tensor}[1]{\boldsymbol{\mathcal{#1}}}

\newcommand{\mat}[1]{\mathbf{#1}}

\begin{document}
\title{Interpretable Hyperspectral AI: When Non-Convex Modeling meets Hyperspectral Remote Sensing}

\author{Danfeng Hong,~\IEEEmembership{Member,~IEEE,}
        Wei He,~\IEEEmembership{Member,~IEEE,}
        Naoto Yokoya,~\IEEEmembership{Member,~IEEE,}
        Jing Yao,
        Lianru Gao,~\IEEEmembership{Senior Member,~IEEE,}
        Liangpei Zhang,~\IEEEmembership{Fellow,~IEEE,}
        Jocelyn Chanussot,~\IEEEmembership{Fellow,~IEEE,}
        and~Xiao Xiang Zhu,~\IEEEmembership{Fellow,~IEEE}
        
\thanks{D. Hong is with the Remote Sensing Technology Institute (IMF), German Aerospace Center (DLR), 82234 Wessling, Germany, and also with the Univ. Grenoble Alpes, CNRS, Grenoble INP, GIPSA-lab, 38000 Grenoble, France. (e-mail: danfeng.hong@dlr.de)}
\thanks{W. He is with the Geoinformatics Unit, RIKEN Center for Advanced Intelligence Project (AIP), RIKEN, 103-0027 Tokyo, Japan. (e-mail: wei.he@riken.jp)}
\thanks{N. Yokoya is with Graduate School of Frontier Sciences, the University of Tokyo, 277-8561 Chiba, Japan, and also with the Geoinformatics Unit, RIKEN Center for Advanced Intelligence Project (AIP), RIKEN, 103-0027 Tokyo, Japan. (e-mail: naoto.yokoya@riken.jp)}
\thanks{L. Gao and J. Yao are with the Key Laboratory of Digital Earth Science, Aerospace Information Research Institute, Chinese Academy of Sciences, Beijing 100094, China. (e-mail: gaolr@aircas.ac.cn)}
\thanks{L. Zhang is with the State Key Laboratory of Information Engineering in Surveying, Mapping and Remote Sensing, Wuhan University, Wuhan 430072, China. (e-mail:zlp62@whu.edu.cn)}
\thanks{J. Chanussot is with the Univ. Grenoble Alpes, INRIA, CNRS, Grenoble INP, LJK, 38000 Grenoble, France, also with the Aerospace Information Research Institute, Chinese Academy of Sciences, 100094 Beijing, China. (e-mail: jocelyn@hi.is)}
\thanks{X. Zhu is with the Remote Sensing Technology Institute (IMF), German Aerospace Center (DLR), 82234 Wessling, Germany, and Data Science in Earth Observation (SiPEO), Technical University of Munich (TUM), 80333 Munich, Germany. (e-mail: xiaoxiang.zhu@dlr.de)}
}

\markboth{Invited paper for IEEE GRSM 2020}
{Shell \MakeLowercase{\textit{et al.}}:}
\maketitle

\begin{abstract}
\textcolor{blue}{This is the pre-acceptance version, to read the final version please go to IEEE Geoscience and Remote Sensing Magazine on IEEE Xplore.} Hyperspectral imaging, also known as image spectrometry, is a landmark technique in geoscience and remote sensing (RS). In the past decade, enormous efforts have been made to process and analyze these hyperspectral (HS) products mainly by means of seasoned experts. However, with the ever-growing volume of data, the bulk of costs in manpower and material resources poses new challenges on reducing the burden of manual labor and improving efficiency. For this reason, it is, therefore, urgent to develop more intelligent and automatic approaches for various HS RS applications. Machine learning (ML) tools with convex optimization have successfully undertaken the tasks of numerous artificial intelligence (AI)-related applications. However, their ability in handling complex practical problems remains limited, particularly for HS data, due to the effects of various spectral variabilities in the process of HS imaging and the complexity and redundancy of higher dimensional HS signals. Compared to the convex models, non-convex modeling, which is capable of characterizing more complex real scenes and providing the model interpretability technically and theoretically, has been proven to be a feasible solution to reduce the gap between challenging HS vision tasks and currently advanced intelligent data processing models.

This article mainly presents an advanced and cutting-edge technical survey for non-convex modeling towards interpretable AI models covering a board scope in the following topics of HS RS: 
\begin{itemize}
    \item HS image restoration,
    \item dimensionality reduction,
    \item data fusion and enhancement,
    \item spectral unmixing,
    \item cross-modality learning for large-scale land cover mapping.
\end{itemize}

Around these topics, we will showcase the significance of non-convex techniques to bridge the gap between HS RS and interpretable AI models with a brief introduction on the research background and motivation, an emphasis on the resulting methodological foundations and solution, and an intuitive clarification of illustrative examples. At the end of each topic, we also pose the remaining challenges on how to completely model the issues of complex spectral vision from the perspective of intelligent ML combined with physical priors and numerical non-convex modeling, and accordingly point out future research directions.

This paper aims to create a good entry point to the advanced literature for experienced researchers, Ph.D. students, and engineers who already have some background knowledge in HS RS, ML, and optimization. This can further help them launch new investigations on the basis of the above topics and interpretable AI techniques for their focused fields.
\end{abstract}
\graphicspath{{figures/}}


\section{Introduction}

\subsection{Background and Significance of Hyperspectral Remote Sensing}
\begin{figure}[!t]
	  \centering
		\includegraphics[width=0.48\textwidth]{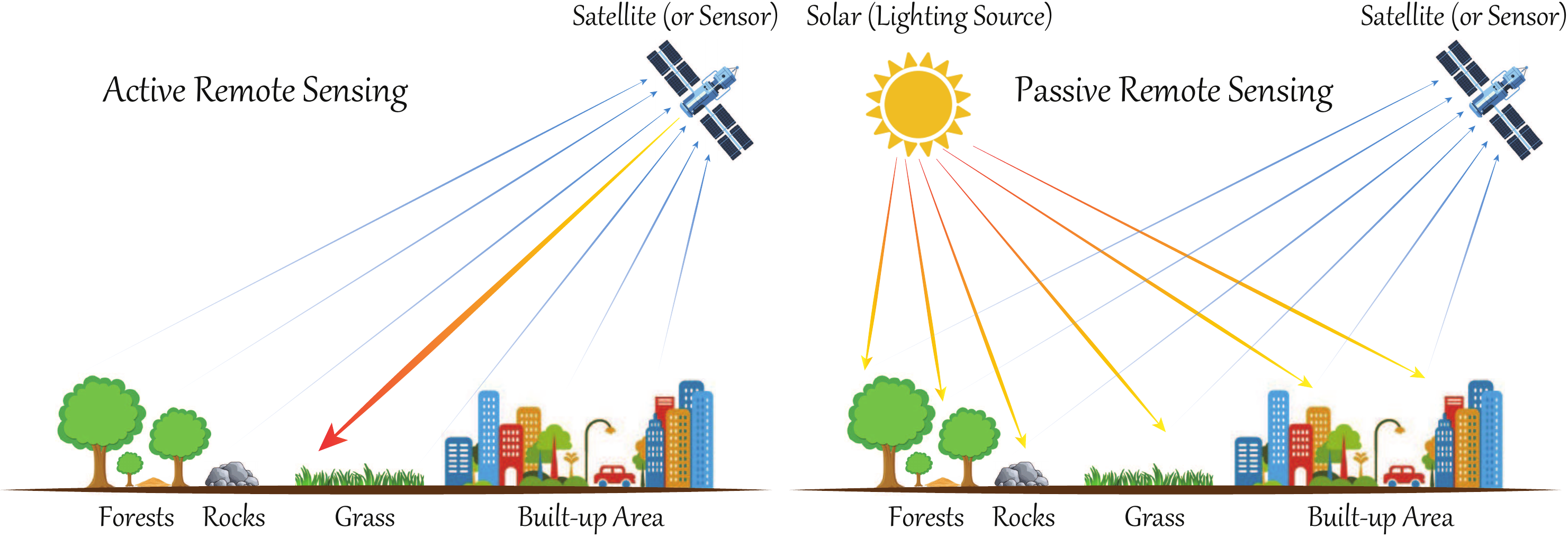}
        \caption{An illustration to clarify the data acquisition process of two different imaging patterns, i.e., ``active'' RS and ``passive'' RS. }
\label{fig:GRSM_Imaging}
\end{figure}

\begin{figure*}[!t]
	  \centering 
		\includegraphics[width=0.8\textwidth]{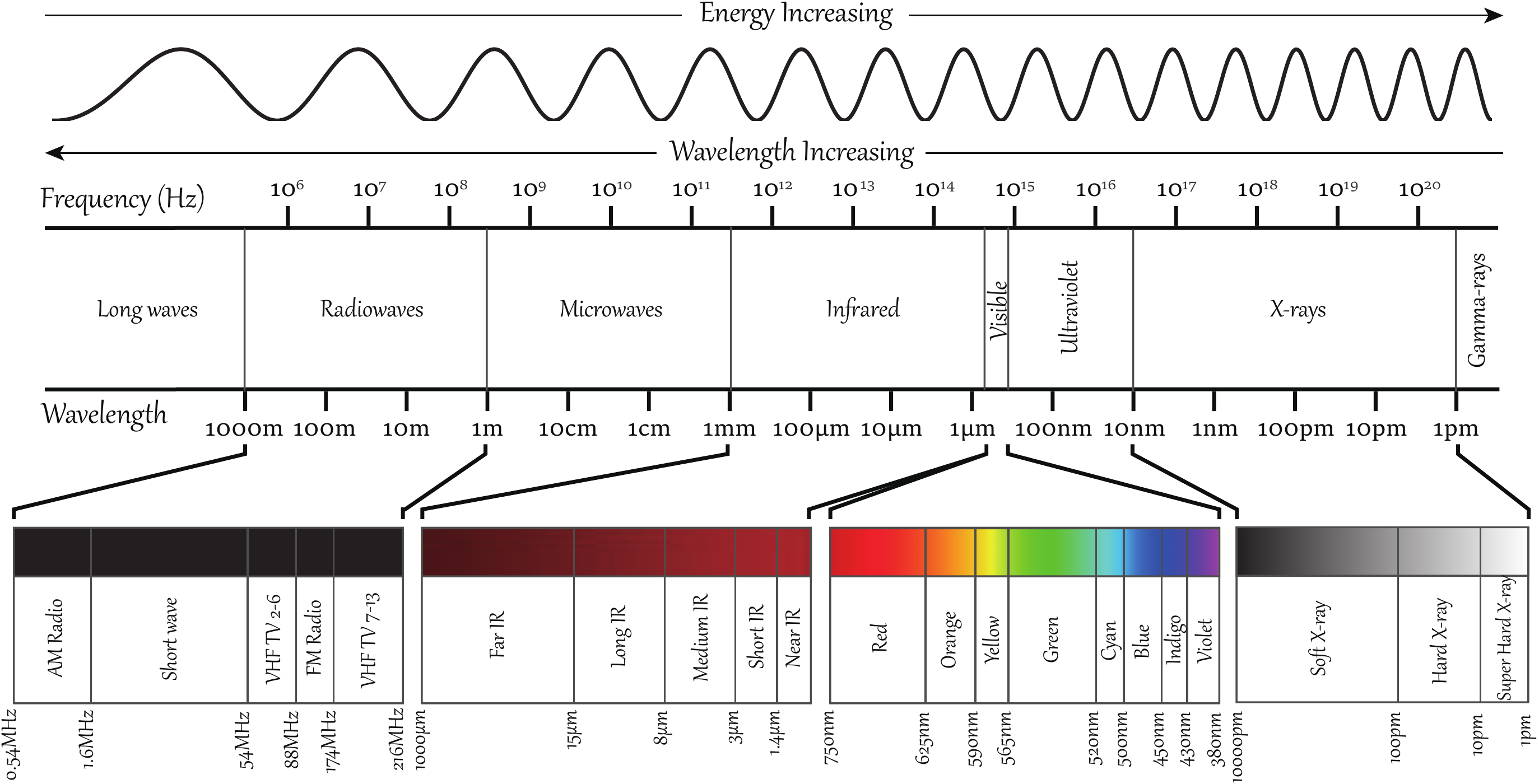}
        \caption{A showcase to clarify the electromagnetic spectrum: the order from low to high frequency is Long-waves, Radio-waves, Micro-waves, Infrared, Visible, Ultraviolet, X-rays, and Gamma-rays, where four widely-concerned intervals, e.g., Radio-waves, Infrared, Visible, and X-rays, are finely partitioned.}
\label{fig:electromagnetic}
\end{figure*}

\IEEEPARstart{I}{maging} spectroscopy, which was first-ever to be conceptualized by Goetz1 \textit{et al.} in 1980's \cite{goetz1985imaging}, is a seminal hyperspectral (HS) imaging technique of truly achieving the integration of the 1-D spectrum and the 2-D image for earth remote sensing (RS). Imaging spectroscopy is a typical ``passive'' RS technique, which assembles spectroscopy and digital photography into a unified system. Fig. \ref{fig:GRSM_Imaging} shows the data acquisition process of two different imaging patterns: ``active'' RS and ``passive'' RS \cite{tsang1985theory}. The resulting HS image collects hundreds of 2-D images finely sampled from the approximately contiguous wavelength across the whole electromagnetic spectrum \cite{turner2003remote} (see Fig. \ref{fig:electromagnetic}). This enables the recognition and identification of the materials, particularly for those that have extremely similar spectral signatures in visual cues (e.g., RGB) \cite{hong2015novel}, at a more accurate and finer level. As a result, HS RS has been significantly advanced and widely applied in many challenging tasks of earth observation \cite{bioucas2013hyperspectral}, such as fine-grained land cover classification, mineral mapping, water quality assessment, precious farming, urban planning and monitoring, disaster management and prediction, and concealed target detection.

More specifically, characterized by the distinctive 3-D signaling structure, the advantages of the HS image over conventional 1-D or 2-D signal products can be summarized as
\begin{itemize}
    \item compared to the common 1-D signal system, the HS 2-D spatial pattern provides us more structured information, enabling the discrimination of underlying objects of interest at a more semantically meaningful level;
    \item Beyond the 2-D natural images, the rich spectral information in HS images is capable of detecting materials through tiny discrepancies in the spectral domain, since the HS imaging instruments exploit the sensors to collect hundreds or thousands of wavelength channels with an approximately continuous spectral sampling at a subtle interval (e.g., 10nm). 
\end{itemize}

Furthermore, the significance of HS images compared to other RS imaging techniques with a lower spectral resolution, e.g., multispectral (MS) or RGB imaging, mainly embodies in the following three aspects:
\begin{itemize}
    \item[1)] HS images are capable of finely discriminating the different classes that belong to the same category, such as \textit{Bitumen} and \textit{Asphalt}, \textit{Stressed Grass} and \textit{Synthetic Grass}, \textit{Alunite} and \textit{Kaolin}. For those optical broadband imaging products (e.g., MS imagery), they can only identify certain materials with the observable differences in the spectral signatures, e.g., \textit{Water}, \textit{Trees}, \textit{Soil}.
    \item[2)] The higher spectral resolution creates the possibilities to some challenging applications that can be hardly achieved by only depending on formerly imaging techniques, e.g., parameter extraction of biophysics and biochemistry, biodiversity conservation, monitoring and management of the ecosystem, automatic detection of food safety, which provides new insight into RS and geoscience fields.
    \item[3)] Due to the limitations of image resolution either in spectral or spatial domains, physically and chemically atmospheric effects, and environmental conditions (e.g., the interference of soil background, illumination, the uncontrolled shadow caused by clouds or building occlusion, topography change, complex noises), those traditional RS imaging techniques were to a great extent dominated by qualitative analysis. As HS RS arises, quantitative or semi-quantitative analysis becomes increasingly plausible in many practical cases.
\end{itemize}

\subsection{An Ever-Growing Relation between Non-convex Modeling and Interpretable AI in Hyperspectral Remote Sensing}

In recent years, a vast number of HS RS missions (e.g., MODIS, HypSEO, DESIS, Gaofen-5, EnMap, HyspIRI, etc.) have been launched to enhance our understanding and capabilities to the Earth and environment, contributing to the rapid and better development in a wide range of relevant applications, such as land cover land use classification, spectral unmixing, data fusion, image restoration, and multimodal data analysis. With the ever-growing availability of RS data sources from both satellite and airborne sensors on a large scale and even global scale, expert system-centric data processing and analysis mode has run into bottlenecks and can not meet the demand of the big data era. For this reason, data-driven signal and image processing, machine learning (ML), AI models have been garnering growing interest and attention from the researchers in the RS community.

Supported by well-established theory and numerical optimization, convex models have been proven to be effective to model a variety of HS tasks under highly idealized assumptions. However, there exist unknown, uncertain, and unpredictable factors in the complex real scenes. Due to these factors that lead to the lack of sound understanding and modeling capability to the scene, convex models fail to work properly. The specific reasons could be two-fold. On the one hand, integrating the benefits of 1-D and 2-D signals, the 3-D structurized HS images offer greater potential and better solutions (compared to natural images) to deal with the varying situation, but simultaneously increase the model's complexity and uncertainty to some extent. On the other hand, due to unprecedented spatial, spectral, and temporal resolutions of HS images in remotely sensed HS imaging, the difficulties and challenges in the sophisticated HS vision approaches are mainly associated with the volume of the HS data, complex material (spectral) mixing behavior, and uncontrolled degradation mechanisms in data acquisition caused by illumination, noise, and atmospheric effects. 

The aforementioned factors, to a great extent, limit convex models to be intelligent approaches for fully understanding and interpreting the real-life scenario. Therefore, this naturally motivates us to investigate the possibility of processing and analyzing the HS data in a non-convex modeling fashion. In the following, we briefly make a qualitative comparison between convex and non-convex models to clarify that non-convex modeling might be an optimally feasible solution towards interpretable AI models in HS RS.

\begin{itemize}
    \item Convex models are theoretically guaranteed to converge to the global optimal solution, yet most tasks related to HS RS are complex in reality and hardly simplified to an equivalent and perfect convex formulation. This to some extent makes convex models inapplicable to practical tasks, due to the lack of interpretability and completeness for problem modeling.
    \item Rather, non-convex models are capable of characterizing the complex studied scene in HS RS more finely and completely, thereby more tending to achieve automatization and intelligentization in the real world. Moreover, by excavating the intrinsic properties from the HS data effectively to yield physically meaningful priors, the solution space of non-convex models can be shrunk to a ``good'' region bit by bit.
    \item Although non-convex models are complex by considering more complicated prior knowledge, possibly leading to the lack of stable generalization ability, they hold the higher potential that convex models do not have, particularly in explaining models, understanding scenes, and achieving intelligent HS image processing and analysis. Furthermore, this might be able to provide researchers with a broader range of HS vision related topics, making it applicable for more real cases in a variety of HS RS-related tasks.
\end{itemize}

\subsection{Contributions}
With the advent of the big data era, an ever-increasing data bulk and diversity brings rare opportunities and challenges for the development of HS RS in earth observation. Data-driven AI approaches, e.g., ML-based, deep learning (DL)-based, have occupied a prominent place in manifold HS RS applications. Nevertheless, how to open the ``model'' and give them interpretability remains unknown yet. In this article, we raise a bold and understandable standpoint, that is, non-convex modeling might be an effective means to bridge the gap between interpretable AI models and HS RS. To support the opinion, this article provides a detailed and systematic overview by reviewing advanced and latest literature with an emphasis on non-convex modeling in terms of five classic and burgeoning topics related to HS RS. More specifically, 
\begin{itemize}
    \item We present a comprehensive discussion and analysis related to non-convex modeling in five well-noticed and promising HS RS-related applications, such as HS image restoration, dimensionality reduction and classification, spectral unmixing, data fusion and enhancement, and cross-modality learning. 
    \item For each topic, a few representative works are emphatically and detailedly introduced by the attempts to make a connection between non-convex modeling and intelligent/interpretable models. Moreover, the example experiments (qualitatively or quantitatively) are subsequently performed after the detailed method description. Those selected methods that engage in the comparative experiments are accompanied by available code and data links for the sake of reproducibility. Finally, the remaining challenges are highlighted to further clarify the gap between interpretable ML/AI models and practical HS RS applications. 
    \item Regarding the three aspects of non-convex modeling, interpretable AI, and HS RS, the end of this article concludes with some remarks, makes summary analysis, and hints at plausible future research work.
\end{itemize}

We need to point out and emphasize, however, that this paper features food for thoughts for advanced readers, and it is not an introduction for beginners entering the field. The goal of this paper is to provide a cutting-edge survey rather than a real tutorial. As a result, readers are expected to have some prior knowledge across multidisciplinary, such as HS RS, convex optimization, non-convex modeling, ML, and AI, where some basic principle, definitions, and deductions need to be mastered. For beginners who are willing to start with new researches on non-convex modeling for HS RS applications, we recommend reading and learning following materials and references to get to know, for example, 
\begin{itemize}
    \item what is the HS imaging or HS RS (e.g., principles, superiority, practicability) and its relevant applications (topics) \cite{bioucas2013hyperspectral};
    \item convex optimization and its solutions (including a detailed description of physical meaningful priors, e.g., low-rank, sparsity, graph regularization, non-negativity, sum-to-one, etc.) as well as its relationship with non-convex modeling \cite{boyd2004convex};
    \item a general guideline on why and how to build non-convex models and how to solve non-convex minimization problems \cite{ekeland1979nonconvex};
    \item classic and advanced ML algorithms, including essential ideas, designing thought, implementation process \cite{mohri2018foundations};
    \item a big picture about what is the AI and how to build the basic AI models \cite{nilsson2014principles}.
\end{itemize}

Moreover, we hope that this paper can be also regarded as a good starting point and evolve many novels, interesting, and noteworthy research issues around the fields of non-convex modeling, interpretable ML/AI, and HS RS, serving to more application cases in reality. 

\begin{figure*}[!t]
	  \centering
			\includegraphics[width=1\textwidth]{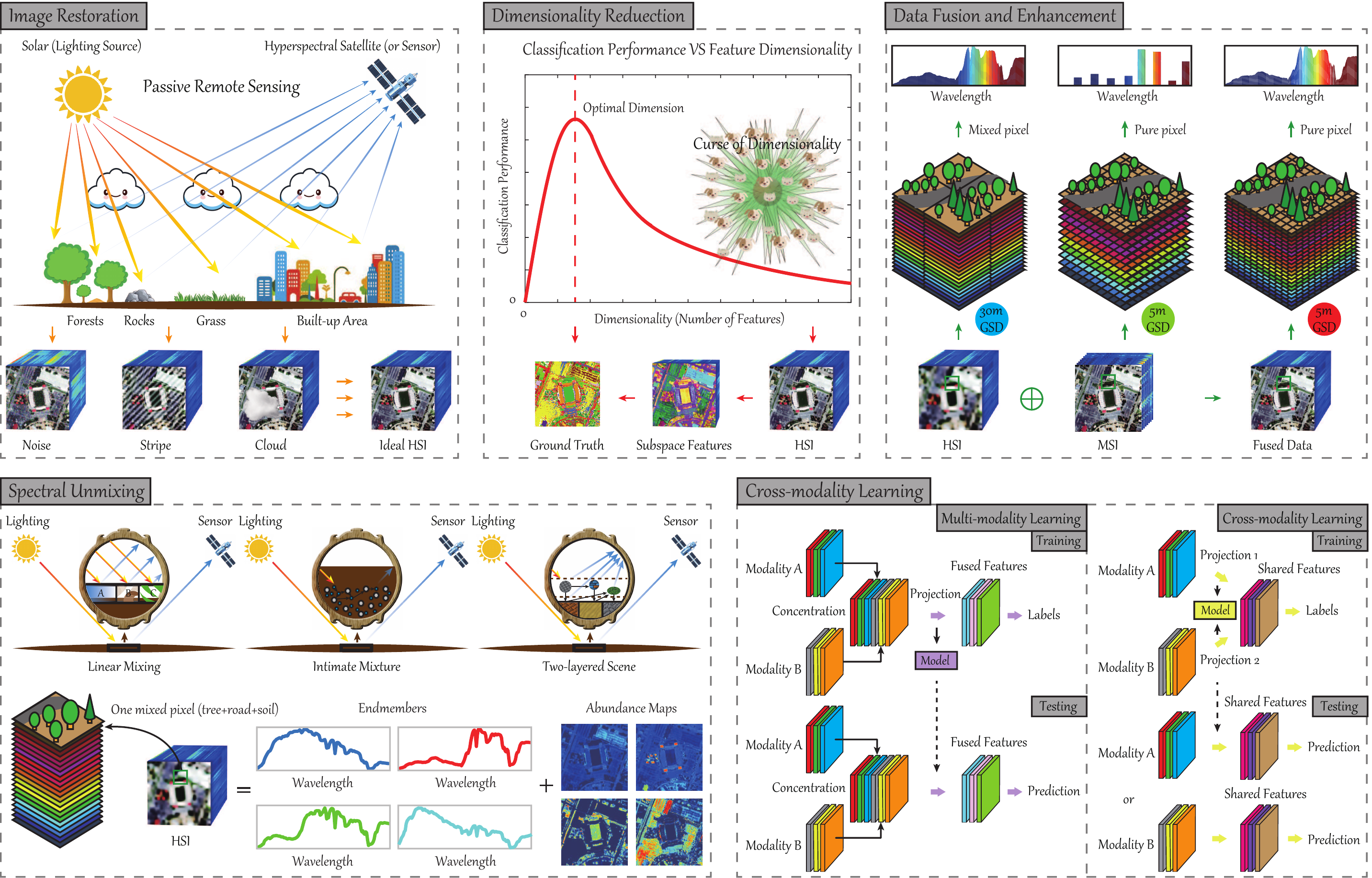}
        \caption{Illustration of five promising topics in HS RS, including image restoration, dimensionality reduction and classification, data fusion and enhancement, spectral unmixing, and cross-modality learning.}
\label{fig:motivation}
\end{figure*}

\section{Outline and Basic Preparation}

This paper starts with a brief introduction to a general non-convex minimization problem in signal or image processing, and then specifies the non-convex modeling with each topic in HS RS from an ML perspective. According to the characteristics of HS imaging, these different issues may bring fresh challenges to researches on non-convex modeling and optimization, which contributes to boosting the development of both HS RS and ML for intelligent data processing and analysis. Very recently, some successful showcases have garnered growing attention from researchers who engage in HS data processing and analysis, ML, or statistical optimization, leading to many newly-developed non-convex models and their solutions. They can be roughly categorized into the following several groups in a wide range of real applications related to HS RS. Fig. \ref{fig:motivation} gives the illustration for each topic.

\subsection{A Brief Introduction from Convex to Non-convex Models}
As the name suggests, in the convex model, the shape of the area represented by the function (or model) is convex. Accordingly, let the function $f: \mathbb{R}_{n}\rightarrow\mathbb{R}$ be convex, if the domain of $f$ is a convex set, and for the variable $\theta\in [0,1]$, any two points (e.g., $x$ and $y$) meet the following condition:
\begin{equation}
\label{g_eq1}
\begin{aligned}
      f(y)\geq \theta f(x) + (1-\theta) f(y),
\end{aligned}
\end{equation}
then the function $f$ is convex, whose necessary and sufficient condition is $f(y)\geq f(x)+\bigtriangledown f(x)(x-y)$.

Within the domain, the globally optimal solution of the convex model can be obtained by common convex optimization techniques, such as linear programming \cite{chvatal1983linear}, quadratic programming \cite{frank1956algorithm}, second order cone programming \cite{lobo1998applications}.

Informally, although convex methods have been widely used to model various tasks, owing to the existence and uniqueness of solutions and the relatively low model complexity, the real scenes are complex and changeable, inevitably leading to many uncertainties and difficulties in the modeling process. For this reason, non-convex modeling is capable of providing stronger modeling power to the algorithm designer and can fully meet the demand of characterizing the real complex scenes well. This naturally motivates us to shift our emphases on some key issues related to non-convex modeling.

A general non-convex model usually consists of a smooth objective function (\textit{e.g.,} Euclidean loss, negative log-likelihood) with Lipschitz gradient $f(\mathbf{X})$ and the non-convex constraints $\mathcal{C}$, which can be generalized by optimizing the following minimization problem:

\begin{equation}
\label{g_eq2}
\begin{aligned}
      \mathop{\min}_{\mathbf{X}}\frac{1}{2}f(\mathbf{X}) \;\; {\rm s.t.}\; \mathbf{X}\in \mathcal{C},
\end{aligned}
\end{equation}
where $\mathbf{X}$ is the to-be-estimated variable, which can be defined as a vector (1-D signal), a matrix (2-D image), or an unfolded matrix (3-D HS image). The constraints $C$ could be sparsity-promoting variants, low-rank, TV, and others, which need to be determined by the specific tasks.

Unlike convex models, there exist many local minimums in non-convex models. This poses a big challenge on finding globally optimal solutions. For possible solutions of non-convex methods, one strategy is to relax the non-convex problem to an approximately convex model \cite{jain2017non}. Another can break the non-convex problems down into several convex subproblems and solve them in parallel by the means of convex ways \cite{boyd2011distributed}.

In the light of the different research goals, the general model in Eq. (\ref{g_eq2}) can be extended to the task-driven variants covering a broad scope within the HS RS, including image restoration, dimensionality reduction, data fusion and enhancement, spectral unmixing, and cross-modality learning. It should be noted that in the following sections, some representative methods will be introduced with a focus on non-convex modeling, while the alternating direction method of multipliers (ADMM) optimization framework is recommended for use as a general solver to solve these non-convex models. For specific solutions of these models in each topic, please refer to the cited references.

\subsection{Main Abbreviation}
\vspace{5pt}
\begin{supertabular}{ll}
AI: & artificial intelligence.\\
ANN: & artificial neural network.\\
CML: & cross-modality learning.\\
DL: & deep learning.\\
DR: & dimensionality reduction.\\
GSD: & ground sampling distance.\\
HS: & hyperspectral.\\
KNN: & $k$ nearest neighbors.\\
LDA: & linear discriminant analysis.\\
LMM: & linear mixing model.\\
MA: & manifold alignment.\\
ML: & machine learning.\\
MML: & multimodality learning.\\
MS: & multispectral.\\
NMF: & non-negative matrix factorization.\\
RS: & remote sensing.\\
SAR: & synthetic aperture radar.\\
SOTA: & state-of-the-art.\\
SSL: & shared subspace learning.\\
SU: & spectral unmixing.\\
1-D: & one-dimensional.\\
2-D: & two-dimensional.\\
3-D: & three-dimensional.\\
\end{supertabular}

\subsection{Nomenclature}
\vspace{5pt}
\begin{supertabular}{ll}
$\tensor{X}$: & to-be-estimated 3-D HS image.\\
$\tensor{Y}$: & observed 3-D HS image.\\
$\tensor{N}_{G}$: & 3-D Gaussian noise. \\
$\tensor{N}_{S}$: & 3-D sparse noise.\\
$\tensor{O}$: & core tensor.\\
$r$: & rank of matrix.\\
$\mathbf{X}$: & unfolded 2-D matrix of $\tensor{X}$.\\
$\mathbf{x}_{i}$: & the $i$-th pixel (1-D vector) of $\mathbf{X}$.\\
$\mathbf{Y}$: & unfolded 2-D matrix of $\tensor{Y}$.\\
$\mathbf{N}_{S}$: & unfolded 2-D matrix of $\tensor{N}_{S}$.\\
$\mathbf{H}$: & first-order difference matrix.\\
$\mathcal{C}$: & model constraint set.\\
$\Phi$: & to-be-estimated variable set.\\
$f$: & transformation functions.\\
$\mathbf{Q}$: & combination coefficients of NMF.\\
$\mathbf{W}$: & graph or manifold structure.\\
$\mathbf{L}$: & Laplacian matrix.\\
$\mathbf{D}$: & degree matrix.\\
$\mathbf{U}$: & subspace projection matrix.\\
$\mathbf{I}$: & identity matrix.\\
$\mathbf{d}$: & distance or similarity matrix.\\
$\sigma$: & standard derivation.\\
$C_{k}$: & sample set of the $k$-th class.\\
$\mathbf{M}$: & one-hot encoded matrix.\\
$\mathbf{P}$: & regression matrix.\\
$\mathbf{E}$: & endmember matrix.\\
$\mathbf{E}_{0}$: & reference endmember matrix.\\
$\mathbf{A}$: & abundance matrix.\\
$\mathbf{S}$: & scaling factors (matrix).\\
$\mathbf{V}$: & spectral variability dictionary (matrix).\\
$\mathbf{J}$: & coefficients corresponding to $\mathbf{V}$.\\ 
$\mathbf{R}$: & spatial degradation function.\\
$\mathbf{G}$: & spectral response function.\\
$\mathbf{N}_{H}$: & HS noise.\\
$\mathbf{N}_{H}$: & MS noise.\\
$\mathbf{Z}$: & high spatial resolution MS image.\\
$m$: & the number of the considered modality.\\
$c$: & scaling constant.\\
\end{supertabular}

\subsection{Notation}
\begin{supertabular}{ll}
$\norm{\mathbf{X}}_{\F}$ & Forbenius norm of $\mathbf{X}$, obtained by $\sqrt{\sum_{i,j}\mathbf{X}_{i,j}^{2}}$\\
$\norm{\mathbf{X}}_{1,1}$ & $\ell_{1}$ norm of $\mathbf{X}$, obtained by $\sum_{i,j}|\mathbf{X}_{i,j}|$\\
$\norm{\mathbf{X}}_{2,1}$ & $\ell_{2,1}$ norm of $\mathbf{X}$, obtained by $\sum_{i}|\sqrt{\sum_{j}\mathbf{X}_{i,j}^{2}}|$\\
$\norm{\mathbf{X}}_{1/2}$ & $\ell_{1/2}$ norm of $\mathbf{X}$, obtained by $\sum_{i,j}\mathbf{x}_{j}(i)^{1/2}$\\
$\norm{\mathbf{X}}_{q}$ & $\ell_{q}$ norm of $\mathbf{X}$, obtained by $\sum_{i,j}\mathbf{x}_{j}(i)^{q}$\\
$\norm{\mathbf{X}}_{{\rm TV}}$ & ${\rm TV}$ norm of  $\mathbf{X}$, obtained by $\norm{\mathbf{H}_{h}\mathbf{X}+\mathbf{H}_{v}\mathbf{X}}_{2,1}$\\
$\norm{\mathbf{X}}_{0}$ & $\ell_{0}$ norm of $\mathbf{X}$, obtained by $\lim_{p\rightarrow0}\sum_{i,j}|\mathbf{X}_{i,j}|^{p}$\\
$\tr(\mathbf{X})$ & trace of $\mathbf{X}$, obtained by $\sum_{i}\mathbf{X}_{i,i}$\\
$\norm{\mathbf{X}}_{*}$ & nuclear norm of $\mathbf{X}$, obtained by $\tr(\sqrt{\mathbf{X}^{\top}\mathbf{X}})$\\
$\norm{\mathbf{x}}_{2}$ & $\ell_{2}$ norm of $\mathbf{x}$, obtained by $\sqrt{\sum_{j}\mathbf{x}_{j}^{2}}$\\
$\odot$ & the element-wise multiplication operator\\
$\phi_{i}$ & the neighbouring pixels of the target pixel $i$\\
\end{supertabular}

\section{Hyperspectral Image Restoration}
\begin{figure}[!t]
	  \centering
		\includegraphics[width=0.48\textwidth]{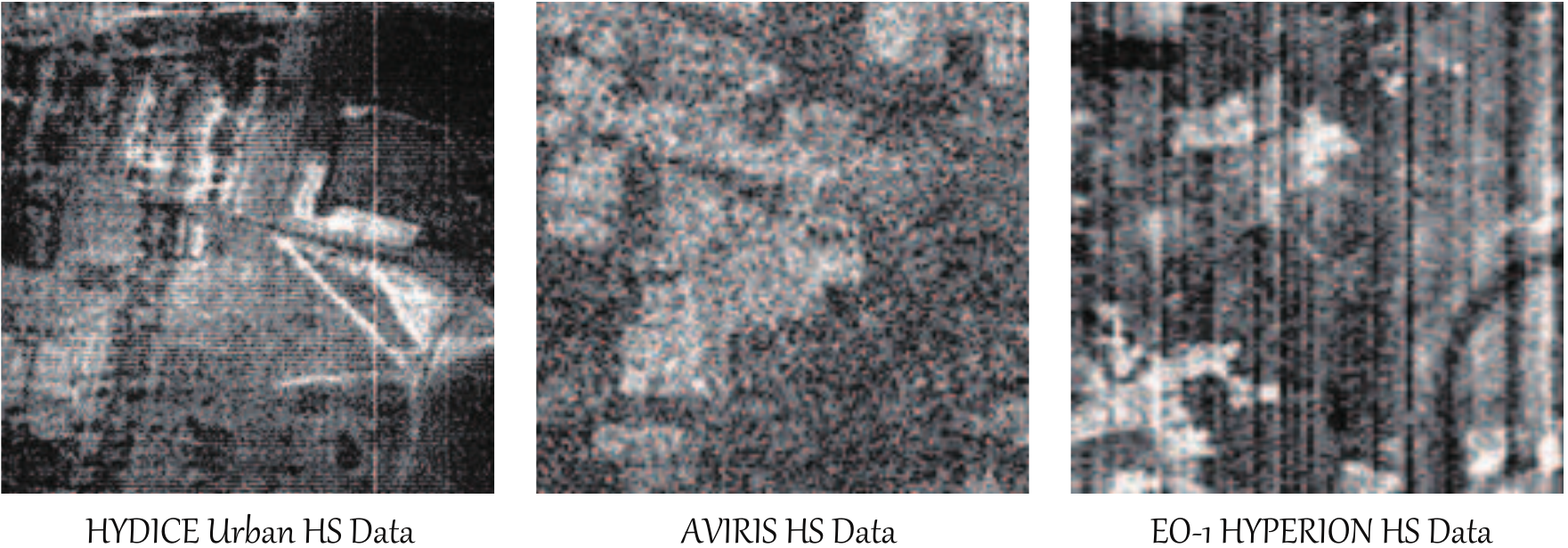}
        \caption{Examples for different noise types of HS Images observed from airborne and spaceborne sensors.}
\label{fig:noiseType}
\end{figure}

Owing to the wealthy spectral information, the HS image has been widely used in different kinds of applications, including urban planning, agriculture, forestry, target detection, and so on. However, due to the limitation of hyperspectral imaging systems and the weather conditions, HS images are always suffering from the pollution of various noise. Fig.~\eqref{fig:noiseType} illustrates different noise types of HS images observed from airborne and spaceborne sensors. Therefore, HS image denoising and restoration is a necessary pre-processing for the subsequent applications to assist the noise.

The statistical distribution of hyperspectral noise is complicated. For instance, the readout noise, which is assumed to obey the Gaussian distribution, is produced by the imaging device (Charge-coupled Device) during the conversation from electrons to the final image~\cite{martin2007anisotropic}. The stripes are generated in the hyperspectral data collected by the pushroom hyperspectral sensors~\cite{datt2003preprocessing}. Due to the weather environment, the obtained HS data are always suffering from the cloud and cloud shadow~\cite{gomez2005cloud}. Besides, the HS images are also suffering from the signal-dependent noise~\cite{acito2011signal}, multiplicative noise, impulsive noise, Laplace noise and so on. Furthermore,  In this paper, we follow the mainstream and focus on the Gaussian noise removal~\cite{Chang1999,yuan2012hyperspectral} and mixed noise removal problem~\cite{zhang2013hyperspectral,he2015total}.

Since 1980s, researchers have paid attention to the HS noise analysis. For example, maximum noise fraction (MNF) transformation ~\cite{green1988transformation} noise-adjusted principal component analysis~\cite{Chang1999} are utilized to extract high-quality components for the subsequent classification and reject the low-quality components. Following the mainstream of gray/color image denoising in computer vision society, various state-of-the-art technologies, such as wavelets~\cite{rasti2014wavelet}, sparse representation~\cite{qian2012hyperspectral,li2016noise}, TV~\cite{yuan2012hyperspectral,wu2017structure}, non-local means processing~\cite{CVPR2014Meng,he2018non}, low-rank matrix representation~\cite{zhang2013hyperspectral,he2015hyperspectral,xie2016hyperspectral,he2015total}, tensor representation~\cite{xie2017kronecker,chen2020TCYB,chen2020TGRS}, DL~\cite{chang2018hsi,yuanHSID-CNN2018} and so on.

The non-convex regularized methods have also been developed for HS image restoration. From~\cite{zhang2013hyperspectral} the HS images are assumed to be corrupted by the additive noise, including Gaussian noise, stripes, deadlines, pixel missing, impulse noise and so on. The observation model is formulated as:
\begin{equation}
\label{eq:ob}
\tensor{Y} = \tensor{X} + \tensor{N}_{G} + \tensor{N}_{S},
\end{equation}
where $\tensor{Y}$ represents the observed noisy image, $\tensor{X}$ stands for the latent clean image, $\tensor{N}_{G}$ is the Gaussian noise, and $\tensor{N}_{S}$ is the sparse noise, including stripes, deadlines, pixel missing and impulse noise. Typically, when sparse noise $\tensor{N}_{S}$ is omitted, the model \eqref{eq:ob} is degraded to the Gaussian noise removal problem. From~\cite{zhang2013hyperspectral}, the low-rank property exploration of clean image $\tensor{X}$ has attracted much attention and achieved state-of-the-art HS image denoising performance~\cite{he2018non,chen2020TGRS}. Generally speaking, the mainstreams are from two points. Firstly, how to explore the low-rank property of clean image $\tensor{X}$. Until now, spectral low-rank property~\cite{zhang2013hyperspectral,he2015hyperspectral}, spatial low-rank property~\cite{liu2012denoising,chen2020TCYB} and non-local low-rank property~\cite{xie2017kronecker,he2018non} have been well studied. Furthermore, how to balance the contribution of different low-rank properties is also a key problem~\cite{chang2017hyper,he2018non}. Secondly, the low-rank constraint of $\tensor{X}$ formulates a non-convex optimization problem. Therein, how to efficiently solve the rank constraint non-convex problem is another key problem. In the next subsection, we review several outstanding works and illustrate how these works formulate the low-rank modeling, and how to solve the non-convex problem.

\begin{table}[!t]
  \centering
  \caption{Prior properties of the selected methods. One, two, and three $\bullet$ denote the low, medium, and high intensity of prior information, respectively.}
  \footnotesize
    \begin{tabular}{c||c|c|c||c|c}
     \toprule[1.5pt]
     \multirow{2}{*}{Methods} & \multicolumn{3}{c||}{Low-rankness} & \multicolumn{2}{c}{Local smoothness}\\
     \cline{2-6}& spatial & spectral & non-local & spatial & spectral   \\
    \hline
    LRTA      & $\bullet \bullet$   & $\bullet \bullet$     &           &             &        \\
    NAILRMA   &           & $\bullet \bullet$     &           &             &          \\
    TDL       &           & $\bullet$            & $\bullet \bullet \bullet$   &             &       \\
    FastHyDe  &           & $\bullet \bullet$     & $\bullet$   &             &        \\
    NGmeet    &           & $\bullet \bullet$     & $\bullet \bullet$   &             &       \\
    \hline
    LRMR      &           & $\bullet \bullet$      &           &             &        \\
    LRTV      &           & $\bullet \bullet$      &           & $\bullet \bullet$      &        \\
    LRTDTV    & $\bullet$    & $\bullet \bullet$      &           & $\bullet \bullet$      & $\bullet \bullet$      \\
    LRTDGS    & $\bullet$   & $\bullet \bullet \bullet$     &           & $\bullet \bullet \bullet$     &      \\
    LRTF-FS   &           & $\bullet \bullet \bullet$     &           & $\bullet \bullet \bullet$     & $\bullet \bullet$     \\
    \bottomrule[1.5pt]
    \end{tabular}%
  \label{tab:Denoising_property}%
\end{table}%

\begin{table*}[!t]
  \centering
  \caption{The restoration results of different selected methods on Gaussian noise and mixed noised, respectively.}
  \footnotesize
    \begin{tabular}{c||ccccc|ccccc}
     \toprule[1.5pt]
    \multirow{2}{*}{Index} & \multicolumn{5}{c|}{Gaussian noise removal} & \multicolumn{5}{c}{Mixed noise removal} \\
\cline{2-11} & LRTA  & NAILRMA & TDL   & FastHyDe & NGmeet & LRMR  & LRTV  & LRTDTV & LRTDGS & LRTF-FS \\
    \hline
    PSNR  & 25.99 & 32.81 & 32.11 & 33.51 & 33.82 & 32.22 & 33.05 & 32.34      &33.33       & 33.26 \\
    SSIM  & 0.7095 & 0.9519 & 0.9443 & 0.9601 & 0.9607 & 0.9401 & 0.9459 &0.9335       &0.9596       & 0.9549 \\
    MSA   & 11.35 & 4.75  & 4.72  & 4.42  & 4.38  & 4.95  & 5.06  & 4.09 & 4.24     & 4.44 \\
    \bottomrule[1.5pt]
    \end{tabular}%
  \label{tab:DenoisResults}%
\end{table*}%

\subsection{Gaussian Noise Removal}
In this section, five methods are selected to represent the state-of-the-art HS image Gaussian noise removal approaches. These methods utilize different low-rank matrix/tensor decomposition models to exploit the spatial, spectral, or non-local low-rank properties of the original clean HS image. The properties of these five methods are summarized in Table \ref{tab:Denoising_property}. The five methods are briefly described in the following.

\subsubsection{LRTA}\footnote{\url{https://www.sandia.gov/tgkolda/TensorToolbox/}}
On the basis of the observation model~\eqref{eq:ob} with ignoring sparse noise $S$, low-rank tensor approximation (LRMA)~\cite{renard2008denoising} tries to restore the HS image from the following objective function
\begin{equation}
\label{LRTA}
\min_{\tensor{X}} \|\tensor{Y}-  \tensor{X}\|_{\F}^2\;\;
{\rm s.t.}\; \tensor{X} = \tensor{O} \times_1 \mat{A} \times_2 \mat{B} \times_3 \mat{C},
\end{equation}
where $\tensor{X} = \tensor{O} \times_1 \mat{A} \times_2 \mat{B} \times_3 \mat{C},$ is the Tucker decomposition,
$\tensor{O} \in  \mathbb{R}^{r_1 \times r_2 \times r_3}$ stands for the core tensor, and $\mat{A} \in \mathbb{R}^{M \times r_1}, \mat{B}  \in  \mathbb{R}^{N \times r_2}, \mat{C}  \in  \mathbb{R}^{B \times r_3}$ are the factors related to different dimensions. With the rank $(r_1, r_2, r_3)$ of Tucker decomposition set in advance, the LRTA model \eqref{LRTA} can simultaneously capture the global spatial and spectral low-rank properties. \eqref{LRTA} provides a simple general model for different kinds of low-rank matrix/tensor decomposition based HS image denoising methods, that's to say, we can change the Tucker decomposition constraint of $\tensor{X}$ to different kinds of matrix/tensor decomposition, such as canonical polyadic (CP) decomposition, tensor train decomposition, tensor ring decomposition and so on.

\subsubsection{NAILRMA}\footnote{\url{https://sites.google.com/site/rshewei/home}}
Noise-adjusted iterative LRMA (NAILRMA)~\cite{he2015hyperspectral} method assumes that the spectral low-rank property is more important than that of spatial ones, and therefore simply spectral low-rank regularizer is utilized to restrict the original spectral image $\tensor{X}$. From other works~\cite{zhang2020}, it also indicates that the spatial TV regularizer is more important than that of the spatial low-rank regularizer. In HS images, the similar signatures representing the same class also appear in the nearby spatial location. To enhance the spectral low-rank property, NAILRMA segmented the HS image into spatial overlapping patches and process each patch individually. The noise intensity in different bands of the HS image is different, which is a big challenge appearing in the HS image Gaussian noise removal, is mitigated by the noise-adjusted iterative strategy~\cite{he2015hyperspectral}. At last, the randomized singular value decomposition (RSVD) is utilized to solve the non-convex low-rank approximation problem.

\subsubsection{TDL}\footnote{\url{http://gr.xjtu.edu.cn/web/dymeng/}}
Tensor dictionary learning (TDL) combines the non-local regularization and low-rank tensor approximation. The noisy HS image is firstly segmented into spatial overlapping patches, and the similar patches are clustered together to formulate a higher-order tensor. In this way, the non-local spatial information is collected. Then the higher-order tensors are denoised as the same of~\eqref{LRTA}, and finally the denoised higher-order tensors are utilized to formulate the final denoised HS image. TDL represents the first method to exploit non-local low-rank property, and the subsequent methods LLRT~\cite{chang2017hyper}, KBR~\cite{xie2017kronecker}, and NLTR~\cite{chen2020TGRS} also achieve remarkable HS image Gaussian noise removal results.

\subsubsection{FastHyDe}\footnote{\url{https://github.com/LinaZhuang/FastHyDe_FastHyIn}}
The main difference between HS images and color/multispectral images is the number of spectral bands. To eliminate this difference and utilize well developed color/multispectral denoising methods for HS image denoising, Zhuang \textit{et.al} proposed the fast HS denoising (FastHyDe)~\cite{zhuang2018fast} method by translating the HS image to low-dimensional reduced image via SVD. By this translation, various state-of-the-art color/multispectral image denoising methods, such as wavelets~\cite{Chen2011TGRS} and BM3D~\cite{zhuang2018fast} are used to denoise the reduced image. Finally, the denoised reduced image is translated back to the denoised HS image via inverse SVD. Generally speaking, under the framework of FastHyDe, the HS image noise removal task is linked to the development of color/multispectral image noise removal tasks.

\subsubsection{NGmeet}\footnote{\url{https://github.com/quanmingyao/NGMeet}}
Spatial non-local low-rank regularizer can produce state-of-the-art HS image noise removal performance. However, as the increase of spectral number, the time cost of non-local related methods also increase incredibly~\cite{chang2017hyper,xie2017kronecker}. Non-local meets global (NGmeet) method also tries to translate the HS image to the reduced image, and utilizes a non-local low-rank method to denoise the reduced image. Different from FastHyDe, NGmeet tries to perfect the framework by iteratively eliminating the error caused by SVD on the noisy HS image, and automatically estimating the spectral rank of the reduced image.

\subsection{Mixed Noise Removal}
In this section, we select five representative methods for the HS image mixed noise removal. These methods are on the basis of the observation model~\eqref{eq:ob}. We focus on the non-convex low-rank regularizer of original image $\tensor{X}$. The properties of these five methods are summarized in Table \ref{tab:Denoising_property}.

\subsubsection{LRMR}
Zhang \textit{et al.} firstly introduced the observation model \eqref{eq:ob} to analysis of complex HS noise~\cite{zhang2013hyperspectral}. LRMR tries to restore the original clean image $\tensor{X}$ from the noisy image via low-rank and sparse decomposition model as follows:
\begin{equation}
\label{LRMR}
\min_{\mat{X}} \norm{\mat{Y}-  \mat{X} - \mat{N}_{S}}_{\F}^2+ \lambda_{1} rank(\mat{X}) + \lambda_{2} card(\mat{N}_{S}),
\end{equation}
where $\mat{Y}, \mat{X}, \mat{N}_{S}$ are the reshaped matrices of $\tensor{Y}, \tensor{X}, \tensor{N}_{S}$ along the spectral dimension, respectively, $\lambda_1$ and $\lambda_2$ are the parameters to trade-off the contributions of $rank(\mat{X})$ and non-zero elements $card(\mat{N}_{S})$. LRMR utilizes ``GoDec'' algorithm~\cite{zhou2011godec} to alternatively update non-convex constraint $\mat{X}$ and $\mat{N}_{S}$, and finally obtains the restored image. To improve the efficiency of the optimization to~\eqref{LRMR}, several non-convex substitutions, such as reweighted nuclear norm~\cite{xie2016hyperspectral}, $\gamma$-norm~\cite{chenyongyong2017TGRS}, smooth rank approximation~\cite{Ye2019TGRS} and normalized $\epsilon$-Penalty~\cite{Xie2020TIP}, are further developed to exploit the spectral low-rank property of $\mat{X}$.

\subsubsection{LRTV}
Low-rank total variation (LRTV) claimed that spectral low-rank property is not enough to describe the property and HS images, and therein introduced TV to explore the spatial smoothness via TV. Generally, low-rank regularization and TV are the two most studied regularizers, and the combination of them to produce the state-of-the-art HS image mixed noise removal is becoming popular. Most of the following works try to either improve the low-rank modeling~\cite{Fan2018TGRS,Wang2018} or the smoothness modeling~\cite{he2017Jstars,chen2020TCYB,ZengTGRS2020} of the HS image to further improve the restoration accuracy. To further combine low-rank and TV, the low-rank exploration of the HS difference image is also developed~\cite{Sun2017letter,Peng2020TIP}.

\subsubsection{LRTDTV}
Low-rank tensor decomposition with TV (LRTDTV)~\cite{Wang2018} tries to improve LRTV by utilizing low-rank tensor decomposition to exploit the low-rank property of HS images, meanwhile spatial-spectral TV (SSTV) to explore the spatial and spectral smoothness simultaneously. Although LRTDTV achieved better mixed noise removal results as reported in~\cite{Wang2018}, the spatial rank utilized in LRTDTV is much larger than that of spectral rank. This is mainly because the spatial low-rank property of HS images is not so important compared to the spectral low-rank property. From another side, the spatial non-local low-rank regularization is proved to be more efficient~\cite{zhang2020} than spatial low-rank property for HS restoration problem.

\subsubsection{LRTDGS}\footnote{\url{ https://chenyong1993.github.io/yongchen.github.io/}}
Low-rank tensor decomposition with group sparse regularization (LRTDGS)~\cite{chen2020TCYB} also utilizes low-rank rank tensor decomposition to exploit the low-rank property of HS images. Differently, LRTDGS explores the group sparsity of the difference image instead of SSTV in LRTDTV. From the mathematical modeling, LRTDGS utilizes weighted $ell_{2,1}$ norm regularization to fulfill the row-group sparsity of the difference image.

\subsubsection{LRTF-FR}\footnote{\url{ https://yubangzheng.github.io/homepage/}}
Following the idea of NGmeet~\cite{he2018non}, factor-regularized low-rank tensor factorization (LRTF-FR)~\cite{ZhengTGRS2020} also utilizes matrix decomposition to decouple the spatial and spectral priors. From one side, the spectral signatures of the HS image are assumed to be of smooth structure. From another side, the reduced image is assumed to have a group sparse structure in the difference domain. The optimization model of LRTF-FR is
\begin{equation}
\label{LTRF-FR}
\begin{aligned}
    \min_{\tensor{X},\tensor{N}_{S}}&\norm{\tensor{Y}-  \tensor{X} - \tensor{N}_{S}}_{\F}^2+ \lambda_1\norm{\tensor{X} \times_3 \mat{H}_3}_{2,1}\\
     &+ \lambda_2 \sum_{k=1}^{2} \norm{\tensor{X} \times_k \mat{H}_k}_{\F}^{2} + \lambda_3 \norm{\tensor{N}_{S}}_{1,1},
\end{aligned}
\end{equation}
where $\mat{H}_k, k=1,2,3$ are the first-order difference matrices. Furthermore, in the optimization to \eqref{LTRF-FR}, the reweighted strategy is utilized to update $\ell_{2,1}$ norm and $L\ell_1$ norm to further improve the restored results.

\subsection{Experimental Study}
We choose the HS image from DLR Earth Sensing Imaging Spectrometer (DESIS) installed on the International Space Station (ISS)\cite{alonso2019data} for the experimental study to compare different methods on the Gaussian and mixed noise removal tasks. We remove the noisy bands and select a sub-image of size $400 \times 400 \times 200$ as the clean reference image, which is normalized to $[0,1]$. Firstly, we add Gaussian noise of noise variance $0.1569$ to simulate the Gaussian noisy image, and apply different Gaussian noise removal methods to remove the Gaussian noise. Furthermore, we add salt \& pepper noise and stripes to simulate the mixed noisy image, and apply mixed noise removal methods to remove the mixed noise. As similar in~\cite{chen2020TCYB}, we choose the mean of peak signal-to-noise rate (PSNR) over all bands, the mean of structural similarity (SSIM) over all bands, and the mean of spectral angle mapping (MSA) overall spectral vectors to evaluate the restored results.

\begin{figure*}[!t]
	  \centering
			\includegraphics[width=1\textwidth]{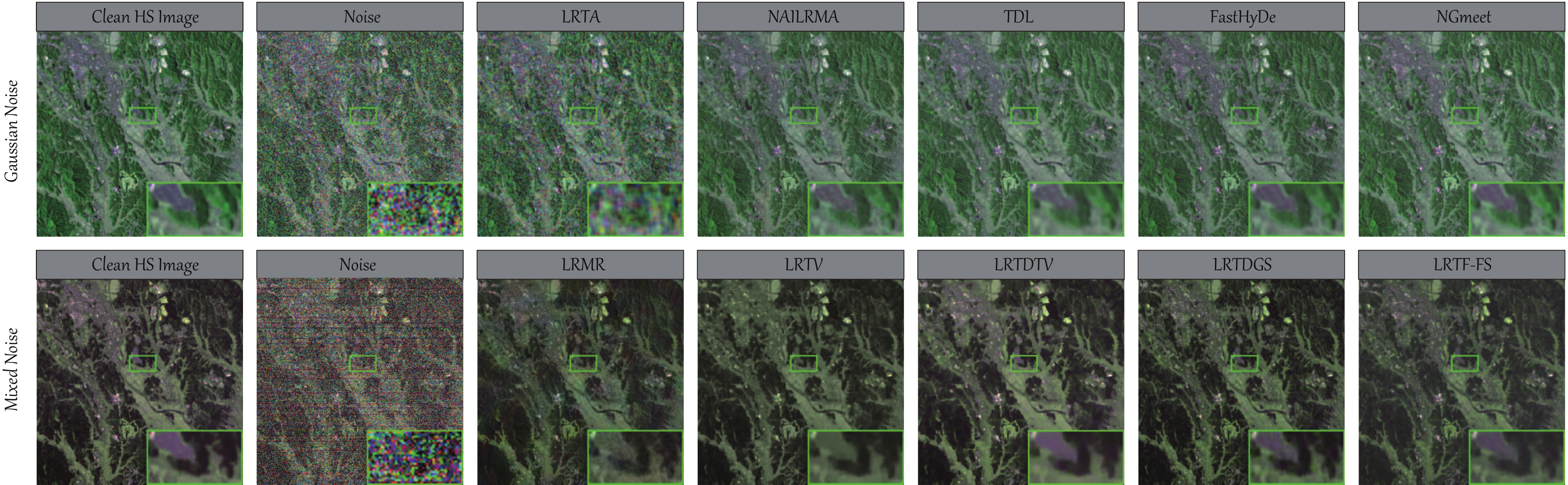}
        \caption{The illustration of different methods on the noise removal results. The first row shows the results of different methods on the Guassian noise remove experiments (R:70, G:100, B:36). The second row displays the results of different methods on the mixed noise remove experiments (R:31, G:80, B:7).}
\label{fig:restoration}
\end{figure*}

Table \ref{tab:DenoisResults} presents the evaluation values of different methods on Gaussian noise and mixed noise removal results, respectively. For the Gaussian noise removal task, NGmeet achieves the best values of three evaluation indices. However, the gap between NGmeet and FastHyDe is limited. For the mixed noise removal task, LRTDGS achieves the best accuracy in PSNR and SSIM values, meanwhile LRTDTV achieved the best MSA value. Combining Tables \ref{tab:Denoising_property} and \ref{tab:DenoisResults}, we can conclude that, firstly, the spectral low-rank prior information is important for HS restoration. Secondly, the contribution of spatial low-rank prior information for HS restoration is limited. Thirdly, on the basis of spectral low-rank regularization, spatial and spectral smoothness prior can further improve the final HS restoration results.

\subsection{Remaining Challenges}
Up to date, many non-convex regularized methods have been proposed to develop the low-rank priors and local smoothness priors, and achieve remarkable HS restoration results for Gaussian and mixed noise removal. However, these methods still face several challenges for further work. We summary these challenges as the following.

\begin{itemize}
    \item \textbf{Efficiency.} Although low-rank related methods have achieved state-of-the-art restoration results, they are time-consuming. For instance, NGmeet and LRTVGS speed more than 10 minutes to process the HS image of size $400 \times 400 \times 200$. Furthermore, the related state-of-the-art restoration methods always exploit multiple priors of the HS image, resulting in the confusion of the parameter chosen. Therein, how to reduce the model complexity and improve the optimization efficiency of the HS image restoration is the key challenge.
    \item \textbf{Scalability.} Previous non-convex related methods always focus on the small HS image processing. However, HS images are used to observe the earth and the spatial size of one scene is usually very large. How to improve the scalability of the restoration approaches is the key challenge. DL provides the possibility for fast and large scale processing of HS images. Whereas DL approaches always rely on the quality of training samples, and the applicable scope of the test data is always limited. To improve the scalability, how to embed well studied non-convex regularizers to the DL architecture should also be further analyzed.

    \item \textbf{Real Application.} Until now, most HS image restoration methods are evaluated on the simulated experiments. However, in most cases, the evaluation indices fail to predict the accuracy of the real HI image restoration results. From another side, the noise distribution in the real noisy HS images is complex. How to testify the related methods on the real HS images should be also further analyzed. From another side, the training samples in the real application are always limited. The blind and unsupervised approaches will become the mainstream of future real HS image restoration.
\end{itemize}

\section{Dimensionality Reduction}
The HS dimensionality reduction (DR) and feature extraction have long been a fundamental but challenging research topic in HS RS \cite{li2018discriminant,rasti2020feature}. The main reasons mainly lie in the following aspects. Due to the highly-correlated characteristic between spectral bands, the HS images are subjected to information redundancy, which could hurt the ability to discriminate the materials under the certain extremely-conditioned cases (\textit{curse of dimensionality}). Plus, as the HS dimension gradually increases along with the spectral domain, large storage capability and high-performance computing are needed. Furthermore, these dimension-reduced features are usually applied for the high-level classification or detection task \cite{wu2019orsim,wu2020fourier}. Recently, many works based on non-convex modeling have shown to be effective for automatically extracting dimension-reduced features of HS images. Linking with Eq. (\ref{g_eq2}), the DR task can be generalized to the following optimization problem:
\begin{equation}
\label{DR_eq1}
\begin{aligned}
      \mathop{\min}_{f_{\Phi},\mathbf{X}}\frac{1}{2}\norm{f_{\Phi}(\mathbf{Y})-\mathbf{X}}_{\F}^{2}\;\; {\rm s.t.}\; \mathbf{X},f_{\Phi}\in \mathcal{C},
\end{aligned}
\end{equation}
where $f_{\Phi}(\bullet)$ denotes the transformation from the original HS space to dimension-reduced subspaces with the respect to the variable set $\Phi$, and $\mathbf{X}$ is the low-dimensional representations of $\mathbf{Y}$. Revolving around the general form in Eq. (\ref{DR_eq1}), we review currently advanced DR methods from three different aspects: unsupervised, supervised, and semi-supervised models.

\begin{figure*}[!t]
	  \centering
			\includegraphics[width=0.9\textwidth]{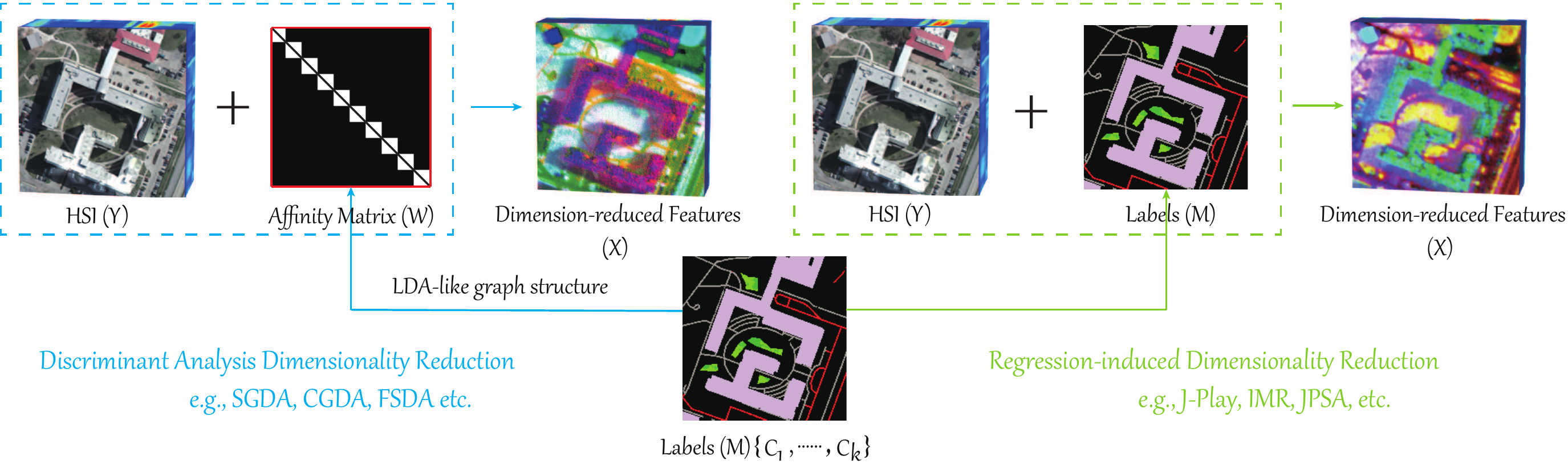}
        \caption{An illustration for supervised DR models in HS images with two different groups: discriminant analysis based DR and regression-induced DR.}
\label{fig:GRSM_SupervisedDR}
\end{figure*}

\subsection{Unsupervised Model}
Non-negative matrix factorization (NMF) \cite{lee2001algorithms} is a common unsupervised learning tool, which has been widely applied in HS DR. These works can be well explained by Eq. (\ref{DR_eq1}), the NMF-based DR problem can be then formulated as
\begin{equation}
\label{DR_eq2}
\begin{aligned}
      \mathop{\min}_{\mathbf{Q}\geq \mathbf{0},\mathbf{X}\geq \mathbf{0}}\frac{1}{2}\norm{\mathbf{Y}-\mathbf{X}\mathbf{Q}}_{\F}^{2} + \Psi(\mathbf{X}) + \Omega(\mathbf{Q}),
\end{aligned}
\end{equation}
where $\mathbf{Q}$ denotes the combination coefficients, $\Phi(\mathbf{X})$ and $\Omega(\mathbf{Q})$ are the potential regularization terms for the variables $\mathbf{X}$ and $\mathbf{Q}$, respectively. Until the current, there have been some advanced NMF-based works in HS DR. Gillis \textit{et al.} \cite{gillis2013sparse} used sparse NMF under approximations for HS data analysis. Yan \textit{et al.} \cite{yan2018novel} proposed a graph-regularized orthogonal NMF (GONMF) model with the application to spatial-spectral DR of HS images. Wen \textit{et al.} \cite{wen2016orthogonal} further extended the GONMF with combining multiple features for HS DR. Rasti \textit{et al.} \cite{rasti2016hyperspectral} designed an orthogonal total variation component analysis (OTVCA) approach for HS feature extraction. Moreover, the HS data are directly regarded as a high-dimensional tensor structure in \cite{an2018tensor}, where the low-rank attribute is fully considered in the process of low-dimensional embedding. In detail, we summarize the regularization and constraints of the above methods as follows:
\begin{itemize}
    \item Sparsity \cite{gillis2013sparse}: $\Omega(\mathbf{Q})=\norm{\mathbf{Q}}_{0}$;
    \item Graph Regularization \cite{yan2018novel}:\\ $\Psi(\mathbf{X})=\tr(\mathbf{X}\mathbf{L}\mathbf{X}^{\top}),\; \rm{s.t.}\; \mathbf{X}\mathbf{X}^{\top}=\mathbf{I}$;
    \item Multi-graph Regularization \cite{wen2016orthogonal}:\\ $\Psi(\mathbf{X})=\sum_{i=1}^{s}\tr(\mathbf{X}\mathbf{L}^{i}\mathbf{X}^{\top}),\; \rm{s.t.}\; \mathbf{X}\mathbf{X}^{\top}=\mathbf{I}$;
    \item Total Variation \cite{rasti2016hyperspectral}:\\ $\Psi(\mathbf{X})=\sum_{i=1}^{r}\norm{\sqrt{(\mathbf{H}_{h}\mathbf{X}_{i})^{2}+(\mathbf{H}_{v}\mathbf{X}_{i})^{2}}}_{1},\\
    \rm{s.t.}\; \mathbf{Q}\mathbf{Q}^{\top}=\mathbf{I}$;
    \item Low-rank Graph \cite{an2018tensor}: $\Psi(\mathbf{X})=\norm{\mathbf{X}}_{*}+\tr(\mathbf{X}\mathbf{L}\mathbf{X}^{\top})$.
\end{itemize}
$\mathbf{L}=\mathbf{D}-\mathbf{W}$ is the Laplacian matrix, where $\mathbf{D}_{i,i}=\sum_{j}\mathbf{W}_{i,j}$ is the degree matrix and $\mathbf{W}$ is the graph (or manifold) structure of $\mathbf{X}$ \cite{belkin2003laplacian}. $\norm{\bullet}_{0}$, $\norm{\bullet}_{2,1}$, and $\norm{\bullet}_{*}$ denote the $\ell_{0}$-norm \cite{aharon2006k}, $\ell_{2,1}$-norm \cite{nie2010efficient}, and nuclear norm \cite{recht2010guaranteed}, respectively.

Another type of unsupervised DR approaches is \textit{graph embedding}, also known as \textit{manifold learning}, which can be also grouped into Eq. (\ref{DR_eq1}) well (according to \cite{wu2017joint}):
\begin{equation}
\label{DR_eq3}
\begin{aligned}
      \mathop{\min}_{\mathbf{U},\mathbf{X}}\frac{1}{2}\norm{\mathbf{X}-\mathbf{U}\mathbf{Y}}_{\F}^{2} + \Psi(\mathbf{X}) + \Omega(\mathbf{U})\; {\rm s.t.}\; \mathbf{X}\mathbf{X}^{\top}=\mathbf{I},
\end{aligned}
\end{equation}
where $\mathbf{U}$ denotes the to-be-estimated projection matrix that bridges the high-dimensional data $\mathbf{Y}$ with the low-dimensional embedding $\mathbf{X}$. The regularization term for the variable $\mathbf{U}$ can be usually expressed as
\begin{equation}
\label{DR_eq4}
\begin{aligned}
      \Omega(\mathbf{U})=\tr(\mathbf{U}\mathbf{Y}\mathbf{L}\mathbf{Y}^{\top}\mathbf{U}^{\top})+\norm{\mathbf{U}}_{\F}^{2},
\end{aligned}
\end{equation}
while the regularizer with respect to $\mathbf{X}$ can be given by
\begin{equation}
\label{DR_eq5}
\begin{aligned}
      \Psi(\mathbf{X})=\tr(\mathbf{X}\mathbf{L}\mathbf{X}^{\top}).
\end{aligned}
\end{equation}
The main difference between these \textit{manifold learning}-based DR approaches lies in the graph construction, i.e., $\mathbf{W}$. Ma \textit{et al.} \cite{ma2010local} integrated the KNN classifier with several representative \textit{manifold learning} algorithms, i.e., locally linear embedding \cite{roweis2000nonlinear}, Laplacian eigenmaps \cite{belkin2003laplacian}, and local tangent space alignment \cite{zhang2004principal}, for HS image classification. Huang \textit{et al.} \cite{huang2015dimensionality} embedded the sparse graph structure, which is performed by solving a $\ell_{1}$-norm optimization problem, for the DR of HS images. He \textit{et al.} \cite{he2016weighted} extended the work of \cite{huang2015dimensionality} by generating a weighted sparse graph. Hong \textit{et al.} \cite{hong2017learning} developed a new spatial-spectral graph for the DR of HS images, called RLMR, by jointly taking neighbouring pixels of a target pixel in spatial and spectral domains into account. An \textit{et al.} \cite{an2018patch} attempted to learn the low-dimensional tensorized HS representations on a sparse and low-rank graph. To sum up, core graphs of the aforementioned methods can be obtained by
\begin{itemize}
    \item Sparse Graph \cite{huang2015dimensionality}: $\min_{\mathbf{W}}\norm{\mathbf{W}}_{1,1}, \;\rm{s.t.}\; \mathbf{Y}=\mathbf{Y}\mathbf{W}$;
    \item Weighted Sparse Graph \cite{he2016weighted}:\\ $\min_{\mathbf{W}}\norm{\mathbf{d}\odot \mathbf{W}}_{1,1}, \;\rm{s.t.}\; \mathbf{Y}=\mathbf{Y}\mathbf{W},$\\
    where $\mathbf{d}$ denotes a weighted matrix on $\mathbf{W}$ and $\odot$ is the element-wise multiplication operator;
    \item Spatial-spectral Graph \cite{hong2017learning}: \\
    $\mathop{\min}_{\mathbf{w}_{i,0}}\sum_{j\in \phi_{i}^{spa}}\norm{\mathbf{y}_{i,j}-\sum_{k\in \phi_{i}^{spe}}\mathbf{y}_{i,k}w_{i,k,j}}_{2}^{2} \\
    {\rm s.t.} \; \norm{\sum_{k\in \phi_{i}^{spe}}\mathbf{y}_{i,k}(4w_{i,k,0}-\sum_{k=1}^{4}w_{i,k,j})}_{2}^{2}\leq \eta, \\
    \qquad \mathbf{w}_{i,j}^{\T}\mathbf{w}_{i,j}=1,$\\
    where $\phi_{i}^{spa}$ and $\phi_{i}^{spe}$ denote the neighbouring pixels in spatial and spectral spaces, respectively;
    \item Sparse and Low-rank Graph \cite{an2018patch}:\\ $\min_{\mathbf{W}}\norm{\mathbf{W}}_{1,1}+\norm{\mathbf{W}}_{*}, \;\rm{s.t.}\; \mathbf{Y}=\mathbf{Y}\mathbf{W}$.
\end{itemize}

\subsection{Supervised Model}

Unlike unsupervised DR that relies on embedding various priors to reduce the dimension of HS data, supervised models are capable of better learning class-separable low-dimensional representations via the use of label information. The supervised DR models can be described from two different categories in this subsection, as shown in Fig. \ref{fig:GRSM_SupervisedDR}. A typical group is the \textit{discriminant analysis} \cite{li2018discriminant} closely related to \textit{graph embedding} and \textit{manifold learning}. Intuitively speaking, these methods belong to a special case of unsupervised \textit{graph embedding}, which means they can be well explained by Eq. (\ref{DR_eq3}). The main difference lies in that the labels are used for constructing the graph structure, i.e., $\mathbf{W}$, thereby yielding a more discriminative subspace.

In the supervised DR, a direct graph structure is written as
\begin{equation}
\label{DR_eq6}
    {\bf W}_{ij}=
    \begin{cases}
      \begin{aligned}
      1, \; \; & \text{if ${\bf y}_{i}$ and ${\bf y}_{j} \in C_{k}$;}\\
      0, \; \; & \text{otherwise,}
      \end{aligned}
    \end{cases}
\end{equation}
where $C_{k}$ means the sample set of the $k$-th class. Furthermore, more advanced supervised graphs have been developed to better represent the HS data in a low-dimensional subspace, such as sparse graph discriminant analysis \cite{ly2013sparse}, collaborative graph discriminant analysis \cite{ly2014collaborative}, feature space discriminant analysis (FSDA) \cite{imani2015feature}, spatial-spectral local discriminant embedding \cite{huang2019spatial}. These approaches sought to construct a \textit{soft} graph instead of a \textit{hard} graph in Eq. (\ref{DR_eq6}). That is, the graph is built by using radial basis function (RBF) to measure the similarity between samples belonging to the same class \cite{hong2020graph1}:
\begin{equation}
\label{DR_eq7}
    {\bf W}_{ij}=
    \begin{cases}
      \begin{aligned}
      \exp\frac{-\norm{\mathbf{y}_{i}-\mathbf{y}_{j}}_{2}^{2}}{2\sigma^{2}}, \; \; & \text{if ${\bf y}_{i}$ and ${\bf y}_{j} \in C_{k}$;}\\
      0, \; \; & \text{otherwise,}
      \end{aligned}
    \end{cases}
\end{equation}
or by solving $\ell_{1}$-norm or $\ell_{2}$-norm optimization functions in the same class set, e.g., \cite{ly2013sparse}, \cite{ly2014collaborative}.

The DR behavior can be also modeled from a regression perspective by directly connecting samples and labels \cite{hong2019regression}, which provides a new insight into the research of the supervised HS DR. A general form for the regression-based supervised DR model can be formulated as
\begin{equation}
\label{DR_eq8}
\begin{aligned}
      \mathop{\min}_{\mathbf{P}, \mathbf{U}}&\frac{1}{2}\norm{\mathbf{M}-\mathbf{P}\mathbf{X}}_{\F}^{2}+\Psi(\mathbf{P})+\Omega(\mathbf{U})\\
      &{\rm s.t.}\; \mathbf{X}=\mathbf{U}\mathbf{Y},\; \mathbf{U}\mathbf{U}^{\top}=\mathbf{I},
\end{aligned}
\end{equation}
where the variable $\mathbf{P}$ denotes the regression coefficients or basis signals, and $\mathbf{M}$ is the one-hot encoded matrix obtained by labels. Eq. (\ref{DR_eq8}) can be, to some extent, regarded as an interpretable linearized artificial neural network (ANN) mode (shallow network). Ji \textit{et al.} \cite{ji2009linear} jointly performed DR and classification, which is a good fit for Eq. (\ref{DR_eq8}) with $\Psi(\mathbf{P})=\norm{\mathbf{P}}_{\F}^{2}$. To enhance the spectrally discriminative ability, Hong \textit{et al.} \cite{hong2018joint} employed a LDA-like graph on the basis of \cite{ji2009linear} to regularize the low-dimensional representations in a Laplacian matrix form, i.e., $\Omega(\mathbf{U})=\tr(\mathbf{U}\mathbf{Y}\mathbf{L}\mathbf{Y}^{\top}\mathbf{U}^{\top})$. In the same work \cite{hong2018joint}, Hong \textit{et al.} further extended their model to a deep version, called JPlay, with a $k$-layered linear regression: 
\begin{equation}
\label{DR_eq9}
\begin{aligned}
      &\mathop{\min}_{\mathbf{P}, \{\mathbf{U}_{i}\}_{i=1}^{k}}\frac{1}{2}\norm{\mathbf{M}-\mathbf{P}\mathbf{X}_{i}}_{\F}^{2}+\Psi(\mathbf{P})+\Omega(\{\mathbf{U}_{i}\}_{i=1}^{k})\\
      &{\rm s.t.}\; \mathbf{X}_{i}=\mathbf{U}_{i}\mathbf{X}_{i-1},\; \mathbf{X}_{1}=\mathbf{U}_{1}\mathbf{Y},\; \mathbf{X}_{i}\geq \mathbf{0},\; \norm{\mathbf{x}_{i}}_{2}\leq 1,
\end{aligned}
\end{equation}
with $\Psi(\mathbf{P})=\norm{\mathbf{P}}_{\F}^{2}$ and
\begin{equation*}
\begin{aligned}
\Omega(\{\mathbf{U}_{i}\}_{i=1}^{k})&=\sum_{i=1}^{k}\tr(\mathbf{U}_{i}\mathbf{X}_{i-1}\mathbf{L}\mathbf{X}_{i-1}^{\top}\mathbf{U}_{i}^{\top}) \\
 &+\sum_{i=1}^{k}\norm{\mathbf{X}_{i-1}-\mathbf{U}_{i}^{\top}\mathbf{U}_{i}\mathbf{X}_{i-1}}_{\F}^{2}.
\end{aligned}
\end{equation*}
The J-Play attempts to open the ``black box'' of deep networks in an explainable way by multi-layered linearized modeling. With explicit mappings and physically meaningful priors, the non-convex J-Play takes a big step towards the interpretable AI model.

\subsection{Semi-supervised Model}
Due to the fact that labeling samples is extremely expensive, particularly for RS images covering a large geographic region, the joint use of labeled and unlabeled information then becomes crucial in DR and classification. 

A simple and feasible strategy for semi-supervised learning is to integrate supervised and unsupervised techniques, e.g., LDA and locality preserving projections \cite{he2004locality}. By simultaneously constructing graphs of labeled and unlabeled samples (e.g., using Eqs. (\ref{DR_eq6}) and (\ref{DR_eq7}), respectively), Eq. (\ref{DR_eq3}) can be easily extended to a semi-supervised version, leading to semi-supervised discriminant analysis (SSDA) \cite{liao2012semisupervised}. Zhao \textit{et al.} \cite{zhao2014general} further improved the SSDA performance by using ``soft'' (or ``pseudo'') labels predicted by label propagation instead of directly using unsupervised similarities between unlabeled samples. Similarly, Wu \textit{et al.} \cite{wu2018semi} generated pseudo-labels using the Dirichlet process mixing model and achieved a novel SSDA approach to learn the low-dimensional HS embedding. These above-mentioned methods are performed surrounding various hand-crafted graph structures ($\mathbf{W}$). 

\begin{figure}[!t]
	  \centering
			\includegraphics[width=0.35\textwidth]{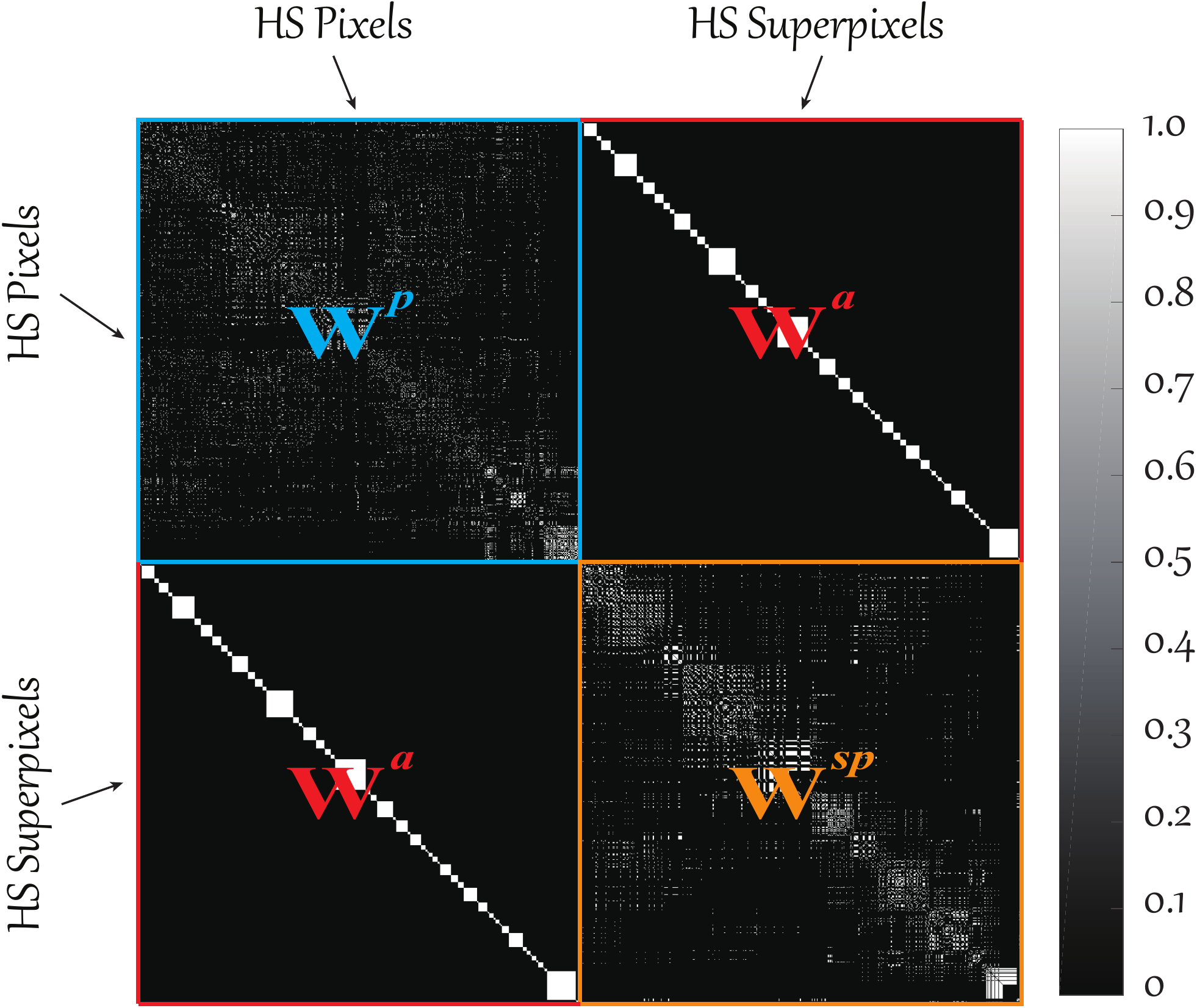}
        \caption{An example to clarify the graph structure of JPSA method, where $\mathbf{W}^{p}$ and $\mathbf{W}^{sp}$ denote the pixel-wise and superpixel-wise subgraphs as well as $\mathbf{W}^{a}$ is the aligned graph between pixels and superpixels.}
\label{fig:graph}
\end{figure}

A different idea is to simulate the brain-like or human-like behaviors in the semi-supervised DR task. It is well known that the feedback reward is a key component that forms the intelligent information processing system. Inspired by it, \cite{hong2019learning} developed an iterative multitask learning (IMR) framework by adaptively learning the label propagation (LP) on graphs to simulate the feedback mechanism, thereby achieving the HS DR process more effectively and efficiently. The IMR is a semi-supervised extension of Eq. (\ref{DR_eq8}) with graph learning, which can be generally modeled as 
\begin{equation}
\label{DR_eq10}
\begin{aligned}
       \mathop{\min}_{\mathbf{P},\mathbf{U},\mathbf{L}}&\sum_{j=1}^{2}\norm{\mathbf{M}_{j}-\mathbf{P}\mathbf{U}\mathbf{Y}_{j}}_{\F}^{2}+\Psi(\mathbf{P})+\Omega(\mathbf{U})\\
       &\mathrm{s.t.} \; \mathbf{U}\mathbf{U}^{\top}=\mathbf{I}, \; \mathbf{L}\in \mathcal{C},
\end{aligned}
\end{equation}
where $\mathbf{Y}_{1}$ and $\mathbf{Y}_{2}$ denote the labeled and unlabeled samples from $\mathbf{Y}$, respectively. $\Psi(\mathbf{P})=\norm{\mathbf{P}}_{\F}^{2}$ and $\Omega(\mathbf{U})=\tr(\mathbf{U}\mathbf{Y}\mathbf{L}\mathbf{Y}^{\top}\mathbf{U}^{\top})$. The non-convex constraint $\mathcal{C}$ with the respect to the variable $\mathbf{L}$ can be summarized as
\begin{equation*}
\begin{aligned}
       \mathcal{C}:=\{\mathbf{L}=\mathbf{L}^{\top}, \; \mathbf{L}_{p,q,p\neq q}\preceq 0, \;\mathbf{L}_{p,q,p=q}\succeq 0, \;\tr(\mathbf{L})=c\},
\end{aligned}
\end{equation*}
where $c>0$ is a scaling constant. Eq. (\ref{DR_eq10}) is a typical data-driven graph learning model, which is capable of automatically learning the graph structure from the data without any hand-crafted priors. By using the iterative strategy to simulate the feedback system, $\mathbf{M}_{2}^{(t+1)}$ in the $t$$+$$1$-step can be updated by the graph-based LP on the learned graph of the $t$-step $\mathbf{W}^{(t)}$:
\begin{equation}
\label{DR_eq11}
\begin{aligned}
\cdots\cdots\mathbf{M}_{2}^{(t+1)}\leftarrow\mathbf{W}^{(t)}\leftarrow\mathbf{M}_{2}^{(t)}\cdots\cdots.
\end{aligned}
\end{equation}

Besides, another intelligent feature extraction algorithm, named JPSA, which is extended from \cite{hong2018joint}, was presented in \cite{hong2020joint} by the attempts to align pixels and superpixels for spatial-spectral semi-supervised HS DR. JPSA basically follows the JPlay framework and the major difference is the graph structure $\mathbf{W}$. The graph in JPSA consists of not only pixel-wise and superpixel-wise similarities but also aligned components between pixels and superpixels. Fig. \ref{fig:graph} gives an example to clarify the graph structure of JPSA. Note that the JPSA's graph can be seen as a full data-driven structure, which can, to a great extent, self-express the intrinsic properties of HS data and further achieves intelligent information extraction and DR.

\begin{table}[!t]
\centering
\caption{Quantitative comparison of different DR algorithms in terms of OA, AA, and $\kappa$ using the NN classifier on the Indian Pines dataset. The best one is shown in bold.}
\resizebox{0.45\textwidth}{!}{
\begin{tabular}{c||c|ccc}
\toprule[1.5pt] Methods & dimension & OA (\%) & AA (\%) & $\kappa$ \\
\hline \hline
OSF & 220 & 65.89 & 75.71 & 0.6148\\
OTVCA \cite{rasti2016hyperspectral} & 16 & 74.18 & 77.61 & 0.7228\\
RLMR \cite{hong2017learning} & 20 & 83.75 & 86.90 & 0.8147\\
FSDA \cite{imani2015feature} & 15 & 64.14 & 74.52 & 0.5964\\
JPlay \cite{hong2018joint} & 20 & 83.92 & 89.35 & 0.8169\\
IMR \cite{hong2019learning} & 20 & 82.80 & 86.27 & 0.8033\\
JPSA \cite{hong2020joint} & 20 & \bf 92.98 & \bf 95.40 & \bf 0.9197\\
\bottomrule[1.5pt]
\end{tabular}}
\label{tab:DR}
\end{table}

\subsection{Experimental Study}
Classification is explored as a potential application to evaluate the performance of state-of-the-art (SOTA) DR algorithms, including original spectral features (OSF), OTVCA\footnote{\url{https://github.com/danfenghong/HyFTech}} \cite{rasti2016hyperspectral}, RLMR\footnote{\url{https://github.com/danfenghong/IEEE_JSTARS_RLML}} \cite{hong2017learning}, FSDA \cite{imani2015feature}, JPlay\footnote{\url{https://github.com/danfenghong/ECCV2018_J-Play}} \cite{hong2018joint}, IMR \cite{hong2019learning}, and JPSA \cite{hong2020joint}. Experiments are performed on the Indian Pine data using the nearest neighbor (NN) classifier in terms of three indices: \textit{Overall Accuracy (OA)}, \textit{Average Accuracy (AA)}, and \textit{Kappa Coefficient} ($\kappa$). The scene consists of $145 \times 145$ pixels and $220$ spectral bands ranging from $0.4 \mu m$ to $2.5 \mu m$. More details regarding training and test samples can be found in \cite{hang2019cascaded}.

Table \ref{tab:DR} lists the quantitative results of different DR methods. Overall, OSF without feature extraction or DR yields the worst classification performance, compared to those SOTA DR methods. This, to a great extend, demonstrates the effectiveness and necessity of DR in the HS image classification task. It is worth noting that the approaches with spatial-spectral modeling, e.g., OTVCA, RLMR, JPSA, tend to obtain better classification results. The performance of RLMR is superior to that of OTVCA, owing to the full consideration of the neighboring information in a graph form rather than the smoothing operation only modeled by the TV regularization. As a linearized deep model, supervised JPlay obviously performs better than others, especially FSDA that is also a supervised DR model. More importantly, the JPSA with a semi-supervised learning strategy dramatically outperforms other competitors, since it can jointly learn richer representations from both pixels and superpixels by means of spatial-spectral manifold alignment and deep regressive regression.

\subsection{Remaining Challenges}
Although extensive SOTA methods have recently shown the effectiveness and superiority in the HS DR and classification, there is still a long way to go towards the AI-guided intelligent information processing. We herein summarize the potential remaining challenges briefly.
\begin{itemize}
    \item \textbf{Optimal Subspace Dimension.} Subspace dimension is a crucial parameter in DR, which is determined experimentally and empirically in most of existing methods. Despite some parameter estimation algorithms, e.g., intrinsic dimension \cite{levina2005maximum}, subspace identification \cite{bioucas2008hyperspectral}, they fail to avoid the pre-survey of various prior knowledge and human intervention in the dimension estimation process.
    \item \textbf{Effects of Noises.} HS images usually suffer from noise degradation in remotely sensed imaging. These noises are complex and closely associated with spectral signatures. Therefore, separating noises from HS data effectively and reducing the noise sensitivity (or preserving spectral discrimination) in the DR process remains challenging. 
    \item \textbf{Robustness and Generalization.} Robust estimation and advanced non-convex regularizers have been widely applied to model the DR behavior, yet the complex noise type, the limited training samples, and noisy labels hinder the robustness and generalization ability to be further improved. For this reason, more robust and intelligent models should be developed in either theory or practice emphatically in the next generation DR technique.
\end{itemize}

\begin{figure*}[!t]
	  \centering
			\includegraphics[width=0.9\textwidth]{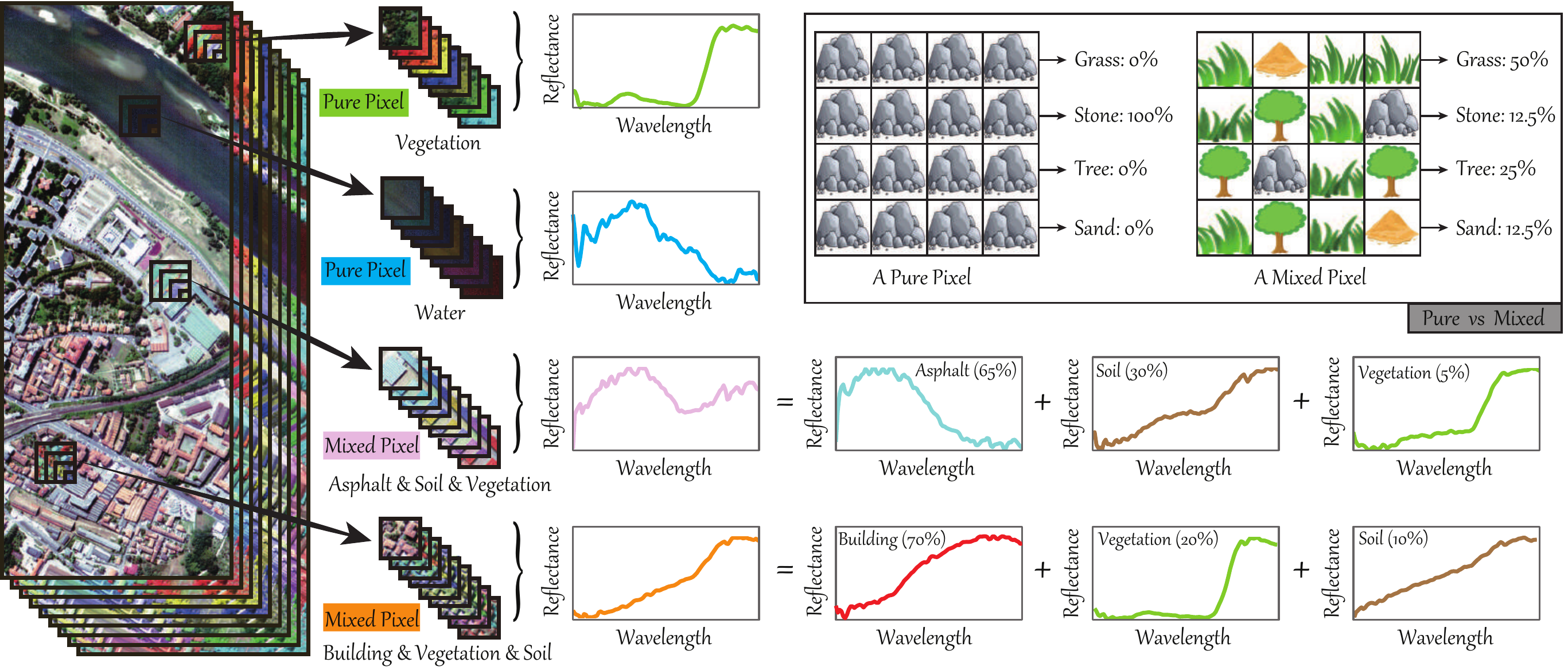}
        \caption{A showcase in a real HS scene (Pavia City Centre) to have a quick look at the 3-D HS cube, spectral signals, and material mixture as well as pure pixels (i.e., \textit{endmember}) and mixed pixels. In the studied scene, the pure pixels correspond to two spectral reflectance curves of \textit{vegetation} and \textit{water}, respectively, while the examples of mixed pixels explain the case of spectral mixing, e.g., the two mixed pixels comprise of three pure components (\textit{endmembers}) in varying proportions. In addition, the figure located in the right upper gives two toy examples to illustrate the material miscibility.}
\label{fig:MaterialMixture}
\end{figure*}

\section{Spectral Unmixing}

Spectral unmixing (SU) can be usually seen as a special case of blind source separation (BSS) problem in ML, referring to a procedure that decomposes the observed pixel spectrum of the HS image into a series of constituent spectral signals (or \textit{endmembers}) of pure materials and a set of corresponding abundance fractions (or \textit{abundance maps}) \cite{bioucas2012hyperspectral}. Due to the meter-level ground sampling distance (GSD) of HS imaging, the spectral signatures for most pixels in HS images are acquired in the form of a complex mixture that consists of at least two types of materials. Fig. \ref{fig:MaterialMixture} gives a showcase to visualize the HS cube, spectral signatures, and material mixing process as well as pure and mixed pixels. Different from general signals, e.g., digital signals, speech signals, there are specific absorption properties in the spectrum signals of different materials. Plus, HS images suffer from miscellaneous unknown degradation, either physically or chemically, in the remotely sensed imaging, inevitably bringing many uncertainties in SU. Therefore, SU plays a unique role in HS RS, yielding many challenging researchable tasks compared to BSS in ML. 

Ideally, a linear mixing model (LMM) can be used to accurately describe the SU process \cite{keshava2002spectral}, which is modeled as the following constrained optimization problem:
\begin{equation}
\label{SU_eq1}
\begin{aligned}
      \mathop{\min}_{\mathbf{E},\mathbf{A}}\frac{1}{2}\norm{\mathbf{Y}-\mathbf{E}\mathbf{A}}_{\F}^{2}\;\;\rm{s.t.}\; \mathbf{E},\mathbf{A}\in \mathcal{C}.
\end{aligned}
\end{equation}
The variables $\mathbf{E}$ and $\mathbf{A}$ in Eq. (\ref{SU_eq1}) stand for the endmembers and abundance maps in the SU issue, respectively. According to the endmembers ($\mathbf{E}$) that are available (given) or not in the process of SU, existing SU models can be loosely divided into \textit{blind SU} and \textit{endmember-guided SU}.

\subsection{Blind Spectral Unmixing}
NMF is a baseline model in a wide range of applications, and the same is true in SU. Up to the present, NMF-based interpretable models have been developed extensively for pursing the intelligent SU with the consideration of physically meaningful priors with respect to $\mathbf{E}$ and $\mathbf{A}$, e.g., the abundance non-negative constraint (ANC), the abundance sum-to-one constraint (ASC). The resulting basic blind SU model can be written as 
\begin{equation}
\label{SU_eq2}
\begin{aligned}
      \mathop{\min}_{\mathbf{E},\mathbf{A}}\frac{1}{2}\norm{\mathbf{Y}-\mathbf{E}\mathbf{A}}_{\F}^{2} + \Phi(\mathbf{E}) + \Omega(\mathbf{A})\;\;\rm{s.t.}\; \mathbf{E},\mathbf{A}\in \mathcal{C},
\end{aligned}
\end{equation}
where the constraint $\mathcal{C}$ is 
\begin{equation*}
\begin{aligned}
      \mathcal{C}:=\{\mathbf{E}\geq \mathbf{0},\; \mathbf{A}\geq \mathbf{0},\; \mathbf{1}^{\top}\mathbf{A}=\mathbf{1}\}.
\end{aligned}
\end{equation*}

On the basis of the model (\ref{SU_eq2}), Yang \textit{et al.} \cite{yang2010blind} proposed sparse NMF for SU with a well-designed S-measure sparseness. Qian \textit{et al.} \cite{qian2011hyperspectral} imposed the sparsity constraint on abundances and used $\ell_{1/2}$-regularized NMF for blind SU, which has shown to be more effective than $\ell_{0}$- and $\ell_{1}$-norm terms. In \cite{sigurdsson2014hyperspectral}, Sigurdsson \textit{et al.} relaxed $\ell_{1/2}$-norm to $\ell_{q}$-norm ($0\leq q\leq1$) for a better estimation of abundances. Thouvenin \textit{et al.} \cite{thouvenin2015hyperspectral} developed an improved LMM, called perturbed LMM (PLMM), by the attempts to model spectral variabilities as perturbed information that simply meets the Gaussian distribution. A similar work is presented in \cite{drumetz2016blind}, where the scaling factor, as a major spectral variability (SV), is modeled into LMM to yield an extended LMM (ELMM) for the blind SU task. He \textit{et al.} \cite{he2017total} employed total variation (TV) and  weighted $\ell_{1}$-norm terms to further enhance the smoothness and sparseness of abundances. Yao \textit{et al.} \cite{yao2019nonconvex} sought to explain the NMF-based SU model by simulating human observations on HS images, e.g., sparsity, non-local, smooth properties, in a non-convex modeling fashion. Another type of interesting SU strategy is to embed the graph or topological structure of the HS data. The local neighboring relation is introduced into the NMF model, showing robust SU results \cite{liu2012enhancing}. Similarly, Lu \textit{et al.} \cite{lu2012manifold} enforced the abundances to follow the manifold structure of spectral signatures in the form of Laplacian regularization form for HS unmixing. Wang \textit{et al.} \cite{wang2016hypergraph} used a structuralized hypergraph regularization in sparse NMF to better depict the underlying manifolds of the HS data. Very recently, Qin \textit{et al.} \cite{qin2020blind} proposed a novel graph TV regularization to estimate endmembers and abundances more effectively. There are still other variants that directly unmix the 3-D HS tensor by preserving spatial structure information as much as possible. For that, Qian \textit{et al.} \cite{qian2016matrix} proposed a matrix-vector non-negative tensor factorization framework for blind SU. Imbiriba \textit{et al.} \cite{imbiriba2019low} modeled the low-rank properties in the HS tensor to address the SV for robust SU. A further modified work based on \cite{imbiriba2019low} is proposed via weighted non-local low-rank tensor decomposition for sparse HS unmixing.

Broadly, these key non-convex priors of the above models can be briefly summarized as follows:
\begin{itemize}
    \item $\ell_{1/2}$-NMF \cite{qian2011hyperspectral}: $\Omega(\mathbf{A})=\norm{\mathbf{A}}_{1/2}=\sum_{k,n=1}^{K,N}\mathbf{a}_{n}(k)^{1/2}$;
    \item $\ell_{q}$-NMF \cite{sigurdsson2014hyperspectral}:
    $\Omega(\mathbf{A})=\norm{\mathbf{A}}_{q}=\sum_{k,n=1}^{K,N}\mathbf{a}_{n}(k)^{q}$;
    \item PLMM \cite{thouvenin2015hyperspectral}:
    $\Phi(\mathbf{E})=\frac{1}{2}\norm{\mathbf{E}-\mathbf{E}_{0}}_{\F}^{2}$,\;
    $\Omega(\mathbf{A})=\frac{1}{2}\norm{\mathbf{A}\mathbf{H}}_{\F}^{2}$,\; $\Psi(\Delta)=\frac{1}{2}\sum_{n=1}^{N}\norm{\Delta_{n}}_{\F}^{2}$,
    where $\mathbf{E}_{0}$, $\mathbf{H}$, and $\Delta$ denote the reference endmembers, the matrix differences in spatial four nearest neighbors, and pixel-wise perturbed information, respectively. 
    \item ELMM \cite{drumetz2016blind}: $\Phi(\mathbf{E})=\sum_{n=1}^{N}\norm{\mathbf{E}_{n}-\mathbf{E}_{0}\mathbf{S}_{n}}_{\F}^{2}$,\; $\Omega(\mathbf{A})=\norm{\mathbf{H}_{h}(\mathbf{A})}_{2,1}+\norm{\mathbf{H}_{v}(\mathbf{A})}_{2,1}$,\; $\Psi(\mathbf{S})=\norm{\mathbf{H}_{h}(\mathbf{S})}_{\F}^{2}+\norm{\mathbf{H}_{v}(\mathbf{S})}_{\F}^{2}$, where $\mathbf{H}_{h}$ and $\mathbf{H}_{v}$ are the horizontal and vertical gradients;
    \item TV-RSNMF \cite{he2017total}: $\Omega(\mathbf{A})=\norm{\mathbf{d}\odot\mathbf{A}}_{1,1}+\norm{\mathbf{A}}_{\rm TV}$;
    \item NLHTV \cite{yao2019nonconvex}:\\ $\Omega(\mathbf{A})=\sum_{n=1}^{N}\norm{J_{w}\mathbf{a}_{n}}_{\mathcal{S}_{1}}+\sum_{i,j}\log(|x_{i,j}|+\epsilon)$, where $J_{w}$ and $\norm{\bullet}_{\mathcal{S}_{1}}$ are defined as non-local Jacobian operator and the Schatten-1 norm, respectively.
    \item Graph $\ell_{1/2}$-NMF \cite{lu2012manifold}: $\Omega(\mathbf{A})=\norm{\mathbf{A}}_{1/2}+\tr(\mathbf{A}\mathbf{L}\mathbf{A}^{\top})$;
    \item Graph TV \cite{qin2020blind}: $\Omega(\mathbf{A})=\norm{\mathbf{A}}_{\rm TV}+\tr(\mathbf{A}\mathbf{L}\mathbf{A}^{\top})$.
\end{itemize}

Owing to the powerful data fitting ability, DL-based SU approaches have recently been paid increasing attention and achieved better unmixing results \cite{palsson2018hyperspectral,su2018stacked,palsson2019spectral,han2020deep}. Although these methods still suffer from the effects of ``black box'', i.e., the lack of model interpretability, yet their performances have preliminary shown the effectiveness and feasibility in unmixing the HS data more accurately.

\subsection{Endmember-Guided Spectral Unmixing}
A mass of blind SU methods has been developed and shown to be effective to simultaneously obtain endmembers and abundance maps. However, these blind methods tend to extract physical meaningless endmembers, e.g., noisy signals, spectral signatures corresponding to non-existent materials, due to the lack of certain interpretable model guidance or prior knowledge. A straightforward solution is to provide nearly real endmembers extracted from HS images. This naturally leads to the researches on endmember-guided SU. As the name suggests, the SU process is performed with given reference endmembers or the guidance of extracted endmembers from the HS image. That is, the endmembers $\mathbf{E}$ in Eq. (\ref{SU_eq2}) are known. Accordingly, the endmember-guided SU can be implemented in a three-stage way.
\begin{itemize}
    \item Firstly, the number of endmembers can be estimated by subspace estimation algorithms, e.g., HySime \cite{bioucas2008hyperspectral};
    \item Secondly, the endmembers can be extracted based on    geometric observations of HS data structure. Several well-known methods are vertex component analysis (VCA) \cite{nascimento2005vertex}, pixel purity index (PPI) \cite{chang2006fast}, and fast autonomous endmember extraction (N-FINDER) \cite{winter1999n}.
    \item Lastly, the abundances of materials are estimated using regression-based methods, which can generally written as 
    \begin{equation}
        \label{SU_eq3}
        \begin{aligned}
              \mathop{\min}_{\mathbf{A}}\frac{1}{2}\norm{\mathbf{Y}-\mathbf{E}\mathbf{A}}_{\F}^{2} + \Omega(\mathbf{A})\;\; {\rm s.t.}\; \mathbf{A}\in \mathcal{C}.
        \end{aligned}
    \end{equation}
\end{itemize}

Following the three steps, many well-working non-convex models have been successfully developed to estimate the abundance maps of different materials at a more accurate level. Heinz \textit{et al.} \cite{heinz2001fully} thoroughly analyzed the spectral mixture in the SU issue, yielding a fully constrained least-squares unmixing (FCLSU) algorithm. Due to the hard ASC, the abundances can not be fully represented in a simplex. For this reason, a partial constraint least-squares unmixing (PCLSU) \cite{heylen2011fully} model emerges as required without ASC. Bioucas-Dias \textit{et al.} \cite{bioucas2010alternating} relaxed the strong $\ell_{0}$-norm to the solvable $\ell_{1}$-norm in the sparse HS unmixing model and designed a fast and generic optimization algorithm based on the ADMM framework \cite{boyd2011distributed}, called sparse unmixing by variable splitting and augmented Lagrangian (SUnSAL). In \cite{iordache2012total}, a TV spatial regularization is considered to further enhance the unmixing performance. Iordache \textit{et al.} \cite{iordache2013collaborative} extended the sparse regression model to the collaborative version regularized by $\ell_{2,1}$-norm for SU. Fu \textit{et al.} \cite{fu2016semiblind} proposed a semi-blind HS unmixing model by correcting the mismatches between estimated endmembers and pure spectral signatures from the library. Huang \textit{et al.} \cite{huang2018joint} jointly imposed sparsity and low-rank properties on the abundances for better estimating abundance maps. Hong \textit{et al.} \cite{hong2018sulora} devised an interesting and effective subspace-based abundance estimation model. The model neatly sidesteps to directly decompose the HS data in the complex high-dimensional space instead of projecting the HS data into a more robust subspace, where the SV tends to be removed in a more generalized way with low-rank attribute embedding. Beyond the current framework, Hong \textit{et al.} \cite{hong2019augmented} further augmented the basic LMM by fully modeling SVs, e.g., principal scaling factors and other SVs that should be incoherent or low-coherent with endmembers, in order to yield an interpretable and more intelligent SU model, called augmented LMM (ALMM). 

\begin{figure}[!t]
	  \centering
			\includegraphics[width=0.47\textwidth]{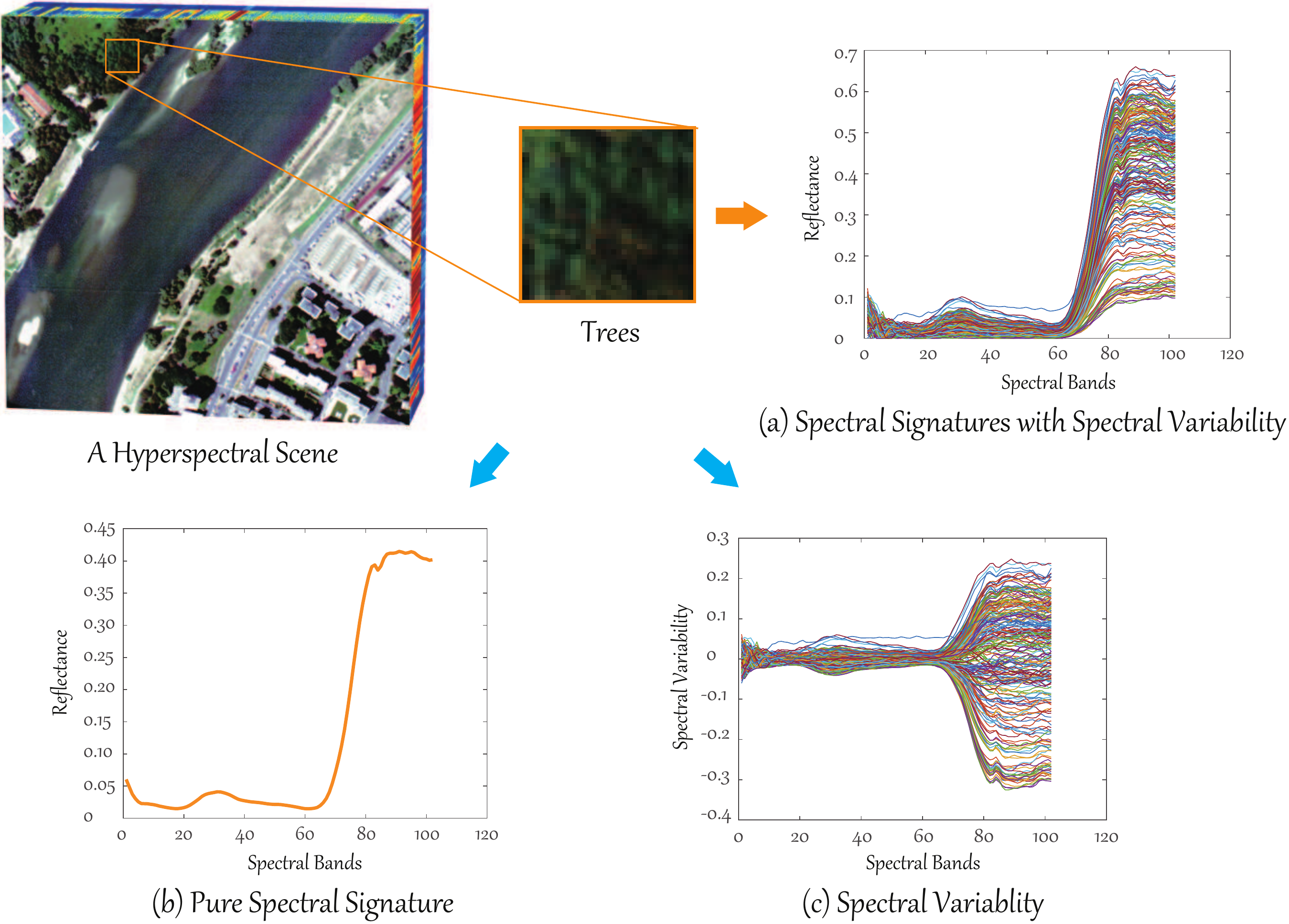}
        \caption{A visual example to clarify SVs in a real HS scene (Pavia City Centre). An image patch cropped from the scene is select to show the spectral bundles involving spectral variations of trees in (a). (b) shows a pure spectral signature (i.e., \textit{endmember}) of trees acquired from the laboratory. (c) represents the differences between (a) and (b), which is seen as SVs.}
\label{fig:SV}
\end{figure}

\begin{figure*}[!t]
	  \centering
			\includegraphics[width=1\textwidth]{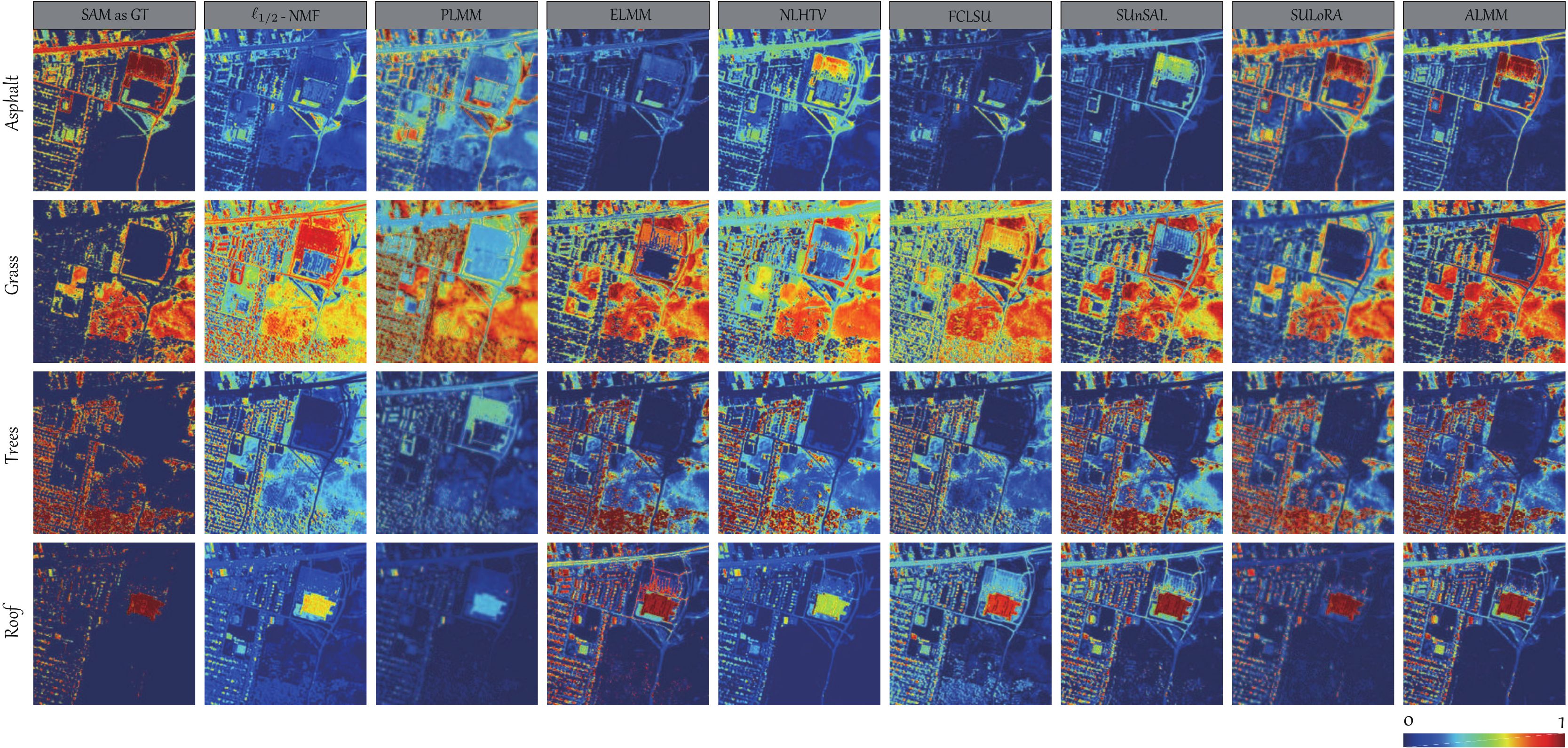}
        \caption{Visualization of abundance maps estimated by different SOTA SU algorithms on the Urban data, where SAM is computed to generate the classification-like maps regarded as the GT to measure the shape similarity of abundance maps obtained by different SU methods.}
\label{fig:SU_AB}
\end{figure*}

The non-convexity of these methods on priors, constraints, or modeling can be summarized as follows:
\begin{itemize}
    \item FCLSU \cite{heinz2001fully}: $\mathbf{A}\geq \mathbf{0}$, $\mathbf{1}^{\top}\mathbf{A}=\mathbf{1}$;
    \item PCLSU \cite{heylen2011fully}: $\mathbf{A}\geq \mathbf{0}$;
    \item SUnSAL \cite{bioucas2010alternating}: $\Omega(\mathbf{A})=\norm{\mathbf{A}}_{1,1}$, $\mathbf{A}\geq \mathbf{0}$, $\mathbf{1}^{\top}\mathbf{A}=\mathbf{1}$;
    \item SUnSAL-TV \cite{iordache2012total}: $\Omega(\mathbf{A})=\norm{\mathbf{A}}_{1,1}+\norm{\mathbf{A}}_{\rm TV}$, $\mathbf{A}\geq \mathbf{0}$;
    \item CSR \cite{iordache2013collaborative}: $\Omega(\mathbf{A})=\norm{\mathbf{a}}_{2,1}=\sum_{n=1}^{N}\norm{\mathbf{a}_{n}}_{2}$, $\mathbf{A}\geq \mathbf{0}$;
    \item DANSER \cite{fu2016semiblind}: $\Omega(\mathbf{A})=\sum_{n=1}^{N}(\norm{\mathbf{a}_{n}}_{2}^{2}+\tau)^{p/2}$, $\mathbf{A}\geq \mathbf{0}$, $\Phi(\mathbf{E})=\norm{\mathbf{E}-\mathbf{E}_{0}}_{\F}^{2}$;
    \item SULoRA \cite{hong2018sulora}: $\Psi(\mathbf{U})=\norm{\mathbf{Y}-\mathbf{U}\mathbf{Y}}_{\F}^{2}+\norm{\mathbf{U}}_{*}$, $\Omega(\mathbf{A})=\norm{\mathbf{A}}_{1,1}$, $\mathbf{A}\geq \mathbf{0}$, where $\mathbf{U}$ denotes the subspace projection and $\norm{\bullet}_{*}$ is the nuclear norm that approximates the rank property of the matrix $\bullet$;
    \item ALMM \cite{hong2019augmented}: $\Phi(\mathbf{A})=\norm{\mathbf{A}}_{1,1}$, $\mathbf{A}\geq \mathbf{0}$, $\Gamma(\mathbf{J})=\norm{\mathbf{J}}_{\F}^{2}$, $\Psi(\mathbf{V})=\norm{\mathbf{A}^{\top}\mathbf{V}}_{\F}^{2}+\norm{\mathbf{V}^{\top}\mathbf{V}-\mathbf{I}}_{\F}^{2}$, where $\mathbf{V}$ and $\mathbf{J}$ denote the SV dictionary and corresponding coefficients, respectively.
\end{itemize}

\subsection{Experimental Study}
A real urban HS data acquired by the HYDICE over the urban area, Texas, USA, in 2015 (the latest version\footnote{\url{http://www.tec.army.mil/Hypercube}}) is used to evaluate the performance of several selected SOTA unmixing methods qualitatively, including $\ell_{1/2}$-NMF \cite{qian2011hyperspectral}, PLMM\footnote{\url{https://pthouvenin.github.io/unmixing-plmm/}} \cite{thouvenin2015hyperspectral}, ELMM\footnote{\url{https://openremotesensing.net/knowledgebase/spectral-variability-and-extended-linear-mixing-model/}} \cite{drumetz2016blind}, NLHTV \cite{yao2019nonconvex}, FCLSU \cite{heinz2001fully}, SUnSAL\footnote{\url{http://www.lx.it.pt/~bioucas/}} \cite{bioucas2010alternating}, SULoRA\footnote{\url{https://github.com/danfenghong/IEEE_JSTSP_SULoRA}} \cite{hong2018sulora}, and ALMM\footnote{\url{https://github.com/danfenghong/ALMM_TIP}} \cite{hong2019augmented}. The HS image consists of $307\times 307$ pixels and 162 spectral bands after removing noisy bands in the wavelength range of $0.4 \mu m$ to $2.5 \mu m$ at a $2m$ GSD. Moreover, four main materials (or endmembers) are investigated in the studied scene, i.e., \textit{asphalt}, \textit{grass}, \textit{trees}, and \textit{roof}. Furthermore, HySime \cite{bioucas2008hyperspectral} and VCA \cite{nascimento2005vertex} algorithms are adopted to determine the number of endmembers and extract endmembers from the HS image (as the initialization for blind SU methods) for all compared algorithms, respectively.

Fig.~\ref{fig:SU_AB} shows the visual comparison between different SOTA unmixing algorithms in terms of abundance maps. Owing to the consideration of real endmembers extracted from the HS scene, the last four endmember-guided SU methods perform evidently better than the blind SU ones. ELMM models the scaling factors, tending to better capture the distributions of different materials. The embedding of non-local spatial information makes the NLHTV method obtain a more similar shape of abundance maps to the GT, yielding comparable unmixing performance with ELMM. Remarkably, the unmixing results with regard to abundance maps of SULoRA and ALMM algorithms are superior to those of other methods, since the SVs can be fully considered by robustly embedding low-rank attributes in a latent subspace using SULoRA and characterizing complex real scenes more finely using ALMM.

\subsection{Remaining Challenges}
SU has long been a challenging and widely concerned topic in HS RS. Over the past decades, tons of SU works have been proposed by the attempts to unmix these mixed spectral pixels more effectively. Yet, some key and essential issues and challenges still remain to be solved.

\begin{itemize}
    \item \textbf{Benchmark Data.} Unlike classification, recognition, and detection tasks, the ground truth of material abundances is able to be hardly collected, due to the immeasurability of abundance values in reality. On the other hand, spectral signatures (i.e., endmembers) of pure materials are often acquired in the lab. This usually leads to uncertain mismatches between real endmembers and lab ones. It turns to be urgent to establish benchmark datasets for SU by drawing support from more advanced imaging techniques or developing interpretable ground truth generation models or processing chain.
    \item \textbf{Evaluation Criteria.} Reconstruction errors (RE) or spectral angle mapper (SAM) are the two commonly used evaluation indices in SU. It should be noted, however, that the results of RE or SAM are not equivalent to those of unmixing. Linking to the issue of benchmark data, the measurement between real results and estimated ones is the optimal choice, if we have the ground truth for abundances and endmembers. If not, developing meaningful and reasonable evaluation indices (e.g., classification accuracy) should give the top priority in future work.
    \item \textbf{Spectral Variability.} Spectral signatures inevitably suffer from various SVs caused by illumination and topography change, noise effects from external conditions and internal equipment, atmospheric interference, and complex mixing of materials in the process of imaging. Fig. \ref{fig:SV} shows a visual example to specify the SVs (e.g., trees) in a real HS scene. Considerable uncertainties brought by these factors have a big negative impact on accurate estimation of abundances and endmembers in SU.
    \item \textbf{Nonlinearity.} The complex interactions (e.g., intimate mixing, multilayered mixing \cite{bioucas2012hyperspectral}) between multiple materials, also known as nonlinearity, inevitably occur in the process of HS imaging. The nonlinearity in SU is a longstanding and pending challenge. Most of existing nonlinear unmixing models only attempt to consider certain special cases \cite{heylen2014review}, e.g., bilinear mixing, intimate mixtures, etc. Consequently, there is still lack of a general and powerful model that can robustly address various nonlinearities in SU.
    \item \textbf{Model Explainability.} The non-negativity and the sum-to-one constraint considered in LMM are the basic priors for spectral signals in HS images. However, only the two constraints fail to model the complex unmixing process in an explainable fashion. To further enhance the explainability, new spectral mixture models should be developed beyond the classic LMM by fully excavating the intrinsic attribute knowledge that lies in the HS image. 
\end{itemize}

\section{Data Fusion and Enhancement}
\begin{figure}[!t]
	  \centering
		\includegraphics[width=0.4\textwidth]{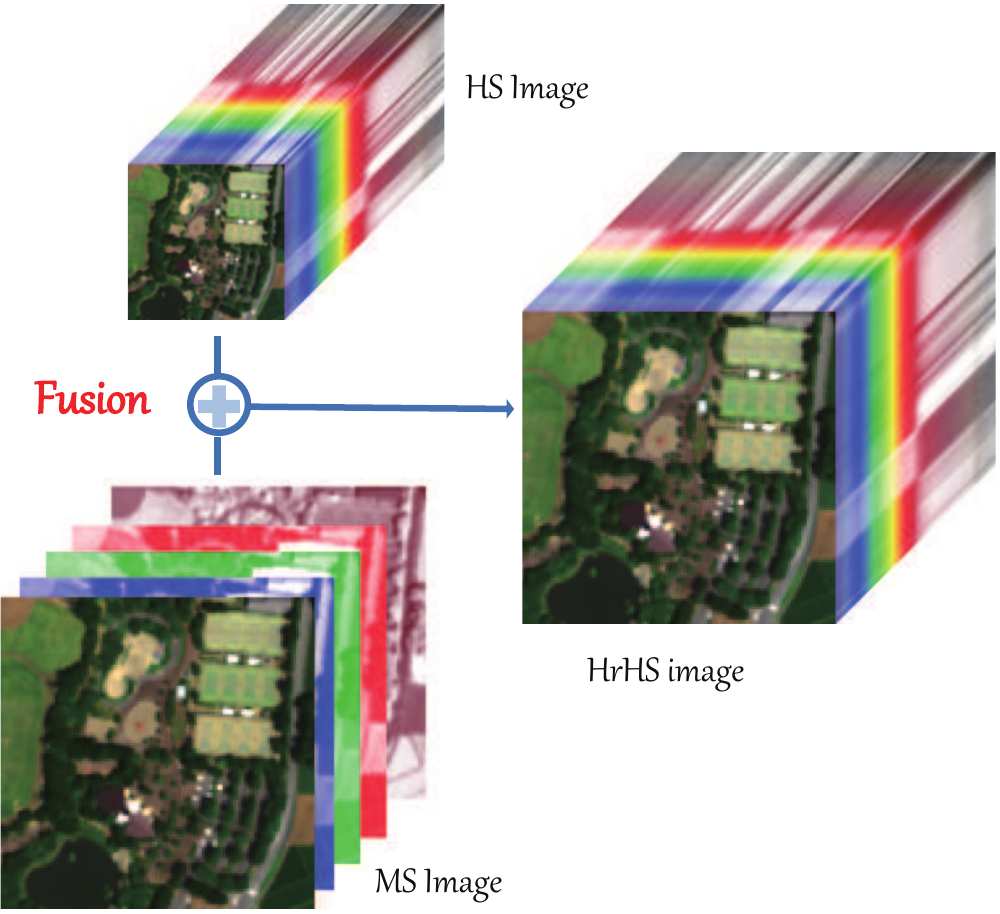}
        \caption{Illustration of MS-HS image fusion to generate the HrHS image.}
\label{fig:datafusion}
\end{figure}

The high spectral resolution of HS images enables the identification and discrimination of materials, meanwhile the high spatial resolution can provide the possibility of the derivation of surface parameters~\cite{yokoya2017hyperspectral}. However, due to the equipment limitation, there is usually a trade-off between the spatial and spectral resolutions, and the HS images obtained by the spaceborne imaging spectrometers are usually with a moderate ground sampling distance~\cite{yokoya2017hyperspectral}. To enhance the spatial resolution, one popular way is to fuse the HS images with high spatial MS images to generate new high spatial-spectral HS (HrHS) images. In particular, enormous effects have been recently made to enhance the spatial or spectral resolutions of HS images by means of ML techniques. Fig.~\ref{fig:datafusion} illustrates the fusion process of HS-MS images to generate the HrHS image. 
Suppose we have the low-spatial resolution HS image $\tensor{Y} \in  \mathbb{R}^{m \times n \times B}$ and high-spatial resolution MS image $\tensor{Z}\in  \mathbb{R}^{M \times N \times b}$ with $M \gg m$, $N \gg n$ and $B \gg b$, the fusion purpose is to generate the high-spatial resolution HS image $\tensor{X}\in \mathbb{R}^{M \times N \times B}$. The degradation models from $\tensor{X}$ to $\tensor{Y}$ and $\tensor{Z}$ are formulated as
\begin{align}
\mat{Y} = \mat{X}\mat{R}+\mat{N}_{H}
\\
\mat{Z} = \mat{G}\mat{X}+\mat{N}_{M}
\label{eq:Fusi}
\end{align}
where $\mat{X}, \mat{Y}, \mat{Z}$ are the reshaped matrices along the spectral dimension of $\tensor{X}, \tensor{Y}, \tensor{Z}$, respectively, $\mat{R}$ is the mixed cyclic convolution and downsampling operator, $\mat{G}$ is the spectral response function (SRF) of the MS image sensor, $\mat{N}_{H}$ and $\mat{N}_{M}$ are the corresponding MS-HS noise. To unify different observation models~\cite{Eismann2004,yokoya2012coupled,QiWei2015TGRS,Simoes2015,yokoya2017hyperspectral,he2020hyperspectral}, $\mat{N}_{H}$ and $\mat{N}_{M}$ are assumed to be the independent identically distributed Gaussian noise. Via the maximum a posteriori (MAP) estimation method and Bayes rule~\cite{Eismann2004,QiWei2015TGRS,Simoes2015}, the following non-convex optimization model is obtained
\begin{align}
&\min_{\mat{X}} \|\mat{Y} - \mat{X}\mat{R}\|_{\F}^2+\|\mat{Z} - \mat{G}\mat{X}\|_{\F}^2,
\label{eq:FusiOpt}
\end{align}
where $\mat{R}$ and $\mat{G}$ are assumed to be known (in~\cite{QiWei2015TGRS,yokoya2017hyperspectral}, $\mat{R}$ and $\mat{G}$ are estimated in advance of the optimization). and As mentioned in~\cite{QiWei2015TGRS,Simoes2015}, the optimization of \eqref{eq:FusiOpt} is a NP-hard problem, and over-estimation of $\mat{Z}$ will result in the unstable fusion results. Therein, additional property of $\mat{X}$ and prior regularizers should be exploited in the optimization model \eqref{eq:FusiOpt}. It should be noted, however, that the two functions $\mathbf{R}$ and $\mathbf{G}$ can be given according to known sensors and also can be learned or automatically estimated from the data itself.

HS pansharpening is a heuristic way to perform the HS-MS fusion \cite{loncan2015hyperspectral}, which has been widely applied in the HS image enhancement task. Component substitution (CS) and multiresolution analysis (MRA) are the two main types of pansharpening techniques. The former one aims to inject detailed information of MS images into the low-resolution HS image, thereby generating the high-resolution HS product. The latter one is to pansharpen the HS image by linearly combining MS bands to synthesize a high-resolution HS band using regression techniques. Another group for the HS-MS fusion task is the subspace-based model, which roughly consists of Bayesian and unmixing based methods (see \cite{yokoya2017hyperspectral}). Different from pansharpening, the subspace-based approaches project the to-be-fused MS and HS images to a new space where the dimension is generally smaller than that of the unknown high-resolution HS image, by the means of the probability-driven Bayesian estimation (Bayesian-based methods) or SU-guided matrix joint factorization (unmixing-based methods).

In the following, we focus on the subspace methods, and review the related HS-MS image fusion methods from the non-convex modeling perspective. A more detailed review can be referred to~\cite{yokoya2017hyperspectral,loncan2015hyperspectral}.

\begin{figure*}[!t]
	  \centering
			\includegraphics[width=1\textwidth]{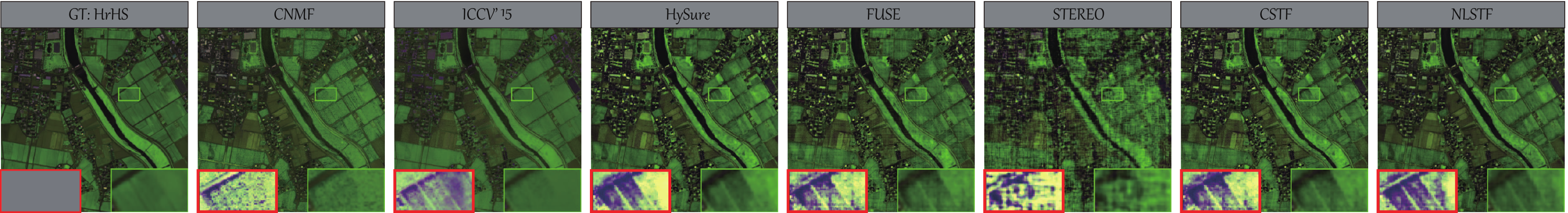}
        \caption{The fusion results of different methods with Chikusei image. The color is composed of bands 70, 100, 36. An enlarged region is framed in {\bf green} and the corresponding residual image between the fused image and MS-GT is framed in {\bf red}.}
\label{fig:Chikusei_reconstruct}
\end{figure*}

\subsection{Unmixing based methods}
Hyperspectral unmixing (HU)~\cite{bioucas2013hyperspectral,he2017total} assumes that the mixed class of an HS image can be decomposed to the collection of constitute spectra (endmembers), and their corresponding proportions (abundances). With LMM assumption, the different endmembers do not interfere with each other~\cite{bioucas2013hyperspectral}. By embedding the LMM model into \eqref{eq:FusiOpt}, we can obtain the following general unmixing based approaches
\begin{align}
\label{eq:FusiOptMix}
&\min_{\mat{X},\mat{E},\mat{A}} \|\mat{Y} - \mat{X}\mat{R}\|_{\F}^2+\|\mat{Z} - \mat{G}\mat{X}\|_{\F}^2,\\
& {\rm s.t.} \;\; \mat{X} = \mat{E}\mat{A}, \; \mat{E},\mat{A} \geq 0,\; \mat{1}_{R}^{\top}\mat{A}=\mat{1}_{MN},
\notag
\end{align}
where $\mat{E}$, $\mat{A}$ are the endmember matrix and abundance matrix, which are assumed to obey the non-negative and abundance sum-to-one constraints. Generally, nonlinear unmixing models~\cite{bioucas2013hyperspectral} can be also utilized for the fusion task of HS-MS images. However, due to the generality of the LMM model, we focus on the review of LMM based fusion approaches.

Eismann \textit{et al.} proposed a maximum a posteriori estimation method to deduce the cost objective function, and introduced a stochastic mixing model (MAP-SMM) to embed LMM into the cost function~\cite{Eismann2004}. MAP-SMM method tries to estimate the prior probabilities for all the mixture classes, including the mean vectors and covariance matrices of the endmember classes. The learned prior probabilities are passed to the cost function to help the reconstruction of the final HrHS image $\mat{X}$.

Yokoya \textit{et al.,} regarded \eqref{eq:FusiOptMix} as the coupled NMF (CNMF) problem~\cite{yokoya2012coupled} and introduced the multiplicative update rules to optimize \eqref{eq:FusiOptMix}. Firstly, CNMF utilizes $\|\mat{Y} - \mat{E}\mat{A}\mat{R}\|_{\F}^2$ as the cost function to update $\mat{E}$ and $\mat{A}_h$ with the endmember matrix $\mat{E}$ initialized by vertex component analysis (VCA). Here, $\mat{A}_h = \mat{A} \mat{R}$ is the abundance matrix from HS images. Secondly, by initializing $\mat{E}_m = \mat{G}\mat{E}$ which is the endmember matrix from the MS image, CNMF again utilizes the multiplicative update rules to update $\mat{E}$ from the cost function $\|\mat{Z} - \mat{G}\mat{E}\mat{A}\|_{\F}^2$. Finally, the HrHS image $\mat{X}$ is reconstructed from $\mat{E}\mat{A}$. The following works~\cite{akhtar2014sparse,akhtar2015bayesian,dong2016hyperspectral} also utilize the CNMF framework to fuse HS-MS images. Differently,~\cite{akhtar2014sparse,dong2016hyperspectral} introduced a non-negative dictionary learning strategy, while \cite{akhtar2015bayesian} proposed the proximal alternating linearized minimisation algorithm to update $\mat{E}$ and $\mat{A}$.

On the basis of \eqref{eq:FusiOptMix}, Wang \textit{et al.,} further regularized $\mat{X}$ with non-local low-rank Tucker decomposition~\cite{Wang2020TGRS}. The improved non-local Tucker decomposition regularized CNMF model~\cite{Wang2020TGRS} was solved by the multi-block ADMM, and achieved remarkable fusion results. It indicates that additional regularizers on $\mat{X}$ can further improve the fusion accuracy. From another side, it is necessary to make a trade-off between the complex models with higher accuracy and the computation efficiency for real large scale HS-MS image fusion task.

\subsection{Orthogonal subspace based methods}
Another common assumption in HS-MS fusion is that the spectral information of $\mat{X}$ underlies a orthogonal subspace, whose dimension is much smaller than the number of bands $B$~\cite{QiWei2015TGRS,he2017total}, $\textit{i.e.,}$ $\mat{X} = \mat{E}\mat{A}$ with $\mat{E}\in  \mathbb{R}^{B \times k}$, $\mat{A}\in  \mathbb{R}^{k \times MN}$, and $k\ll B$.  $\mat{E}^{\top}$ is an orthogonal matrix with $\mat{E}^{\top}\mat{E} = \mat{I}_{k}$. Therefore, the subspace based model is formulated as
\begin{align}
\label{eq:FusiOpt2}
&\min_{\mat{X},\mat{E},\mat{A}} \|\mat{Y} - \mat{X}\mat{R}\|_{\F}^2+\|\mat{Z} - \mat{G}\mat{X}\|_{\F}^2, \\
& {\rm s.t.}\;\; \mat{X} = \mat{E}\mat{A}, \; \mat{E}^{\top}\mat{E} = \mat{I}_{k}.
\notag
\end{align}
Although additional spectral subspace prior is exploited, the optimization of \eqref{eq:FusiOpt2} still faces several challenges. Firstly, if $k\gg b$, that's to say, the dimension number of the subspace is larger than the bands' number of the MS image, the optimization of \eqref{eq:FusiOpt2} is the under-estimate problem. Therefore, to ensure a reasonable solution, prior information of coefficient $\mat{A}$ need to be exploited.~\cite{QiWei2015TGRS} pre-trains a dictionary to represent $\mat{A}$, and updates $\mat{A}$ via sparse representation. Hyperspectral super-resolution (HySure) in~\cite{Simoes2015} assumes that $\mat{A}$ appears the spatial smoothness structure and regularize $\mat{A}$ with band-by-band TV. \cite{wei2015fast} translates the optimization of $\mat{A}$ to a Sylvester equation and proposes a fast fusion method for  \eqref{eq:FusiOpt2} (FUSE).

Secondly, the optimization of orthogonal matrix $\mat{E}^{\top}$ is another challenge due to the non-convex of \eqref{eq:FusiOpt2}. One appearing approach~\cite{QiWei2015TGRS,Simoes2015,wei2015fast} is to pre-estimate $\mat{E}$ from $\mat{Y}$ in advance, and fix the variable $\mat{E}$ during the optimization of \eqref{eq:FusiOpt2}. Specially, FUSE~\cite{wei2015fast} adopted principal component analysis (PCA), meanwhile, HySure utilized VCA to extract $\mat{E}$ from $\mat{Y}$. Another strategy is to regularize the update of $\mat{E}$ and $\mat{A}$ as a coupled matrix factorization problem, and blind dictionary learning strategy is utilized to update $\mat{E}$~\cite{kawakami2011high}. A hybrid inexact block coordinate descent~\cite{wu2020hybrid} is introduced to exactly estimate $\mat{E}$.

\subsection{Tensor based methods}
The above subspace based methods utilize low-rank matrix decomposition to exploit the low-rank property of the reshaped high-spatial resolution HS image $\mat{X}$. However, the original HS image is a 3-D tensor, and therein, the researchers introduce the tensor decomposition to simultaneously capture the spatial and spectral low-rank property. The coupled sparse tensor factorization (CSTF) approach~\cite{li2018fusing} utilized Tucker decomposition, presented as follows:
\begin{align}
\label{eq:TucD}
&\tensor{X} = \tensor{O}\times_1 \mat{E}_1\times_2 \mat{E}_2\times_3 \mat{E}_3, \\
&{\rm subject \;to} \;\; \mat{E}_i^{\top}\mat{E}_i = \mat{I},\; \|\tensor{O}\|_0 \leq \mathcal{C},
\notag
\end{align}
to regularize the high-spatial resolution HS image $\tensor{X}$. In \eqref{eq:TucD}, the core tensor $\tensor{O}$ is assumed to obey the sparse property, and $\mat{E}_i$ is the orthogonal matrix of the $i$-th dimension. Subsequently, CP decomposition~\cite{kanatsoulis2018hyperspectral}, tensor train decomposition~\cite{dian2019learning}, tensor ring decomposition~\cite{xu2020hyperspectral,he2020hyperspectral}, and so on, are utilized to regularize $\tensor{X}$. Furthermore, non-local LRTD is also investigated for the fusion task~\cite{dian2017hyperspectral,wang2017hyperspectral,xu2019nonlocal}.

It is worth noting that unmixing, orthogonal subspace, and tensor based methods share the common idea that spectral space of $\mat{X}$ should lie in the low-dimensional space. Unmixing based approaches interpret the low-rank property as the endmembers and abundances, which are assumed to be non-negative, meanwhile Orthogonal subspace and tensor based methods ignore the non-negative restrict. Unmixing based approaches are interpretable from the physical meaning, but suffering from the unstable convergence in the optimization. Orthogonal subspace and tensor based methods lose physical meaning, but can be optimized more elegantly. 

Very recently, there are some preliminary works to perform the fusion task by means of DL-based methods \cite{mei2017hyperspectral,haut2018new,liu2019stfnet,liu2019efficient,zheng2020coupled,uezato2020guided,yao2020cross} and show the effective and competitive fusion performance. A similar problem existed in these methods is the model interpretability and rationality. Clearly explaining the intrinsic meaning in each layer of deep networks would contribute to better modeling the fusion task and further obtaining higher-quality products.

\begin{table}[!t]
\centering
\caption{Quantitative comparison of different algorithms on the HS-MS image fusion experiments. The best one is shown in bold.}
\resizebox{0.4\textwidth}{!}{
\begin{tabular}{c||cccc}
\toprule[1.5pt] Methods & RMSE & ERGAS & SA & SSIM \\
\hline \hline
CNMF \cite{yokoya2012coupled} & 6.404 & 0.715 & 4.89 & 0.8857\\
ICCV'15 \cite{lanaras2015hyperspectral} & \bf 5.203 & \bf 0.589 & \bf 4.64 & \bf 0.9139\\
HySure \cite{Simoes2015} & 8.537 & 0.812 & 9.45 & 0.8527\\
FUSE \cite{wei2015fast} & 8.652 & 0.869 & 9.51 & 0.8401\\
CSTF \cite{li2018fusing} & 8.32 & 0.841 & 8.34 & 0.8419\\
STEREO \cite{kanatsoulis2018hyperspectral} & 9.4425 & 0.891 & 9.78  & 0.8231\\
NLSTF \cite{dian2017hyperspectral} & 8.254 & 0.819 & 8.36 & 0.8424\\
\bottomrule[1.5pt]
\end{tabular}}
\label{tab:fusion}
\end{table}

\begin{figure*}[!t]
	  \centering
		\subfigure[Training for MML and CML]{
			\includegraphics[width=0.315\textwidth]{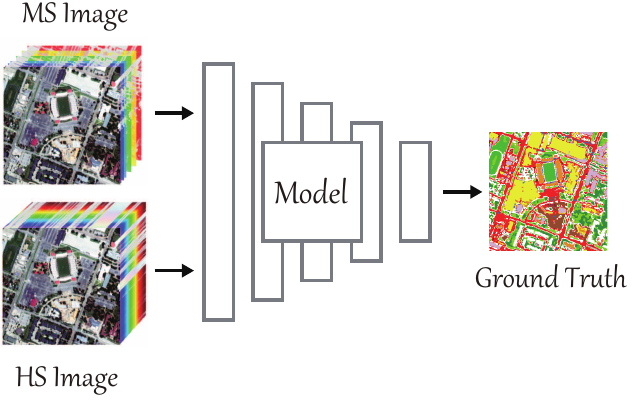}
		}
		\subfigure[Testing for MML]{
			\includegraphics[width=0.315\textwidth]{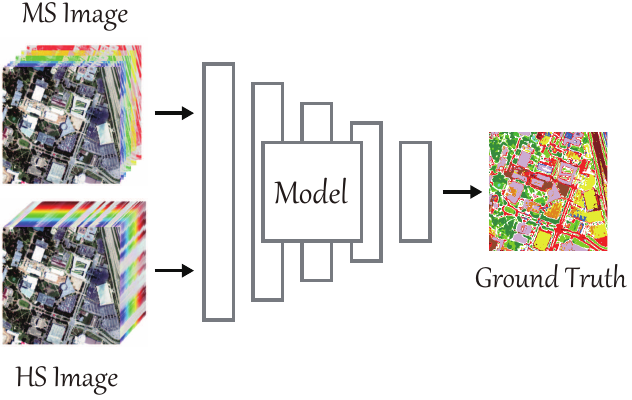}
		}
		\subfigure[Testing for CML]{
			\includegraphics[width=0.315\textwidth]{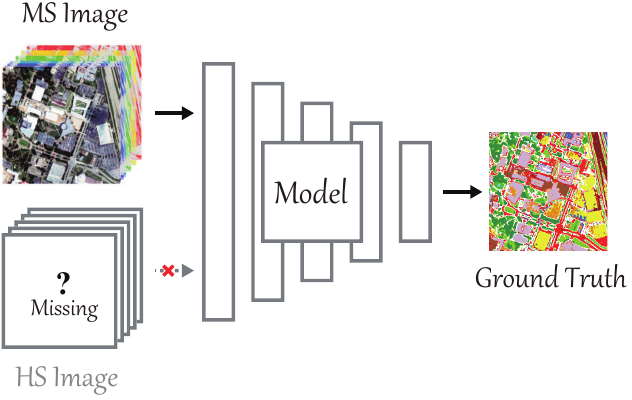}
		}
		\caption{An illustration for the model's training and testing in MML- and CML-based classification tasks (take the bi-modality as an example). (a) They share the same training process, i.e., two modalities are used for model training. The main difference lies in the testing phase, (b) MML still needs the input of two modalities, (c) while one modality is absent in CML.}
\label{fig:CML}
\end{figure*}

\subsection{Experimental Study}
In this section, we select  unmixing based methods: CNMF\footnote{\url{ http://naotoyokoya.com/Download.html}}~\cite{yokoya2012coupled}, ICCV'15\footnote{\url{https://github.com/lanha/SupResPALM}} \cite{lanaras2015hyperspectral}; subspace based methods: HySure\footnote{\url{ https://github.com/alfaiate}}~\cite{Simoes2015}, FUSE\footnote{\url{ https://github.com/qw245/BlindFuse}}~\cite{wei2015fast}; tensor decomposition regularized methods: STEREO\footnote{\url{ https://sites.google.com/site/harikanats/}}~\cite{kanatsoulis2018hyperspectral}, CSTF~\cite{li2018fusing}; finally non-local tensor decomposition regularized method: NLSTF\footnote{\url{ https://sites.google.com/view/renweidian/}}~\cite{dian2017hyperspectral} for the comparison and analysis. We use evaluation indices, including the root mean square error (RMSE), relative dimensional global error in synthesis (ERGAS) \cite{wald2000quality}, MSA, and SSIM \cite{wang2004image} as evaluation criteria for the fusion results of different methods.

The selected dataset for the experiment is the Chikusei dataset obtained at Chikusei, Ibaraki, Japan, on 29 July 2014~\cite{yokoya2017hyperspectral}. The selected high-spatial resolution HS image is of size $448 \times 448 \times 128$, and the simulated HS-MS images are of size $448 \times 448 \times 3$ and $14 \times 14 \times 128$, respectively. Tab.~\eqref{tab:fusion} presents the quantitative comparison results of different algorithms on the HS-MS image fusion, meanwhile Fig.~\eqref{fig:Chikusei_reconstruct} presents the visual illustration. From the results, it can be observed that even the HS image is spatially degraded by 32 times, the fusion methods can efficiently reconstruct the spatial details with the help of a 3 band MS image. On this tested toy dataset, ICCV'15 performed the best. However, different datasets need different kinds of regularizers. The fusion of HS-MS images for efficient and large scale applications is still a challenge for further research. 

\subsection{Remaining Challenges}
Subspace based non-convex methods for the fusion of HS-MS images have been well developed. However, most remarkable results are achieved on the simulated experiments. For the real applications with HS-MS images from two different satellite sensors, there still remain several challenges.

\begin{itemize}
    \item \textbf{Blind.} Most fusion methods assume the linear spatial and spectral downsampling from HR-HS image to HS-MS images. However, in real applications, the degradation is complex and unknown in advance. how to blindly reconstruct the HR-HS image is a challenge in future research. 
    \item \textbf{Regularizer.} We reviewed the subspace based fusion methods from unmixing, orthogonal subspace, and tensor decomposition perspectives. Different assumptions are suitable for the exploited for the different structure of the HS image. How to mine the essence of HS images and develop efficient regularizers for large scale processing still remains a challenge.
    \item \textbf{Evaluation.} In the real cases, the enhanced HrHS images from HS and MS images are not existed as the reference images in the real scenario. How to evaluate the final enhanced HrHS images is also a key problem for the future fusion approach development of HS-MS. 
\end{itemize}

\section{Cross-modality Learning for Large-scale Land Cover Mapping}
With the ever-growing availability of diverse RS data sources from both satellite and airborne sensors, multimodal data processing and analysis in RS \cite{dalla2015challenges,hong2020more} can provide potential possibilities to break the performance bottleneck in many high-level applications, e.g., land cover classification.

HS data are featured by rich spectral information, which enables the high discrimination ability for material recognition at a more accurate and fine level. It should be noted, however, that the HS image coverage from space is much narrow compared to MS imaging due to the limitations of imaging principles and devices. That means the HS-dominated multimodal learning (MML) fails to identify the materials on a large geographic coverage and even global scale \cite{hong2020x}. But fortunately, large-scale MS or synthetic aperture radar (SAR) images are openly available from e.g., Sentinel-1, Sentinel-2, Landsat-8. This, therefore, drives us to ponder over a problem: \textit{can HS images acquired only in a limited area improve the land cover mapping performance using a larger area covered by the MS or SAR images?} This is a typical issue of cross-modality learning (CML) from a ML's point of view.

Take the bi-modality as an example, CML for simplicity refers to that training a model using two modalities and one modality is absent in the testing phase, or \emph{vice versa} (only one modality is available for training and bi-modality for testing) \cite{ngiam2011multimodal}. Such a CML problem that exists widely in a variety of RS tasks is more applicable to real-world cases. Fig. \ref{fig:CML} illustrates the differences between MML and CML in terms of training and testing process. The core idea of CML is to find a new data space, where the information can be exchanged effectively across different modalities. Thereupon, we formulate this process in a general way as follows: 
\begin{equation}
\label{CML_eq1}
\begin{aligned}
      \mathop{\min}_{\mathbf{X},\{\mathbf{U}_{s}\}_{s=1}^{m}}\sum_{s=1}^{m}\frac{1}{2}\norm{\mathbf{X}-\mathbf{U}_{s}\mathbf{Y}_{s}}_{\F}^{2} \;\; {\rm s.t.}\; \mathbf{X},\{\mathbf{U}_{s}\}_{s=1}^{m}\in \mathcal{C},
\end{aligned}
\end{equation}
where $m$ is the number of input modality. For simplicity, we only consider the bi-modality case in this topic, i.e., $m=2$. According to different learning strategies on modalities, CML can be roughly categorized into two groups: manifold alignment (MA) and shared subspace learning (SSL). The differences between the two types of approaches mainly lie in 
\begin{itemize}
    \item MA learns the low-dimensional embedding by preserving the aligned manifold (or graph) structure between different modalities. In the process of graph construction, the similarities between samples (unsupervised MA) and indirect label information (supervised or semi-supervised MA) are used. Despite the competitive performance obtained by MA-based approaches for the CML task, the discrimination ability of learned features remains limited due to the lack of directly bridging low-dimensional features with label information.
    \item SSL, as the name suggests, aims to find a latent shared subspace, where the features of different modalities are linked via a manifold alignment regularizer. Also, the learned features are further connected with label information. The two steps are jointly optimized in a SSL model, tending to yield more discriminative feature representations.
\end{itemize}
More specifically, we will briefly review and detail some representative approaches belonging to the aforementioned two groups as follows.

\subsection{Manifold Alignment based Approach}
As the name suggests, MA is capable of aligning multiple modalities on manifolds into a latent subspace, achieving a highly effective knowledge transfer \cite{wang2009general}. Due to the interactive learning ability, MA has a good fit for large-scale RS image classification. In \cite{matasci2011transfer}, the domain adaptation was investigated to reduce the gap between the source and target domains of HS data for land cover classification. By simultaneously considering labeled and unlabeled samples, Tuia \textit{et al.} \cite{tuia2014semisupervised} used semi-supervised MA (SSMA) techniques \cite{wang2011heterogeneous} to align the multi-view RS images onto the manifold space by the attempts to eliminate the effects of image variants caused by different views. Matasci \textit{et al.} \cite{matasci2015semisupervised} modified the classic transfer component analysis \cite{pan2011domain}, making it applicable to land cover classification of RS images. Moreover, a kernelized MA approach presented in \cite{tuia2016kernel} projected the multimodal RS data to a higher dimensional space and aligned them in a nonlinear way. Hu \textit{et al.} \cite{hu2019comparative} deeply reviewed the semi-supervised MA methods with respect to the fusion classification of HS and polarimetric SAR images. Based on the work in \cite{hu2019comparative}, the same investigators made full use of topological data analysis and designed a new graph structure for optical (e.g., HS) and SAR data fusion \cite{hu2019mima}. 

Mathematically, the MA idea can be implemented by solving the following non-convex model:
\begin{equation}
\label{CML_eq2}
\begin{aligned}
      \mathop{\min}_{\{\mathbf{U}\}_{s=1}^{m}}\frac{A+C}{B},
\end{aligned}
\end{equation}
where $A$, $B$, and $C$ are 
\begin{equation*}
\begin{aligned}
      &A = \frac{1}{2}\sum_{p=1}^{m}\sum_{q=1}^{m}\sum_{i=1}^{n}\sum_{j=1}^{n}\norm{\mathbf{U}_{p}\mathbf{y}_{p}^{i}-\mathbf{U}_{q}\mathbf{y}_{q}^{j}}_{2}^{2}\mathbf{W}_{sim}^{i,j},\\
      &B = \frac{1}{2}\sum\limits_{p=1}^{m}\sum\limits_{q=1}^{m}\sum\limits_{i=1}^{n}\sum\limits_{j=1}^{n}\norm{\mathbf{U}_{p}\mathbf{y}_{p}^{i}-\mathbf{U}_{q}\mathbf{y}_{q}^{j}}_{2}^{2}\mathbf{W}_{dis}^{i,j},\\
      &C = \frac{1}{2}\sum_{t=1}^{m}\sum_{i=1}^{n}\sum_{j=1}^{n}\norm{\mathbf{U}_{t}\mathbf{y}_{t}^{i}-\mathbf{U}_{t}\mathbf{y}_{t}^{j}}_{2}^{2}\mathbf{W}_{t}^{i,j}.
\end{aligned}
\end{equation*}
By minimizing the problem (\ref{CML_eq2}), the $\{\mathbf{U}\}_{s=1}^{m}$ can be estimated via generalized eigenvalues decomposition. We then have $\mathbf{X}=\mathbf{U}_{s}\mathbf{Y}_{s}$. Three different graphs need to be pre-computed in Eq. (\ref{CML_eq2}), including the similarity graph, i.e., $\mathbf{W}_{sim}$:
\begin{equation}
\label{CML_eq3}
\begin{aligned}
	  \mathbf{W}_{sim}=
	      \left[
        	 \begin{matrix}
        	  \mathbf{W}_{sim}^{1,1}&\mathbf{W}_{sim}^{1,2}&\cdots&  \mathbf{W}_{sim}^{1,m}\\
        	  \mathbf{W}_{sim}^{2,1}&\mathbf{W}_{sim}^{2,2}&\cdots&  \mathbf{W}_{sim}^{2,m}\\
        	  \vdots&\vdots&\ddots&\vdots\\
        	  \mathbf{W}_{sim}^{m,1}&\mathbf{W}_{sim}^{m,2}&\cdots&  \mathbf{W}_{sim}^{m,m}\\
             \end{matrix}
         \right],
\end{aligned}
\end{equation}
the dissimilarity matrix, i.e., $\mathbf{W}_{dis}$:
\begin{equation}
\label{CML_eq4}
\begin{aligned}
	  \mathbf{W}_{dis}=
	      \left[
        	 \begin{matrix}
        	  \mathbf{W}_{dis}^{1,1}&\mathbf{W}_{dis}^{1,2}&\cdots&  \mathbf{W}_{dis}^{1,m}\\
        	  \mathbf{W}_{dis}^{2,1}&\mathbf{W}_{dis}^{2,2}&\cdots&  \mathbf{W}_{dis}^{2,m}\\
        	  \vdots&\vdots&\ddots&\vdots\\
        	  \mathbf{W}_{dis}^{m,1}&\mathbf{W}_{dis}^{m,2}&\cdots&  \mathbf{W}_{dis}^{m,m}\\
             \end{matrix}
         \right],
\end{aligned}
\end{equation}
and the topology structure for each single modality obtained by $knn$ graph, i.e., $\mathbf{W}_{t}$:
\begin{equation}
\label{CML_eq5}
\begin{aligned}
	  \mathbf{W}_{t}=
	      \left[
        	 \begin{matrix}
        	  \mathbf{W}_{t}^{1,1}&\mathbf{0}&\cdots&  \mathbf{0}\\
        	  \mathbf{0}&\mathbf{W}_{d}^{2,2}&\cdots&  \mathbf{0}\\
        	  \vdots&\vdots&\ddots&\vdots\\
        	  \mathbf{0}&\mathbf{0}&\cdots&  \mathbf{W}_{t}^{m,m}\\
             \end{matrix}
         \right].
\end{aligned}
\end{equation}
In Eqs. (\ref{CML_eq3})$-$(\ref{CML_eq5}), $\mathbf{W}_{sim}^{i,j}$, $\mathbf{W}_{dis}^{i,j}$, and $\mathbf{W}_{t}^{i,j}$ are given, respectively, by
\begin{equation*}
    \mathbf{W}_{sim}^{i,j}=
    \begin{cases}
      \begin{aligned}
          1, \; \; & \text{if $\mathbf{y}_{p}^{i}$ and $\mathbf{y}_{q}^{j} \in C_{k}$}\\
          0, \; \; & \text{otherwise,}
      \end{aligned}
    \end{cases}
\end{equation*} 
\begin{equation*}
    \mathbf{W}_{dis}^{i,j}=
    \begin{cases}
      \begin{aligned}
          1, \; \; & \text{if $\mathbf{y}_{p}^{i}$ and $\mathbf{y}_{q}^{j}\notin C_{k}$ }\\
          0, \; \; & \text{otherwise,}
      \end{aligned}
    \end{cases}
\end{equation*} 
\begin{equation*}
    \mathbf{W}_{t}^{i,j}=
    \begin{cases}
      \begin{aligned}
      \exp\frac{\norm{\mathbf{y}^{i}-\mathbf{y}^{j}}_{2}^{2}}{2\sigma^{2}}, \; \; & \text{if $\mathbf{y}_{p}^{i}\in\phi_{k}(\mathbf{y}_{q}^{j})$;}\\
      0, \; \; & \text{otherwise,}
      \end{aligned}
    \end{cases}
\end{equation*} 
where $\phi_{k}(\bullet)$ denotes the $k$ nearest neighbors of $\bullet$.

\subsection{Shared Subspace Learning based Approach}
Due to the lack of the direct relational modeling between the learned features and label information, MA-based approaches fail to activate the connections across modalities effectively \cite{ngiam2011multimodal}, thereby yielding the relatively weak transferability between different modalities, particularly heterogeneous data. There have been some tentative works in recent years, providing potential solutions to overcome the aforementioned challenges. For example, Hong \textit{et al.} \cite{hong2019cospace} for the first time proposed a supervised CoSpace model to learn a latent discriminative subspace from HS-MS correspondences for the CML-related classification problem. Beyond it, the same authors \cite{hong2019learnable} fully tapped the potential of the CoSpace by learning the data-driven graph structure from both labeled and unlabeled samples, yielding a learnable manifold alignment (LeMA) approach. Moreover, \cite{hong2020learning} deeply investigated and analyzed different regression techniques, i.e., $\ell_2$-norm ridge regression, $\ell_1$-norm sparse regression, in CoSpace. In \cite{hong2020graph}, a semi-supervised graph-induced aligned learning (GiAL) was developed by jointly regressing labels and pseudo-labels.

Accordingly, these methods can be generalized to be a unified model \cite{hong2019cospace} to address the CML's problem in a regression-based fashion:
\begin{equation}
\label{CML_eq6}
\begin{aligned}
      \mathop{\min}_{\mathbf{P}, \{\mathbf{U}_{s}\}_{s=1}^{m}}&\frac{1}{2}\norm{\mathbf{M}-\mathbf{P}\mathbf{U}_{s}\mathbf{Y}_{s}}_{\F}^{2}+\Psi(\mathbf{P})+\Omega(\{\mathbf{U}_{s}\}_{s=1}^{m})\\
      & {\rm s.t.}\;\; \mathbf{U}_{s}\mathbf{U}_{s}^{\top}=\mathbf{I}, \;s=1,\cdots,m,
\end{aligned}
\end{equation}
where $\{\mathbf{U}_{s}\}_{s=1}^{m}$ denote the projections linking to the shared features for different modalities. To avoid the over-fitting of the model and stabilize the learning process, $\mathbf{P}$ can be regularized by the Frobenius-norm \cite{hong2019cospace} or $\ell_{1,1}$-norm \cite{hong2020learning}:
\begin{equation}
\label{CML_eq7}
\begin{aligned}
      \Psi(\mathbf{P})=\norm{\mathbf{P}}_{\F}^{2}, \; \text{or}\; \norm{\mathbf{P}}_{1,1},
\end{aligned}
\end{equation}
and $\Omega(\{\mathbf{U}_{s}\}_{s=1}^{m})$ is specified as a manifold alignment term on the multimodal data, which is written as
\begin{equation}
\label{CML_eq8}
\begin{aligned}
      \Omega(\{\mathbf{U}_{s}\}_{s=1}^{m})=\tr(\mathbf{U}\mathbf{Y}\mathbf{L}\mathbf{Y}^{\top}\mathbf{U}^{\top}),
\end{aligned}
\end{equation}
where $\mathbf{U}=[\mathbf{U}_{1},\mathbf{U}_{2},\cdots,\mathbf{U}_{m}]$ and
\begin{equation*}
\begin{aligned}
   \mathbf{Y}=
	      \left[
        	 \begin{matrix}
        	  \mathbf{Y}_{1} & \mathbf{0} & \cdots &  \mathbf{0}\\
        	  \mathbf{0} & \mathbf{Y}_{2} & \cdots &  \mathbf{0}\\
        	  \vdots&\vdots&\ddots&\vdots\\
        	  \mathbf{0} & \mathbf{0} & \cdots &  \mathbf{Y}_{m}\\
             \end{matrix}
         \right].    
\end{aligned}
\end{equation*}
Similar to Fig. \ref{fig:graph}, $\mathbf{L}$ is a joint Laplacian matrix. 

Using the general model in Eq. (\ref{CML_eq6}), 
\begin{itemize}
    \item \cite{hong2019cospace} considers the HS-MS correspondences that exist in an overlapped region as the model input. The learned shared representations (e.g., $\mathbf{X}=\mathbf{U}_{s}\mathbf{Y}_{s}$) can be then used for classification on a larger area, even though only MS data are available in the inference phase;
    \item Differently, \cite{hong2019learnable} inputs not only the labeled HS-MS pairs but also unlabeled MS data in large quantity. With the graph learning, i.e., the variable $\mathbf{L}$ is to be learned from the data rather than fixed by a given RBF, the unlabeled information can be made use of to find a better decision boundary. According to the equivalent form of Eq. (\ref{CML_eq8}), we then have
    \begin{equation}
        \label{CML_eq9}
        \begin{aligned}
             \tr(\mathbf{U}\mathbf{Y}\mathbf{L}\mathbf{Y}^{\top}\mathbf{U}^{\top})=\frac{1}{2}\tr(\mathbf{W}\mathbf{d})=\frac{1}{2}\norm{\mathbf{W}\odot\mathbf{d}}_{1,1},
        \end{aligned}
    \end{equation}
    where $\mathbf{d}_{i,j}=\norm{\mathbf{x}_{i}-\mathbf{x}_{j}}_{2}^{2}$ denotes the pair-wise distance in Euclidean space. Using Eq. (\ref{CML_eq9}), the resulting optimization problem with respect to the variable $\mathbf{W}$ is 
    \begin{equation}
        \label{CML_eq10}
        \begin{aligned}
              \frac{1}{2}&\norm{\mathbf{W}\odot\mathbf{d}}_{1,1}\\
              &{\rm s.t.} \; \mathbf{W}=\mathbf{W}^{\top},\; \mathbf{W}_{i,j}\geq 0,\; \norm{\mathbf{W}}_{1,1}=c.
        \end{aligned}
    \end{equation}
    \item Inspired by the brain-like feedback mechanism presented in \cite{hong2019learning}, a more intelligent CML model was proposed  \cite{hong2020graph}. With the joint use of labels and pseudo-labels updated by the graph feedback in each iteration, more representative features can be also learned (even if a certain modality is absent, i.e., the CML case). 
\end{itemize}

\begin{table}[!t]
\centering
\caption{Quantitative comparison of SOTA algorithms related to the CML's issue in terms of OA, AA, and $\kappa$ using the NN classifier on the Houston2013 datasets. The best one is shown in bold.}
\resizebox{0.35\textwidth}{!}{
\begin{tabular}{c||ccc}
\toprule[1.5pt] Methods & OA (\%) & AA (\%) & $\kappa$ \\
\hline \hline
O-Baseline & 62.12 & 65.97 & 0.5889\\
USMA \cite{he2004locality} & 65.54 & 68.81 & 0.6251\\
SMA \cite{wang2009general} & 68.01 & 70.50 & 0.6520\\
SSMA \cite{wang2011heterogeneous} & 69.29 & 72.00 & 0.6659\\
CoSpace \cite{hong2019cospace} & 69.38 & 71.69 & 0.6672\\
LeMA \cite{hong2019learnable} & 73.42 & 74.76 & 0.7110\\
GiAL \cite{hong2020graph} & \bf 80.66 & \bf 81.31 & \bf 0.7896\\
\bottomrule[1.5pt]
\end{tabular}}
\label{tab:CML}
\end{table}

\subsection{Experimental Study}
We evaluate the performance of several SOTA algorithms related to the CML's issue both quantitatively and qualitatively. They are O-Baseline (i.e., using original image features), unsupervised MA (USMA) \cite{he2004locality}, supervised MA (SMA)\footnote{\url{https://sites.google.com/site/changwangnk/home/ma-html}} \cite{wang2009general}, SSMA \cite{wang2011heterogeneous}, CoSpace\footnote{\url{https://github.com/danfenghong/IEEE_TGRS_CoSpace}} \cite{hong2019cospace}, LeMA\footnote{\url{https://github.com/danfenghong/ISPRS_LeMA}} \cite{hong2019learnable}, and GiAL \cite{hong2020graph}. Three common indices, e.g., \textit{OA}, \textit{AA}, and $\kappa$, are adopted to quantify the classification performance using the SVM classifier on the Houston2013 HS-MS datasets that have been widely used in many researches \cite{hong2019cospace,hong2019learnable,hong2020learning,hong2020graph}. 

\begin{figure}[!t]
	  \centering
		\subfigure{
			\includegraphics[width=0.42\textwidth]{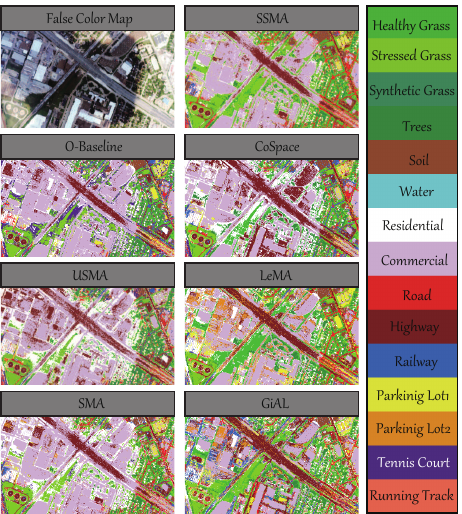}
		}
        \caption{ROI visualization of classification maps using different SOTA methods related to the CML's issue.}
\label{fig:CML_CM_H}
\end{figure}

Table \ref{tab:CML} gives the quantitative comparison between the above-mentioned methods for the CML-related classification, while Fig. \ref{fig:CML_CM_H} visualizes a region of interest (ROI) of classification maps. By and large, the classification accuracy of O-Baseline, i.e., only using MS data, is much lower than other methods. By aligning multimodal data on manifolds, MA-based approaches perform better than O-Baseline with the approximated increase of $3\%$ OA in USMA, $6\%$ OA in SMA, and $7\%$ OA in SSMA. As expected, the classification performance of SSL-based models, e.g., CoSpace, LeMA, and GiAL, is obviously superior to that of MA-based ones. In particular, GiAL dramatically outperforms other competitors, owing to the use of the brain-like feedback mechanism and graph-driven pseudo-label learning. Visually, shared learning methods tend to capture more robust spectral properties and achieve more realistic classification results. As can be seen from Fig. \ref{fig:CML_CM_H}, the shadow region covered by clouds can be finely classified by CoSpace, LeMA, and GiAL, while MA-based models fail to identify the materials well in the region.

\subsection{Remaining Challenges}
CML has drawn growing interest from researchers in computer vision and ML, yet it is rarely investigated in the RS community. In other words, CML is a relatively emerging topic in RS, which means there are lots of difficulties (or challenges) to be overcome. In detail,

\begin{itemize}
    \item \textbf{Data Preparation.} Since multimodal data are acquired by different contexts, sensors, resolutions, etc., this inevitably poses a great challenge to data collection and processing. For example, the errors caused by interpolation of different resolutions, registration methods of geographical coordinates, pixel-wise biases of different sensors, and uncertainties of image degradation in the imaging process easily generate unregistered multimodal data to a great extent.
    \item \textbf{Model Transferability.} Due to different imaging mechanisms and principles, the viscosity between pixels from the same modality is much stronger than from the different modalities. This might lead to difficulties in fusing multimodal information at a deep level, particularly heterogeneous data (e.g., HS and SAR data), further limiting the model's transferability.
    \item \textbf{Labeling.} Unlike natural images or street view images that are relatively easy and accurate to be labeled manually, labeling RS scenes (field trips needed) is extremely expensive and time-consuming. Consequently, a limited number of labeled samples are available for training and even worse, there are many noisy labels in these samples. These problems will be noteworthy and to-be-solved key points of the next generation interpretable AI models in the RS-related CML task.
\end{itemize}

\section{Conclusion and Future Prospect}
Characterized by the nearly continuous spectral profile that is capable of sampling and representing the whole electromagnetic spectrum, HS images play an important role in both promoting developments of new techniques and accelerating the practical applications, not only limiting to the fields of RS, geoscience, signal and image processing, numerical optimization and modeling, ML, and AI. However, there still exist severe difficulties and challenges that need to be carefully considered in the development and application of HS RS techniques. One sign reveals that HS data analysis methods dominated by expert systems have been unable to meet the demand of an ever-growing volume of HS data whether in performance gain or in processing efficiency. Another sign is that despite the currently unprecedented progress made on computer vision, ML, and AI techniques, the model compatibility and interpretability for HS RS applications remain limited.

Due to the SVs of HS data caused by various degradation mechanisms (e.g., environmental condition, atmospheric effects, spectral nonlinear mixing, etc.), the redundancy of high-dimensional HS signals, and the complex practical cases underlying in the HS products (e.g., low spatial resolution, narrow imaging range, instrumental noises), convex models under ideal circumstances usually fail to extract useful and diagnostic information from HS images (especially those products that are corrupted seriously) and thereby understand our environment. Considering that non-convex modeling is capable of characterizing more complex real scenes and better providing the model interpretability, in this article we present a comprehensive and technical survey over five promising and representative research topics related to HS RS with a focus on non-convex modeling, such as HS image restoration, dimensionality reduction and classification, data fusion and enhancement, spectral unmixing, and cross-modality learning. Among these topics, we review the current state-of-the-art methods with illustrations, show the significance and superiority of non-convex modeling to bridge the gap between HS RS and interpretable AI, and point out remaining challenges and future research directions.

It is well-known that the HS image processing and analysis chain is wide-ranging. Apart from the five topics covered in this paper, we are not able to detailedly report all the important and promising applications related to HS RS missions. There are several very noteworthy and active fields that should be paid more attention in future work, including target/change detection, time-series analysis, multitemporal fusion/classification, physical parameter inversion, image quality assessment, and various practical applications (e.g., precious farming, disaster management and response). Moreover, some crucial steps for the HS image pre-processing algorithms are also missing, such as atmospheric and geometric corrections, geographic coordinate registration, etc. Furthermore, the methodologies summarized and reported in this article mainly focus on the survey of shallow non-convex models. Undeniably, deep models, e.g., DL-based methods, are capable of excavating deeper and intrinsic properties of HS data. There is, therefore, room for improvement in the development of more intelligent DL-related non-convex modeling with application to HS RS. For example, embedding more physically meaningful priors and devising advanced and novel deep unfolding \cite{hershey2014deep} or unrolling \cite{diamond2017unrolled} strategies to closely integrate data-driven DL and theoretically-guaranteed optimization technique is to open and interpret the so-called ``black box'' in DL models.

Finally, we have to admit that non-convex modeling and optimization is a powerful tool across multidisciplinary and the relevant studies along the direction have been made tremendous progress theoretically and technically. This provides the possibility of creating new methodologies and implementing interpretable AI for various HS RS applications. In this paper, we attempt to ``intellectualize'' these models by introducing more interpretable and physically meaningful knowledge to meet the actual needs in a non-convex modeling fashion. In other words, we hope that non-convex modeling can play the role as a bridge to connect interpretable AI models and various research topics in HS RS. Our efforts in this paper are made to foster curiosity and create a good starting point for post-graduate, Ph.D. students, and senior researchers working in the HS-related fields, thereby further looking for new and advanced research directions in the interdisciplinary involving signal and image processing, ML, AI, and RS.

\section*{Acknowledgement}
The authors would like to thank Prof. D. Landgrebe from
Purdue University for providing the AVIRIS Indian Pines data, Prof. P. Gamba from the University of Pavia for providing the ROSIS-3 Pavia University and Centre data, the Hyperspectral Image Analysis group at the University of Houston for providing the CASI University of Houston dataset used in the IEEE GRSS DFC2013 and DFC2018, and the Hyperspectral Digital Imagery Collection Experiment (HYDICE) for sharing the urban dataset free of charge. 

This work from the D. Hong and X. Zhu sides is jointly supported by the German Research Foundation (DFG) under grant ZH 498/7-2, by the Helmholtz Association through the framework of Helmholtz Artificial Intelligence (HAICU) - Local Unit ``Munich Unit @Aeronautics, Space and Transport (MASTr)'' and Helmholtz Excellent Professorship ``Data Science in Earth Observation - Big Data Fusion for Urban Research'', by the German Federal Ministry of Education and Research (BMBF) in the framework of the international AI future lab ``AI4EO -- Artificial Intelligence for Earth Observation: Reasoning, Uncertainties, Ethics and Beyond''. This work from the L. Gao side is supported by the National Natural Science Foundation of China under Grant 42030111 and Grant 41722108. This work from the W. He and N. Yokoya sides is supported by the Japan Society for the Promotion of Science under KAKENHI 19K20308 and KAKENHI 18K18067. This work from the J. Chanussot side has been partially supported by MIAI@Grenoble Alpes, (ANR-19-P3IA-0003) and by the AXA Research Fund. 

The corresponding authors of this paper are Dr. Wei He and Prof. Lianru Gao.

\bibliographystyle{ieeetr}
\bibliography{HDF_ref}

\begin{thebibliography}{100}

\bibitem{goetz1985imaging}
A.~Goetz, G.~Vane, J.~Solomon, and B.~Rock, ``Imaging spectrometry for earth
  remote sensing,'' {\em Science}, vol.~228, no.~4704, pp.~1147--1153, 1985.

\bibitem{tsang1985theory}
L.~Tsang, J.~A. Kong, and R.~T. Shin, ``Theory of microwave remote sensing,''
  1985.

\bibitem{turner2003remote}
W.~Turner, S.~Spector, N.~Gardiner, M.~Fladeland, E.~Sterling, and
  M.~Steininger, ``Remote sensing for biodiversity science and conservation,''
  {\em Trends Ecol. Evol.}, vol.~18, no.~6, pp.~306--314, 2003.

\bibitem{hong2015novel}
D.~Hong, W.~Liu, J.~Su, Z.~Pan, and G.~Wang, ``A novel hierarchical approach
  for multispectral palmprint recognition,'' {\em Neurocomputing}, vol.~151,
  pp.~511--521, 2015.

\bibitem{bioucas2013hyperspectral}
J.~Bioucas-Dias, A.~Plaza, G.~Camps-Valls, P.~Scheunders, N.~Nasrabadi, and
  J.~Chanussot, ``Hyperspectral remote sensing data analysis and future
  challenges,'' {\em IEEE Geosci. Remote Sens. Mag.}, vol.~1, no.~2, pp.~6--36,
  2013.

\bibitem{boyd2004convex}
S.~Boyd, S.~P. Boyd, and L.~Vandenberghe, {\em Convex optimization}.
\newblock Cambridge university press, 2004.

\bibitem{ekeland1979nonconvex}
I.~Ekeland, ``Nonconvex minimization problems,'' {\em Bull. Amer. Math. Soc.},
  vol.~1, no.~3, pp.~443--474, 1979.

\bibitem{mohri2018foundations}
M.~Mohri, A.~Rostamizadeh, and A.~Talwalkar, {\em Foundations of machine
  learning}.
\newblock MIT press, 2018.

\bibitem{nilsson2014principles}
N.~J. Nilsson, {\em Principles of artificial intelligence}.
\newblock Morgan Kaufmann, 2014.

\bibitem{chvatal1983linear}
V.~Chvatal, V.~Chvatal, {\em et~al.}, {\em Linear programming}.
\newblock Macmillan, 1983.

\bibitem{frank1956algorithm}
M.~Frank, P.~Wolfe, {\em et~al.}, ``An algorithm for quadratic programming,''
  {\em Nav. Res. Logist. Q.}, vol.~3, no.~1-2, pp.~95--110, 1956.

\bibitem{lobo1998applications}
M.~S. Lobo, L.~Vandenberghe, S.~Boyd, and H.~Lebret, ``Applications of
  second-order cone programming,'' {\em Linear Algebra Appl.}, vol.~284,
  no.~1-3, pp.~193--228, 1998.

\bibitem{jain2017non}
P.~Jain, P.~Kar, {\em et~al.}, ``Non-convex optimization for machine
  learning,'' {\em Found. Trends{\textregistered} Mach. Learn.}, vol.~10,
  no.~3-4, pp.~142--363, 2017.

\bibitem{boyd2011distributed}
S.~Boyd, N.~Parikh, and E.~Chu, {\em Distributed optimization and statistical
  learning via the alternating direction method of multipliers}.
\newblock Now Publishers Inc, 2011.

\bibitem{martin2007anisotropic}
J.~Martin-Herrero, ``Anisotropic diffusion in the hypercube,'' {\em IEEE Trans.
  Geosci. Remote Sens.}, vol.~45, no.~5, pp.~1386--1398, 2007.

\bibitem{datt2003preprocessing}
B.~Datt, T.~R. McVicar, T.~G. Van~Niel, D.~L. Jupp, and J.~S. Pearlman,
  ``Preprocessing eo-1 hyperion hyperspectral data to support the application
  of agricultural indexes,'' {\em IEEE Trans. Geosci. Remote Sens.}, vol.~41,
  no.~6, pp.~1246--1259, 2003.

\bibitem{gomez2005cloud}
L.~G{\'o}mez-Chova, J.~Amor{\'o}s, G.~Camps-Valls, J.~D. Martin, J.~Calpe,
  L.~Alonso, L.~Guanter, J.~C. Fortea, and J.~Moreno, ``Cloud detection for
  chris/proba hyperspectral images,'' in {\em Remote Sensing of Clouds and the
  Atmosphere X}, vol.~5979, p.~59791Q, International Society for Optics and
  Photonics, 2005.

\bibitem{acito2011signal}
N.~Acito, M.~Diani, and G.~Corsini, ``Signal-dependent noise modeling and model
  parameter estimation in hyperspectral images,'' {\em IEEE Trans. Geosci.
  Remote Sens.}, vol.~49, no.~8, pp.~2957--2971, 2011.

\bibitem{Chang1999}
C.-I. Chang and Q.~Du, ``Interference and noise-adjusted principal components
  analysis,'' {\em IEEE Trans. Geosci. Remote Sens.}, vol.~37, pp.~2387--2396,
  Sep. 1999.

\bibitem{yuan2012hyperspectral}
Q.~Yuan, L.~Zhang, and H.~Shen, ``Hyperspectral image denoising employing a
  spectral--spatial adaptive total variation model,'' {\em IEEE Trans. Geosci.
  Remote Sens.}, vol.~50, no.~10, pp.~3660--3677, 2012.

\bibitem{zhang2013hyperspectral}
H.~Zhang, W.~He, L.~Zhang, H.~Shen, and Q.~Yuan, ``Hyperspectral image
  restoration using low-rank matrix recovery,'' {\em IEEE Trans. Geosci. Remote
  Sens.}, vol.~52, no.~8, pp.~4729--4743, 2013.

\bibitem{he2015total}
W.~He, H.~Zhang, L.~Zhang, and H.~Shen, ``Total-variation-regularized low-rank
  matrix factorization for hyperspectral image restoration,'' {\em IEEE Trans.
  Geosci. Remote Sens.}, vol.~54, no.~1, pp.~178--188, 2015.

\bibitem{green1988transformation}
A.~A. Green, M.~Berman, P.~Switzer, and M.~D. Craig, ``A transformation for
  ordering multispectral data in terms of image quality with implications for
  noise removal,'' {\em IEEE Trans. Geosci. Remote Sens.}, vol.~26, no.~1,
  pp.~65--74, 1988.

\bibitem{rasti2014wavelet}
B.~Rasti, J.~R. Sveinsson, and M.~O. Ulfarsson, ``Wavelet-based sparse
  reduced-rank regression for hyperspectral image restoration,'' {\em IEEE
  Trans. Geosci. Remote Sens.}, vol.~52, no.~10, pp.~6688--6698, 2014.

\bibitem{qian2012hyperspectral}
Y.~Qian and M.~Ye, ``Hyperspectral imagery restoration using nonlocal
  spectral-spatial structured sparse representation with noise estimation,''
  {\em IEEE J. Sel. Top. Appl. Earth Obs. Remote Sens.}, vol.~6, no.~2,
  pp.~499--515, 2012.

\bibitem{li2016noise}
J.~Li, Q.~Yuan, H.~Shen, and L.~Zhang, ``Noise removal from hyperspectral image
  with joint spectral--spatial distributed sparse representation,'' {\em IEEE
  Trans. Geosci. Remote Sens.}, vol.~54, no.~9, pp.~5425--5439, 2016.

\bibitem{wu2017structure}
Z.~Wu, Q.~Wang, J.~Jin, and Y.~Shen, ``Structure tensor total
  variation-regularized weighted nuclear norm minimization for hyperspectral
  image mixed denoising,'' {\em Signal Process.}, vol.~131, pp.~202--219, 2017.

\bibitem{CVPR2014Meng}
Y.~Peng, D.~Meng, Z.~Xu, C.~Gao, Y.~Yang, and B.~Zhang, ``Decomposable nonlocal
  tensor dictionary learning for multispectral image denoising,'' in {\em Proc.
  CVPR}, pp.~2949--2956, 2014.

\bibitem{he2018non}
W.~{He}, Q.~{Yao}, C.~{Li}, N.~{Yokoya}, Q.~{Zhao}, H.~{Zhang}, and L.~{Zhang},
  ``Non-local meets global: An integrated paradigm for hyperspectral image
  restoration,'' {\em IEEE Trans. Pattern Anal. Mach. Intell.}, pp.~1--1,
  10.1109/TPAMI.2020.3027563, 2020.

\bibitem{he2015hyperspectral}
W.~He, H.~Zhang, L.~Zhang, and H.~Shen, ``Hyperspectral image denoising via
  noise-adjusted iterative low-rank matrix approximation,'' {\em IEEE J. Sel.
  Top. Appl. Earth Obs. Remote Sens.}, vol.~8, no.~6, pp.~3050--3061, 2015.

\bibitem{xie2016hyperspectral}
Y.~Xie, Y.~Qu, D.~Tao, W.~Wu, Q.~Yuan, W.~Zhang, {\em et~al.}, ``Hyperspectral
  image restoration via iteratively regularized weighted schatten p-norm
  minimization.,'' {\em IEEE Trans. Geosci. Remote Sens.}, vol.~54,
  pp.~4642--4659, Aug. 2016.

\bibitem{xie2017kronecker}
Q.~Xie, Q.~Zhao, D.~Meng, and Z.~Xu, ``Kronecker-basis-representation based
  tensor sparsity and its applications to tensor recovery,'' {\em IEEE Trans.
  Pattern Anal. Mach. Intell.}, vol.~40, no.~8, pp.~1888--1902, 2018.

\bibitem{chen2020TCYB}
Y.~{Chen}, W.~{He}, N.~{Yokoya}, and T.~{Huang}, ``Hyperspectral image
  restoration using weighted group sparsity-regularized low-rank tensor
  decomposition,'' {\em IEEE Trans. Cybern.}, vol.~50, no.~8, pp.~3556--3570,
  2020.

\bibitem{chen2020TGRS}
Y.~{Chen}, W.~{He}, N.~{Yokoya}, T.~{Huang}, and X.~{Zhao}, ``Nonlocal
  tensor-ring decomposition for hyperspectral image denoising,'' {\em IEEE
  Trans. Geosci. Remote Sens.}, vol.~58, no.~2, pp.~1348--1362, 2020.

\bibitem{chang2018hsi}
Y.~Chang, L.~Yan, H.~Fang, S.~Zhong, and W.~Liao, ``Hsi-denet: Hyperspectral
  image restoration via convolutional neural network,'' {\em IEEE Trans.
  Geosci. Remote Sens.}, pp.~1--16, 2018.

\bibitem{yuanHSID-CNN2018}
Q.~Yuan, Q.~Zhang, J.~Li, H.~Shen, and L.~Zhang, ``Hyperspectral image
  denoising employing a spatial-spectral deep residual convolutional neural
  network,'' {\em IEEE Trans. Geosci. Remote Sens.}, pp.~1--14, 2018.

\bibitem{liu2012denoising}
X.~Liu, S.~Bourennane, and C.~Fossati, ``Denoising of hyperspectral images
  using the parafac model and statistical performance analysis,'' {\em IEEE
  Trans. Geosci. Remote Sens.}, vol.~50, no.~10, pp.~3717--3724, 2012.

\bibitem{chang2017hyper}
Y.~Chang, L.~Yan, and S.~Zhong, ``Hyper-laplacian regularized unidirectional
  low-rank tensor recovery for multispectral image denoising,'' in {\em Proc.
  CVPR}, pp.~4260--4268, 2017.

\bibitem{renard2008denoising}
N.~{Renard}, S.~{Bourennane}, and J.~{Blanc-Talon}, ``Denoising and
  dimensionality reduction using multilinear tools for hyperspectral images,''
  {\em IEEE Geosci. Remote Sens. Lett.}, vol.~5, pp.~138--142, April 2008.

\bibitem{zhang2020}
H.~{Zhang}, L.~{Liu}, W.~{He}, and L.~{Zhang}, ``Hyperspectral image denoising
  with total variation regularization and nonlocal low-rank tensor
  decomposition,'' {\em IEEE Trans. Geosci. Remote Sens.}, vol.~58,
  pp.~3071--3084, May 2020.

\bibitem{zhuang2018fast}
L.~Zhuang and J.~M. Bioucas-Dias, ``Fast hyperspectral image denoising and
  inpainting based on low-rank and sparse representations,'' {\em IEEE J. Sel.
  Top. Appl. Earth Obs. Remote Sens.}, vol.~11, no.~3, pp.~730--742, 2018.

\bibitem{Chen2011TGRS}
G.~Chen and S.-E. Qian, ``Denoising of hyperspectral imagery using principal
  component analysis and wavelet shrinkage,'' {\em IEEE Trans. Geosci. Remote
  Sens.}, vol.~49, pp.~973--980, Mar. 2011.

\bibitem{zhou2011godec}
T.~Zhou and D.~Tao, ``Godec: Randomized low-rank and sparse matrix
  decomposition in noisy case,'' in {\em Proc. ICML}, 2011.

\bibitem{chenyongyong2017TGRS}
Y.~{Chen}, Y.~{Guo}, Y.~{Wang}, D.~{Wang}, C.~{Peng}, and G.~{He}, ``Denoising
  of hyperspectral images using nonconvex low rank matrix approximation,'' {\em
  IEEE Trans. Geosci. Remote Sens.}, vol.~55, no.~9, pp.~5366--5380, 2017.

\bibitem{Ye2019TGRS}
H.~{Ye}, H.~{Li}, B.~{Yang}, F.~{Cao}, and Y.~{Tang}, ``A novel rank
  approximation method for mixture noise removal of hyperspectral images,''
  {\em IEEE Trans. Geosci. Remote Sens.}, vol.~57, no.~7, pp.~4457--4469, 2019.

\bibitem{Xie2020TIP}
T.~{Xie}, S.~{Li}, and B.~{Sun}, ``Hyperspectral images denoising via nonconvex
  regularized low-rank and sparse matrix decomposition,'' {\em IEEE Trans.
  Image Process.}, vol.~29, pp.~44--56, 2020.

\bibitem{Fan2018TGRS}
H.~{Fan}, C.~{Li}, Y.~{Guo}, G.~{Kuang}, and J.~{Ma}, ``Spatial-spectral total
  variation regularized low-rank tensor decomposition for hyperspectral image
  denoising,'' {\em IEEE Trans. Geosci. Remote Sens.}, vol.~56, no.~10,
  pp.~6196--6213, 2018.

\bibitem{Wang2018}
Y.~{Wang}, J.~{Peng}, Q.~{Zhao}, Y.~{Leung}, X.~{Zhao}, and D.~{Meng},
  ``Hyperspectral image restoration via total variation regularized low-rank
  tensor decomposition,'' {\em IEEE J. Sel. Top. Appl. Earth Obs. Remote
  Sens.}, vol.~11, no.~4, pp.~1227--1243, 2018.

\bibitem{he2017Jstars}
W.~{He}, H.~{Zhang}, H.~{Shen}, and L.~{Zhang}, ``Hyperspectral image denoising
  using local low-rank matrix recovery and global spatial-spectral total
  variation,'' {\em IEEE J. Sel. Top. Appl. Earth Obs. Remote Sens.}, vol.~11,
  no.~3, pp.~713--729, 2018.

\bibitem{ZengTGRS2020}
H.~{Zeng}, X.~{Xie}, H.~{Cui}, H.~{Yin}, and J.~{Ning}, ``Hyperspectral image
  restoration via global l\_\{1-2\} spatial-spectral total variation
  regularized local low-rank tensor recovery,'' {\em IEEE Trans. Geosci. Remote
  Sens.}, pp.~1--17, 2020.

\bibitem{Sun2017letter}
L.~{Sun}, B.~{Jeon}, Y.~{Zheng}, and Z.~{Wu}, ``Hyperspectral image restoration
  using low-rank representation on spectral difference image,'' {\em IEEE
  Geosci. Remote Sens. Lett.}, vol.~14, no.~7, pp.~1151--1155, 2017.

\bibitem{Peng2020TIP}
J.~{Peng}, Q.~{Xie}, Q.~{Zhao}, Y.~{Wang}, L.~{Yee}, and D.~{Meng}, ``Enhanced
  3dtv regularization and its applications on hsi denoising and compressed
  sensing,'' {\em IEEE Tran. Image Process.}, vol.~29, pp.~7889--7903, 2020.

\bibitem{ZhengTGRS2020}
Y.~{Zheng}, T.~{Huang}, X.~{Zhao}, Y.~{Chen}, and W.~{He},
  ``Double-factor-regularized low-rank tensor factorization for mixed noise
  removal in hyperspectral image,'' {\em IEEE Trans. Geosci. Remote Sens.},
  pp.~1--15, 2020.

\bibitem{alonso2019data}
K.~Alonso, M.~Bachmann, K.~Burch, E.~Carmona, D.~Cerra, R.~De~los Reyes,
  D.~Dietrich, U.~Heiden, A.~H{\"o}lderlin, J.~Ickes, {\em et~al.}, ``Data
  products, quality and validation of the dlr earth sensing imaging
  spectrometer (desis),'' {\em Sensors}, vol.~19, no.~20, p.~4471, 2019.

\bibitem{li2018discriminant}
W.~Li, F.~Feng, H.~Li, and Q.~Du, ``Discriminant analysis-based dimension
  reduction for hyperspectral image classification: A survey of the most recent
  advances and an experimental comparison of different techniques,'' {\em IEEE
  Geosci. Remote Sens. Mag.}, vol.~6, no.~1, pp.~15--34, 2018.

\bibitem{rasti2020feature}
B.~Rasti, D.~Hong, R.~Hang, P.~Ghamisi, X.~Kang, J.~Chanussot, and
  J.~Benediktsson, ``Feature extraction for hyperspectral imagery: The
  evolution from shallow to deep (overview and toolbox),'' {\em IEEE Geosci.
  Remote Sens. Mag.}, vol.~8, no.~4, pp.~60--88, 2020.

\bibitem{wu2019orsim}
X.~Wu, D.~Hong, J.~Tian, J.~Chanussot, W.~Li, and R.~Tao, ``Orsim detector: A
  novel object detection framework in optical remote sensing imagery using
  spatial-frequency channel features,'' {\em IEEE Trans. Geosci. Remote Sens.},
  vol.~57, no.~7, pp.~5146--5158, 2019.

\bibitem{wu2020fourier}
X.~Wu, D.~Hong, J.~Chanussot, Y.~Xu, R.~Tao, and Y.~Wang, ``Fourier-based
  rotation-invariant feature boosting: An efficient framework for geospatial
  object detection,'' {\em IEEE Geosci. Remote Sens. Lett.}, vol.~17, no.~2,
  pp.~302--306, 2020.

\bibitem{lee2001algorithms}
D.~D. Lee and H.~S. Seung, ``Algorithms for non-negative matrix
  factorization,'' in {\em Proc. NIPS}, pp.~556--562, 2001.

\bibitem{gillis2013sparse}
N.~Gillis and R.~J. Plemmons, ``Sparse nonnegative matrix underapproximation
  and its application to hyperspectral image analysis,'' {\em Linear Algebra
  its Appl.}, vol.~438, no.~10, pp.~3991--4007, 2013.

\bibitem{yan2018novel}
W.~Yan, B.~Zhang, and Z.~Yang, ``A novel regularized nonnegative matrix
  factorization for spectral-spatial dimension reduction of hyperspectral
  imagery,'' {\em IEEE Access}, vol.~6, pp.~77953--77964, 2018.

\bibitem{wen2016orthogonal}
J.~Wen, J.~E. Fowler, M.~He, Y.-Q. Zhao, C.~Deng, and V.~Menon, ``Orthogonal
  nonnegative matrix factorization combining multiple features for
  spectral--spatial dimensionality reduction of hyperspectral imagery,'' {\em
  IEEE Trans. Geosci. Remote Sens.}, vol.~54, no.~7, pp.~4272--4286, 2016.

\bibitem{rasti2016hyperspectral}
B.~Rasti, M.~O. Ulfarsson, and J.~R. Sveinsson, ``Hyperspectral feature
  extraction using total variation component analysis,'' {\em IEEE Trans.
  Geosci. Remote Sens.}, vol.~54, no.~12, pp.~6976--6985, 2016.

\bibitem{an2018tensor}
J.~An, X.~Zhang, H.~Zhou, and L.~Jiao, ``Tensor-based low-rank graph with
  multimanifold regularization for dimensionality reduction of hyperspectral
  images,'' {\em IEEE Trans. Geosci. Remote Sens.}, vol.~56, no.~8,
  pp.~4731--4746, 2018.

\bibitem{belkin2003laplacian}
M.~Belkin and P.~Niyogi, ``Laplacian eigenmaps for dimensionality reduction and
  data representation,'' {\em Neural Comput.}, vol.~15, no.~6, pp.~1373--1396,
  2003.

\bibitem{aharon2006k}
M.~Aharon, M.~Elad, and A.~Bruckstein, ``K-svd: An algorithm for designing
  overcomplete dictionaries for sparse representation,'' {\em IEEE Trans.
  Signal Process.}, vol.~54, no.~11, pp.~4311--4322, 2006.

\bibitem{nie2010efficient}
F.~Nie, H.~Huang, X.~Cai, and C.~H. Ding, ``Efficient and robust feature
  selection via joint $\ell$2,1-norms minimization,'' in {\em Proc. NIPS},
  pp.~1813--1821, 2010.

\bibitem{recht2010guaranteed}
B.~Recht, M.~Fazel, and P.~A. Parrilo, ``Guaranteed minimum-rank solutions of
  linear matrix equations via nuclear norm minimization,'' {\em SIAM Review},
  vol.~52, no.~3, pp.~471--501, 2010.

\bibitem{wu2017joint}
J.~Wu, Z.~Lin, and H.~Zha, ``Joint latent subspace learning and regression for
  cross-modal retrieval,'' in {\em Proc. ACM SIGIR}, pp.~917--920, 2017.

\bibitem{ma2010local}
L.~Ma, M.~M. Crawford, and J.~Tian, ``Local manifold learning-based
  $k$-nearest-neighbor for hyperspectral image classification,'' {\em IEEE
  Trans. Geosci. Remote Sens.}, vol.~48, no.~11, pp.~4099--4109, 2010.

\bibitem{roweis2000nonlinear}
S.~T. Roweis and L.~K. Saul, ``Nonlinear dimensionality reduction by locally
  linear embedding,'' {\em Science}, vol.~290, no.~5500, pp.~2323--2326, 2000.

\bibitem{zhang2004principal}
Z.~Zhang and H.~Zha, ``Principal manifolds and nonlinear dimensionality
  reduction via tangent space alignment,'' {\em SIAM J. Sci. Comput.}, vol.~26,
  no.~1, pp.~313--338, 2004.

\bibitem{huang2015dimensionality}
H.~Huang, F.~Luo, J.~Liu, and Y.~Yang, ``Dimensionality reduction of
  hyperspectral images based on sparse discriminant manifold embedding,'' {\em
  ISPRS J. Photogramm. Remote Sens.}, vol.~106, pp.~42--54, 2015.

\bibitem{he2016weighted}
W.~He, H.~Zhang, L.~Zhang, W.~Philips, and W.~Liao, ``Weighted sparse graph
  based dimensionality reduction for hyperspectral images,'' {\em IEEE Geosci.
  Remote Sens. Lett.}, vol.~13, no.~5, pp.~686--690, 2016.

\bibitem{hong2017learning}
D.~Hong, N.~Yokoya, and X.~X. Zhu, ``Learning a robust local manifold
  representation for hyperspectral dimensionality reduction,'' {\em IEEE J.
  Sel. Top. Appl. Earth Obs. Remote Sens.}, vol.~10, no.~6, pp.~2960--2975,
  2017.

\bibitem{an2018patch}
J.~An, X.~Zhang, H.~Zhou, J.~Feng, and L.~Jiao, ``Patch tensor-based sparse and
  low-rank graph for hyperspectral images dimensionality reduction,'' {\em IEEE
  J. Sel. Top. Appl. Earth Obs. Remote Sens.}, vol.~11, no.~7, pp.~2513--2527,
  2018.

\bibitem{ly2013sparse}
N.~H. Ly, Q.~Du, and J.~E. Fowler, ``Sparse graph-based discriminant analysis
  for hyperspectral imagery,'' {\em IEEE Trans. Geosci. Remote Sens.}, vol.~52,
  no.~7, pp.~3872--3884, 2013.

\bibitem{ly2014collaborative}
N.~H. Ly, Q.~Du, and J.~E. Fowler, ``Collaborative graph-based discriminant
  analysis for hyperspectral imagery,'' {\em IEEE J. Sel. Top. Appl. Earth Obs.
  Remote Sens.}, vol.~7, no.~6, pp.~2688--2696, 2014.

\bibitem{imani2015feature}
M.~Imani and H.~Ghassemian, ``Feature space discriminant analysis for
  hyperspectral data feature reduction,'' {\em ISPRS J. Photogramm. Remote
  Sens.}, vol.~102, pp.~1--13, 2015.

\bibitem{huang2019spatial}
H.~Huang, Y.~Duan, H.~He, G.~Shi, and F.~Luo, ``Spatial-spectral local
  discriminant projection for dimensionality reduction of hyperspectral
  image,'' {\em ISPRS J. Photogramm. Remote Sens.}, vol.~156, pp.~77--93, 2019.

\bibitem{hong2020graph1}
D.~Hong, L.~Gao, J.~Yao, B.~Zhang, P.~Antonio, and J.~Chanussot, ``Graph
  convolutional networks for hyperspectral image classification,'' {\em IEEE
  Trans. Geosci. Remote Sens.}, 2020.
\newblock DOI: 10.1109/TGRS.2020.3015157.

\bibitem{hong2019regression}
D.~Hong, {\em Regression-Induced Representation Learning and Its Optimizer: A
  Novel Paradigm to Revisit Hyperspectral Imagery Analysis}.
\newblock PhD thesis, Technische Universit{\"a}t M{\"u}nchen, 2019.

\bibitem{ji2009linear}
S.~Ji and J.~Ye, ``Linear dimensionality reduction for multi-label
  classification,'' in {\em Proc. IJCAI}, 2009.

\bibitem{hong2018joint}
D.~Hong, N.~Yokoya, J.~Xu, and X.~Zhu, ``Joint \& progressive learning from
  high-dimensional data for multi-label classification,'' in {\em Proc. ECCV},
  pp.~469--484, 2018.

\bibitem{he2004locality}
X.~He and P.~Niyogi, ``Locality preserving projections,'' in {\em Proc. NIPS},
  pp.~153--160, 2004.

\bibitem{liao2012semisupervised}
W.~Liao, A.~Pizurica, P.~Scheunders, W.~Philips, and Y.~Pi, ``Semisupervised
  local discriminant analysis for feature extraction in hyperspectral images,''
  {\em IEEE Trans. Geosci. Remote Sens.}, vol.~51, no.~1, pp.~184--198, 2012.

\bibitem{zhao2014general}
M.~Zhao, Z.~Zhang, T.~W. Chow, and B.~Li, ``A general soft label based linear
  discriminant analysis for semi-supervised dimensionality reduction,'' {\em
  Neural Netw.}, vol.~55, pp.~83--97, 2014.

\bibitem{wu2018semi}
H.~Wu and S.~Prasad, ``Semi-supervised dimensionality reduction of
  hyperspectral imagery using pseudo-labels,'' {\em Pattern Recognit.},
  vol.~74, pp.~212--224, 2018.

\bibitem{hong2019learning}
D.~Hong, N.~Yokoya, J.~Chanussot, J.~Xu, and X.~X. Zhu, ``Learning to propagate
  labels on graphs: An iterative multitask regression framework for
  semi-supervised hyperspectral dimensionality reduction,'' {\em ISPRS J.
  Photogramm. Remote Sens.}, vol.~158, pp.~35--49, 2019.

\bibitem{hong2020joint}
D.~Hong, N.~Yokoya, J.~Chanussot, J.~Xu, and X.~X. Zhu, ``Joint and progressive
  subspace analysis (jpsa) with spatial-spectral manifold alignment for
  semi-supervised hyperspectral dimensionality reduction,'' {\em IEEE Trans.
  Cybern.}, 2020.
\newblock DOI: 10.1109/TCYB.2020.3028931.

\bibitem{hang2019cascaded}
R.~Hang, Q.~Liu, D.~Hong, and P.~Ghamisi, ``Cascaded recurrent neural networks
  for hyperspectral image classification,'' {\em IEEE Trans. Geosci. Remote
  Sens.}, vol.~57, no.~8, pp.~5384--5394, 2019.

\bibitem{levina2005maximum}
E.~Levina and P.~J. Bickel, ``Maximum likelihood estimation of intrinsic
  dimension,'' in {\em Proc. NIPS}, pp.~777--784, 2005.

\bibitem{bioucas2008hyperspectral}
J.~M. Bioucas-Dias and J.~M. Nascimento, ``Hyperspectral subspace
  identification,'' {\em IEEE Trans. Geosci. Remote Sens.}, vol.~46, no.~8,
  pp.~2435--2445, 2008.

\bibitem{bioucas2012hyperspectral}
J.~M. Bioucas-Dias, A.~Plaza, N.~Dobigeon, M.~Parente, Q.~Du, P.~Gader, and
  J.~Chanussot, ``Hyperspectral unmixing overview: Geometrical, statistical,
  and sparse regression-based approaches,'' {\em IEEE J. Sel. Top. Appl. Earth
  Obs. Remote Sens.}, vol.~5, no.~2, pp.~354--379, 2012.

\bibitem{keshava2002spectral}
N.~Keshava and J.~F. Mustard, ``Spectral unmixing,'' {\em IEEE Signal Process.
  Mag.}, vol.~19, no.~1, pp.~44--57, 2002.

\bibitem{yang2010blind}
Z.~Yang, G.~Zhou, S.~Xie, S.~Ding, J.-M. Yang, and J.~Zhang, ``Blind spectral
  unmixing based on sparse nonnegative matrix factorization,'' {\em IEEE Tran.
  Image Process.}, vol.~20, no.~4, pp.~1112--1125, 2010.

\bibitem{qian2011hyperspectral}
Y.~Qian, S.~Jia, J.~Zhou, and A.~Robles-Kelly, ``Hyperspectral unmixing via
  $l_{1/2}$ sparsity-constrained nonnegative matrix factorization,'' {\em IEEE
  Trans. Geosci. Remote Sens.}, vol.~49, no.~11, pp.~4282--4297, 2011.

\bibitem{sigurdsson2014hyperspectral}
J.~Sigurdsson, M.~O. Ulfarsson, and J.~R. Sveinsson, ``Hyperspectral unmixing
  with $l_{q}$ regularization,'' {\em IEEE Trans. Geosci. Remote Sens.},
  vol.~52, no.~11, pp.~6793--6806, 2014.

\bibitem{thouvenin2015hyperspectral}
P.~Thouvenin, N.~Dobigeon, and J.~Tourneret, ``Hyperspectral unmixing with
  spectral variability using a perturbed linear mixing model,'' {\em IEEE
  Trans. Signal Process.}, vol.~64, no.~2, pp.~525--538, 2015.

\bibitem{drumetz2016blind}
L.~Drumetz, M.~Veganzones, S.~Henrot, R.~Phlypo, J.~Chanussot, and C.~Jutten,
  ``Blind hyperspectral unmixing using an extended linear mixing model to
  address spectral variability,'' {\em IEEE Trans. Image Process.}, vol.~25,
  no.~8, pp.~3890--3905, 2016.

\bibitem{he2017total}
W.~He, H.~Zhang, and L.~Zhang, ``Total variation regularized reweighted sparse
  nonnegative matrix factorization for hyperspectral unmixing,'' {\em IEEE
  Trans. Geosci. Remote Sens.}, vol.~55, no.~7, pp.~3909--3921, 2017.

\bibitem{yao2019nonconvex}
J.~Yao, D.~Meng, Q.~Zhao, W.~Cao, and Z.~Xu, ``Nonconvex-sparsity and
  nonlocal-smoothness-based blind hyperspectral unmixing,'' {\em IEEE Trans.
  Image Process.}, vol.~28, no.~6, pp.~2991--3006, 2019.

\bibitem{liu2012enhancing}
J.~Liu, J.~Zhang, Y.~Gao, C.~Zhang, and Z.~Li, ``Enhancing spectral unmixing by
  local neighborhood weights,'' {\em IEEE J. Sel. Top. Appl. Earth Obs. and
  Remote Sens.}, vol.~5, no.~5, pp.~1545--1552, 2012.

\bibitem{lu2012manifold}
X.~Lu, H.~Wu, Y.~Yuan, P.~Yan, and X.~Li, ``Manifold regularized sparse nmf for
  hyperspectral unmixing,'' {\em IEEE Trans. Geosci. Remote Sens.}, vol.~51,
  no.~5, pp.~2815--2826, 2012.

\bibitem{wang2016hypergraph}
W.~Wang, Y.~Qian, and Y.~Y. Tang, ``Hypergraph-regularized sparse nmf for
  hyperspectral unmixing,'' {\em IEEE J. Sel. Top. Appl. Earth Obs. and Remote
  Sens.}, vol.~9, no.~2, pp.~681--694, 2016.

\bibitem{qin2020blind}
J.~Qin, H.~Lee, J.~T. Chi, L.~Drumetz, J.~Chanussot, Y.~Lou, and A.~L.
  Bertozzi, ``Blind hyperspectral unmixing based on graph total variation
  regularization,'' {\em IEEE Trans. Geosci. Remote Sens.}, 2020.
\newblock DOI: 10.1109/TGRS.2020.3020810.

\bibitem{qian2016matrix}
Y.~Qian, F.~Xiong, S.~Zeng, J.~Zhou, and Y.~Y. Tang, ``Matrix-vector
  nonnegative tensor factorization for blind unmixing of hyperspectral
  imagery,'' {\em IEEE Trans. Geosci. Remote Sens.}, vol.~55, no.~3,
  pp.~1776--1792, 2016.

\bibitem{imbiriba2019low}
T.~Imbiriba, R.~A. Borsoi, and J.~C.~M. Bermudez, ``Low-rank tensor modeling
  for hyperspectral unmixing accounting for spectral variability,'' {\em IEEE
  Trans. Geosci. Remote Sens.}, vol.~58, no.~3, pp.~1833--1842, 2019.

\bibitem{palsson2018hyperspectral}
B.~Palsson, J.~Sigurdsson, J.~R. Sveinsson, and M.~O. Ulfarsson,
  ``Hyperspectral unmixing using a neural network autoencoder,'' {\em IEEE
  Access}, vol.~6, pp.~25646--25656, 2018.

\bibitem{su2018stacked}
Y.~Su, A.~Marinoni, J.~Li, J.~Plaza, and P.~Gamba, ``Stacked nonnegative sparse
  autoencoders for robust hyperspectral unmixing,'' {\em IEEE Geosci. Remote
  Sens.Lett.}, vol.~15, no.~9, pp.~1427--1431, 2018.

\bibitem{palsson2019spectral}
B.~Palsson, J.~R. Sveinsson, and M.~O. Ulfarsson, ``Spectral-spatial
  hyperspectral unmixing using multitask learning,'' {\em IEEE Access}, vol.~7,
  pp.~148861--148872, 2019.

\bibitem{han2020deep}
Z.~Han, D.~Hong, L.~Gao, B.~Zhang, and J.~Chanussot, ``Deep half-siamese
  networks for hyperspectral unmixing,'' {\em IEEE Geosci. Remote Sens.Lett.},
  2020.
\newblock DOI:10.1109/LGRS.2020.3011941.

\bibitem{nascimento2005vertex}
J.~M. Nascimento and J.~M. Dias, ``Vertex component analysis: A fast algorithm
  to unmix hyperspectral data,'' {\em IEEE Trans. Geosci. Remote Sens.},
  vol.~43, no.~4, pp.~898--910, 2005.

\bibitem{chang2006fast}
C.-I. Chang and A.~Plaza, ``A fast iterative algorithm for implementation of
  pixel purity index,'' {\em IEEE Geosci. Remote Sens. Lett.}, vol.~3, no.~1,
  pp.~63--67, 2006.

\bibitem{winter1999n}
M.~E. Winter, ``N-findr: An algorithm for fast autonomous spectral end-member
  determination in hyperspectral data,'' in {\em Imaging Spectrometry V},
  vol.~3753, pp.~266--275, International Society for Optics and Photonics,
  1999.

\bibitem{heinz2001fully}
D.~C. Heinz and C.-I. Chang, ``Fully constrained least squares linear spectral
  mixture analysis method for material quantification in hyperspectral
  imagery,'' {\em IEEE Trans. Geosci. Remote Sens.}, vol.~39, no.~3,
  pp.~529--545, 2001.

\bibitem{heylen2011fully}
R.~Heylen, D.~Burazerovic, and P.~Scheunders, ``Fully constrained least squares
  spectral unmixing by simplex projection,'' {\em IEEE Trans. Geosci. Remote
  Sens.}, vol.~49, no.~11, pp.~4112--4122, 2011.

\bibitem{bioucas2010alternating}
J.~M. Bioucas-Dias and M.~A. Figueiredo, ``Alternating direction algorithms for
  constrained sparse regression: Application to hyperspectral unmixing,'' in
  {\em Proc. WHISPERS}, pp.~1--4, IEEE, 2010.

\bibitem{iordache2012total}
M.-D. Iordache, J.~M. Bioucas-Dias, and A.~Plaza, ``Total variation spatial
  regularization for sparse hyperspectral unmixing,'' {\em IEEE Trans. Geosci.
  Remote Sens.}, vol.~50, no.~11, pp.~4484--4502, 2012.

\bibitem{iordache2013collaborative}
M.-D. Iordache, J.~M. Bioucas-Dias, and A.~Plaza, ``Collaborative sparse
  regression for hyperspectral unmixing,'' {\em IEEE Trans. Geosci. Remote
  Sens.}, vol.~52, no.~1, pp.~341--354, 2013.

\bibitem{fu2016semiblind}
X.~Fu, W.~Ma, J.~Bioucas-Dias, and T.~Chan, ``Semiblind hyperspectral unmixing
  in the presence of spectral library mismatches,'' {\em IEEE Trans. Geosci.
  Remote Sens.}, vol.~54, no.~9, pp.~5171--5184, 2016.

\bibitem{huang2018joint}
J.~Huang, T.-Z. Huang, L.-J. Deng, and X.-L. Zhao, ``Joint-sparse-blocks and
  low-rank representation for hyperspectral unmixing,'' {\em IEEE Trans.
  Geosci. Remote Sens.}, vol.~57, no.~4, pp.~2419--2438, 2018.

\bibitem{hong2018sulora}
D.~Hong and X.~Zhu, ``S{UL}o{RA}: Subspace unmixing with low-rank attribute
  embedding for hyperspectral data analysis,'' {\em IEEE J. Sel. Topics Signal
  Process.}, vol.~12, no.~6, pp.~1351--1363, 2018.

\bibitem{hong2019augmented}
D.~Hong, N.~Yokoya, J.~Chanussot, and X.~Zhu, ``An augmented linear mixing
  model to address spectral variability for hyperspectral unmixing,'' {\em IEEE
  Trans. Image Process.}, vol.~28, no.~4, pp.~1923--1938, 2019.

\bibitem{heylen2014review}
R.~Heylen, M.~Parente, and P.~Gader, ``A review of nonlinear hyperspectral
  unmixing methods,'' {\em IEEE J. Sel. Top. Appl. Earth Obs. Remote Sens.},
  vol.~7, no.~6, pp.~1844--1868, 2014.

\bibitem{yokoya2017hyperspectral}
N.~Yokoya, C.~Grohnfeldt, and J.~Chanussot, ``Hyperspectral and multispectral
  data fusion: A comparative review of the recent literature,'' {\em IEEE
  Geosci. Remote Sens. Mag.}, vol.~5, no.~2, pp.~29--56, 2017.

\bibitem{Eismann2004}
M.~T. {Eismann} and R.~C. {Hardie}, ``Application of the stochastic mixing
  model to hyperspectral resolution enhancement,'' {\em IEEE Trans. Geosci.
  Remote Sens.}, vol.~42, no.~9, pp.~1924--1933, 2004.

\bibitem{yokoya2012coupled}
N.~Yokoya, T.~Yairi, and A.~Iwasaki, ``Coupled nonnegative matrix factorization
  unmixing for hyperspectral and multispectral data fusion,'' {\em IEEE Trans.
  Geosci. Remote Sens.}, vol.~50, no.~2, pp.~528--537, 2012.

\bibitem{QiWei2015TGRS}
Q.~{Wei}, J.~{Bioucas-Dias}, N.~{Dobigeon}, and J.~{Tourneret}, ``Hyperspectral
  and multispectral image fusion based on a sparse representation,'' {\em IEEE
  Trans. Geosci. Remote Sens.}, vol.~53, no.~7, pp.~3658--3668, 2015.

\bibitem{Simoes2015}
M.~{Simoes}, J.~{Bioucas‐Dias}, L.~B. {Almeida}, and J.~{Chanussot}, ``A
  convex formulation for hyperspectral image superresolution via subspace-based
  regularization,'' {\em IEEE Trans. Geosci. Remote Sens.}, vol.~53, no.~6,
  pp.~3373--3388, 2015.

\bibitem{he2020hyperspectral}
W.~He, Y.~Chen, N.~Yokoya, C.~Li, and Q.~Zhao, ``Hyperspectral super-resolution
  via coupled tensor ring factorization,'' {\em arXiv preprint
  arXiv:2001.01547}, 2020.

\bibitem{loncan2015hyperspectral}
L.~Loncan, L.~B. De~Almeida, J.~M. Bioucas-Dias, X.~Briottet, J.~Chanussot,
  N.~Dobigeon, S.~Fabre, W.~Liao, G.~A. Licciardi, M.~Simoes, {\em et~al.},
  ``Hyperspectral pansharpening: A review,'' {\em IEEE Geosci. Remote Sens.
  Mag.}, vol.~3, no.~3, pp.~27--46, 2015.

\bibitem{akhtar2014sparse}
N.~Akhtar, F.~Shafait, and A.~Mian, ``Sparse spatio-spectral representation for
  hyperspectral image super-resolution,'' in {\em Proc. ECCV}, pp.~63--78,
  Springer, 2014.

\bibitem{akhtar2015bayesian}
N.~Akhtar, F.~Shafait, and A.~Mian, ``Bayesian sparse representation for
  hyperspectral image super resolution,'' in {\em Proc. CVPR}, pp.~3631--3640,
  2015.

\bibitem{dong2016hyperspectral}
W.~Dong, F.~Fu, G.~Shi, X.~Cao, J.~Wu, G.~Li, and X.~Li, ``Hyperspectral image
  super-resolution via non-negative structured sparse representation,'' {\em
  IEEE Trans. Image Process.}, vol.~25, no.~5, pp.~2337--2352, 2016.

\bibitem{Wang2020TGRS}
K.~{Wang}, Y.~{Wang}, X.~L. {Zhao}, J.~C.~W. {Chan}, Z.~{Xu}, and D.~{Meng},
  ``Hyperspectral and multispectral image fusion via nonlocal low-rank tensor
  decomposition and spectral unmixing,'' {\em IEEE Trans. Geosci. Remote
  Sens.}, vol.~58, no.~11, pp.~7654--7671, 2020.

\bibitem{wei2015fast}
Q.~Wei, N.~Dobigeon, and J.-Y. Tourneret, ``Fast fusion of multi-band images
  based on solving a sylvester equation,'' {\em IEEE Trans. Image Process.},
  vol.~24, no.~11, pp.~4109--4121, 2015.

\bibitem{kawakami2011high}
R.~Kawakami, Y.~Matsushita, J.~Wright, M.~Ben-Ezra, Y.-W. Tai, and K.~Ikeuchi,
  ``High-resolution hyperspectral imaging via matrix factorization,'' in {\em
  Proc. CVPR}, pp.~2329--2336, IEEE, 2011.

\bibitem{wu2020hybrid}
R.~Wu, H.-T. Wai, and W.-K. Ma, ``Hybrid inexact bcd for coupled structured
  matrix factorization in hyperspectral super-resolution,'' {\em IEEE Trans.
  Signal Process.}, vol.~68, pp.~1728--1743, 2020.

\bibitem{li2018fusing}
S.~Li, R.~Dian, L.~Fang, and J.~M. Bioucas-Dias, ``Fusing hyperspectral and
  multispectral images via coupled sparse tensor factorization,'' {\em IEEE
  Trans. Image Process.}, vol.~27, no.~8, pp.~4118--4130, 2018.

\bibitem{kanatsoulis2018hyperspectral}
C.~I. Kanatsoulis, X.~Fu, N.~D. Sidiropoulos, and W.-K. Ma, ``Hyperspectral
  super-resolution: A coupled tensor factorization approach,'' {\em IEEE Trans.
  Signal Process.}, vol.~66, no.~24, pp.~6503--6517, 2018.

\bibitem{dian2019learning}
R.~Dian, S.~Li, and L.~Fang, ``Learning a low tensor-train rank representation
  for hyperspectral image super-resolution,'' {\em IEEE Trans. Neural Netw.
  Learn. Syst.}, vol.~30, no.~9, pp.~2672--2683, 2019.

\bibitem{xu2020hyperspectral}
Y.~Xu, Z.~Wu, J.~Chanussot, and Z.~Wei, ``Hyperspectral images super-resolution
  via learning high-order coupled tensor ring representation,'' {\em IEEE
  Trans. Neural Netw. Learn. Syst.}, 2020.

\bibitem{dian2017hyperspectral}
R.~Dian, L.~Fang, and S.~Li, ``Hyperspectral image super-resolution via
  non-local sparse tensor factorization,'' in {\em Proc. CVPR}, pp.~5344--5353,
  2017.

\bibitem{wang2017hyperspectral}
Y.~Wang, X.~Chen, Z.~Han, S.~He, {\em et~al.}, ``Hyperspectral image
  super-resolution via nonlocal low-rank tensor approximation and total
  variation regularization,'' {\em Remote Sens.}, vol.~9, no.~12, p.~1286,
  2017.

\bibitem{xu2019nonlocal}
Y.~Xu, Z.~Wu, J.~Chanussot, and Z.~Wei, ``Nonlocal patch tensor sparse
  representation for hyperspectral image super-resolution,'' {\em IEEE Trans.
  Image Process.}, vol.~28, no.~6, pp.~3034--3047, 2019.

\bibitem{mei2017hyperspectral}
S.~Mei, X.~Yuan, J.~Ji, Y.~Zhang, S.~Wan, and Q.~Du, ``Hyperspectral image
  spatial super-resolution via 3d full convolutional neural network,'' {\em
  Remote Sens.}, vol.~9, no.~11, p.~1139, 2017.

\bibitem{haut2018new}
J.~M. Haut, R.~Fernandez-Beltran, M.~E. Paoletti, J.~Plaza, A.~Plaza, and
  F.~Pla, ``A new deep generative network for unsupervised remote sensing
  single-image super-resolution,'' {\em IEEE Trans. Geosci.Remote Sens.},
  vol.~56, no.~11, pp.~6792--6810, 2018.

\bibitem{liu2019stfnet}
X.~Liu, C.~Deng, J.~Chanussot, D.~Hong, and B.~Zhao, ``Stfnet: A two-stream
  convolutional neural network for spatiotemporal image fusion,'' {\em IEEE
  Trans. Geosci. Remote Sens.}, vol.~57, no.~9, pp.~6552--6564, 2019.

\bibitem{liu2019efficient}
W.~Liu and J.~Lee, ``An efficient residual learning neural network for
  hyperspectral image superresolution,'' {\em IEEE J. Sel. Top. Appl. Earth
  Obs. Remote Sens.}, vol.~12, no.~4, pp.~1240--1253, 2019.

\bibitem{zheng2020coupled}
K.~Zheng, L.~Gao, W.~Liao, D.~Hong, B.~Zhang, X.~Cui, and J.~Chanussot,
  ``Coupled convolutional neural network with adaptive response function
  learning for unsupervised hyperspectral super resolution,'' {\em IEEE Trans.
  Geosci. Remote Sens.}, 2020.

\bibitem{uezato2020guided}
T.~Uezato, D.~Hong, N.~Yokoya, and W.~He, ``Guided deep decoder: Unsupervised
  image pair fusion,'' in {\em Proc. ECCV}, pp.~87--102, Springer, 2020.

\bibitem{yao2020cross}
J.~Yao, D.~Hong, J.~Chanussot, D.~Meng, X.~Zhu, and Z.~Xu, ``Cross-attention in
  coupled unmixing nets for unsupervised hyperspectral super-resolution,'' in
  {\em Proc. ECCV}, pp.~208--224, Springer, 2020.

\bibitem{lanaras2015hyperspectral}
C.~Lanaras, E.~Baltsavias, and K.~Schindler, ``Hyperspectral super-resolution
  by coupled spectral unmixing,'' in {\em Proc. ICCV}, pp.~3586--3594, 2015.

\bibitem{wald2000quality}
L.~Wald, ``Quality of high resolution synthesised images: Is there a simple
  criterion?,'' in {\em Proc. ICFED}, pp.~99--103, 2000.

\bibitem{wang2004image}
Z.~Wang, A.~C. Bovik, H.~R. Sheikh, and E.~P. Simoncelli, ``Image quality
  assessment: from error visibility to structural similarity,'' {\em IEEE
  Trans. Image Process.}, vol.~13, no.~4, pp.~600--612, 2004.

\bibitem{dalla2015challenges}
M.~Dalla~Mura, S.~Prasad, F.~Pacifici, P.~Gamba, J.~Chanussot, and J.~A.
  Benediktsson, ``Challenges and opportunities of multimodality and data fusion
  in remote sensing,'' {\em Proc. IEEE}, vol.~103, no.~9, pp.~1585--1601, 2015.

\bibitem{hong2020more}
D.~Hong, L.~Gao, N.~Yokoya, J.~Yao, J.~Chanussot, D.~Qian, and B.~Zhang, ``More
  diverse means better: Multimodal deep learning meets remote-sensing imagery
  classification,'' {\em IEEE Trans. Geosci. Remote Sens.}, 2020.
\newblock DOI: 10.1109/TGRS.2020.3016820.

\bibitem{hong2020x}
D.~Hong, N.~Yokoya, G.-S. Xia, J.~Chanussot, and X.~X. Zhu, ``X-modalnet: A
  semi-supervised deep cross-modal network for classification of remote sensing
  data,'' {\em ISPRS J. Photogramm. Remote Sens.}, vol.~167, pp.~12--23, 2020.

\bibitem{ngiam2011multimodal}
J.~Ngiam, A.~Khosla, M.~Kim, J.~Nam, H.~Lee, and A.~Ng, ``Multimodal deep
  learning,'' in {\em Proc. ICML}, pp.~689--696, 2011.

\bibitem{wang2009general}
C.~Wang and S.~Mahadevan, ``A general framework for manifold alignment.,'' in
  {\em Proc. AAAI}, pp.~79--86, 2009.

\bibitem{matasci2011transfer}
G.~Matasci, M.~Volpi, D.~Tuia, and M.~Kanevski, ``Transfer component analysis
  for domain adaptation in image classification,'' in {\em Image and Signal
  Processing for Remote Sensing XVII}, vol.~8180, p.~81800F, International
  Society for Optics and Photonics, 2011.

\bibitem{tuia2014semisupervised}
D.~Tuia, M.~Volpi, M.~Trolliet, and G.~Camps-Valls, ``Semisupervised manifold
  alignment of multimodal remote sensing images,'' {\em IEEE Trans. Geosci.
  Remote Sens.}, vol.~52, no.~12, pp.~7708--7720, 2014.

\bibitem{wang2011heterogeneous}
C.~Wang and S.~Mahadevan, ``Heterogeneous domain adaptation using manifold
  alignment,'' in {\em Proc. IJCAI}, vol.~22, p.~1541, 2011.

\bibitem{matasci2015semisupervised}
G.~Matasci, M.~Volpi, M.~Kanevski, L.~Bruzzone, and D.~Tuia, ``Semisupervised
  transfer component analysis for domain adaptation in remote sensing image
  classification,'' {\em IEEE Trans. Geosci. Remote Sens.}, vol.~53, no.~7,
  pp.~3550--3564, 2015.

\bibitem{pan2011domain}
S.~J. Pan, I.~W. Tsang, J.~T. Kwok, and Q.~Yang, ``Domain adaptation via
  transfer component analysis,'' {\em IEEE Trans. Neural Netw.}, vol.~22,
  no.~2, pp.~199--210, 2011.

\bibitem{tuia2016kernel}
D.~Tuia and G.~Camps-Valls, ``Kernel manifold alignment for domain
  adaptation,'' {\em PloS One}, vol.~11, no.~2, p.~e0148655, 2016.

\bibitem{hu2019comparative}
J.~Hu, D.~Hong, Y.~Wang, and X.~X. Zhu, ``A comparative review of manifold
  learning techniques for hyperspectral and polarimetric sar image fusion,''
  {\em Remote Sens.}, vol.~11, no.~6, p.~681, 2019.

\bibitem{hu2019mima}
J.~Hu, D.~Hong, and X.~X. Zhu, ``Mima: Mapper-induced manifold alignment for
  semi-supervised fusion of optical image and polarimetric sar data,'' {\em
  IEEE Trans. Geosci. Remote Sens.}, vol.~57, no.~11, pp.~9025--9040, 2019.

\bibitem{hong2019cospace}
D.~Hong, N.~Yokoya, J.~Chanussot, and X.~X. Zhu, ``Co{S}pace: Common subspace
  learning from hyperspectral-multispectral correspondences,'' {\em IEEE Trans.
  Geos. Remote Sens.}, vol.~57, no.~7, pp.~4349--4359, 2019.

\bibitem{hong2019learnable}
D.~Hong, N.~Yokoya, N.~Ge, J.~Chanussot, and X.~Zhu, ``Learnable manifold
  alignment ({L}e{MA}): A semi-supervised cross-modality learning framework for
  land cover and land use classification,'' {\em ISPRS J. Photogramm. Remote
  Sens.}, vol.~147, pp.~193--205, 2019.

\bibitem{hong2020learning}
D.~Hong, J.~Chanussot, N.~Yokoya, J.~Kang, and X.~X. Zhu, ``Learning-shared
  cross-modality representation using multispectral-lidar and hyperspectral
  data,'' {\em IEEE Geosci. Remote Sens. Lett.}, vol.~17, no.~8,
  pp.~1470--1474, 2020.

\bibitem{hong2020graph}
D.~Hong, J.~Kang, N.~Yokoya, and J.~Chanussot, ``Graph-induced aligned learning
  on subspaces for hyperspectral and multispectral data,'' {\em IEEE Trans.
  Geosci. Remote Sens.}, 2020.
\newblock DOI: 10.1109/TGRS.2020.3021140.

\bibitem{hershey2014deep}
J.~R. Hershey, J.~L. Roux, and F.~Weninger, ``Deep unfolding: Model-based
  inspiration of novel deep architectures,'' {\em arXiv preprint
  arXiv:1409.2574}, 2014.

\bibitem{diamond2017unrolled}
S.~Diamond, V.~Sitzmann, F.~Heide, and G.~Wetzstein, ``Unrolled optimization
  with deep priors,'' {\em arXiv preprint arXiv:1705.08041}, 2017.

\end{thebibliography}
\begin{IEEEbiography}[{\includegraphics[width=1in,height=1.25in,clip,keepaspectratio]{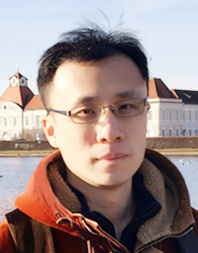}}]{Danfeng Hong}
(S'16--M'19) received the M.Sc. degree (summa cum laude) in computer vision, College of Information Engineering, Qingdao University, Qingdao, China, in 2015, the Dr. -Ing degree (summa cum laude) in Signal Processing in Earth Observation (SiPEO), Technical University of Munich (TUM), Munich, Germany, in 2019. 

Since 2015, he worked as a Research Associate at the Remote Sensing Technology Institute (IMF), German Aerospace Center (DLR), Oberpfaffenhofen, Germany. Currently, he is a research scientist and leads a Spectral Vision working group at IMF, DLR, and also an adjunct scientist in GIPSA-lab, Grenoble INP, CNRS, Univ. Grenoble Alpes, Grenoble, France. 

His research interests include signal / image processing and analysis, hyperspectral remote sensing, machine / deep learning, artificial intelligence and their applications in Earth Vision.
\end{IEEEbiography}

\vskip -2\baselineskip plus -1fil

\begin{IEEEbiography}[{\includegraphics[width=1in,height=1.25in,clip,keepaspectratio]{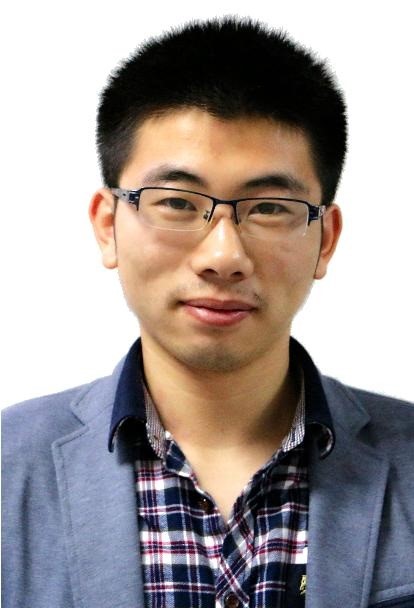}}]{Wei He}
(S'14-M'17)
 received the B.S. degree in School of Mathematics and statistics and the Ph.D degree in Surveying, Mapping and Remote Sensing (LIESMARS) from Wuhan University, Wuhan, China, in 2012 and 2017, respectively. He was a researcher at RIKEN Center for Advanced Intelligence Project, from 2018 to 2020. \par
 He is currently a research scientist with the Geoinformatics unit, RIKEN Center for Advanced Intelligence Project, Japan. His research interests include image quality improvement, remote sensing image processing, matrix/tensor analysis and deep learning.
\end{IEEEbiography}

\vskip -2\baselineskip plus -1fil

\begin{IEEEbiography}[{\includegraphics[width=1in,height=1.25in,clip,keepaspectratio]{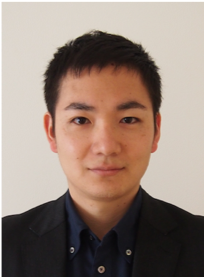}}]{Naoto Yokoya} (S'10--M'13) received the M.Eng. and Ph.D. degrees from the Department of Aeronautics and Astronautics, the University of Tokyo, Tokyo, Japan, in 2010 and 2013, respectively.

He is currently a Lecturer at the University of Tokyo and a Unit Leader at the RIKEN Center for Advanced Intelligence Project, Tokyo, Japan, where he leads the Geoinformatics Unit. He was an Assistant Professor at the University of Tokyo from 2013 to 2017. In 2015-2017, he was an Alexander von Humboldt Fellow, working at the German Aerospace Center (DLR), Oberpfaffenhofen, and Technical University of Munich (TUM), Munich, Germany. His research is focused on the development of image processing, data fusion, and machine learning algorithms for understanding remote sensing images, with applications to disaster management.

Dr. Yokoya won the first place in the 2017 IEEE Geoscience and Remote Sensing Society (GRSS) Data Fusion Contest organized by the Image Analysis and Data Fusion Technical Committee (IADF TC). He is the Chair (2019-2021) and was a Co-Chair (2017-2019) of IEEE GRSS IADF TC and also the secretary of the IEEE GRSS All Japan Joint Chapter since 2018. He is an Associate Editor for the IEEE Journal of Selected Topics in Applied Earth Observations and Remote Sensing (JSTARS) since 2018. He is/was a Guest Editor for the IEEE JSTARS in 2015-2016, for Remote Sensing in 2016-2020, and for the IEEE Geoscience and Remote Sensing Letters (GRSL) in 2018-2019.
\end{IEEEbiography}

\vskip -2\baselineskip plus -1fil

\begin{IEEEbiography}[{\includegraphics[width=1in,height=1.25in,clip,keepaspectratio]{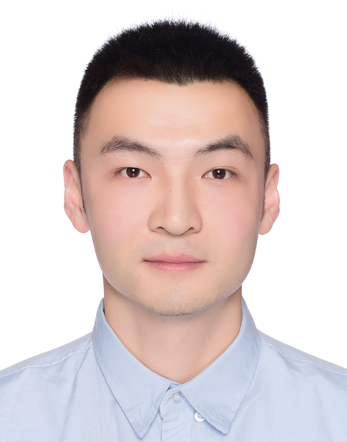}}]{Jing Yao} received the B.Sc. degree from Northwest University, Xi’an, China, in 2014, and the Ph.D. degree in the School of Mathematics and Statistics, Xi’an Jiaotong University, Xi’an, China, in 2021. 

He is currently an Assistant Professor with the Key Laboratory of Digital Earth Science, Aerospace Information Research Institute, Chinese Academy of Sciences, Beijing, China. From 2019 to 2020, he was a visiting student at Signal Processing in Earth Observation (SiPEO), Technical University of Munich (TUM), Munich, Germany, and at the Remote Sensing Technology Institute (IMF), German Aerospace Center (DLR), Oberpfaffenhofen, Germany.

His research interests include low-rank modeling, hyperspectral image analysis and deep learning-based image processing methods.
\end{IEEEbiography}

\vskip -2\baselineskip plus -1fil

\begin{IEEEbiography}[{\includegraphics[width=1in,height=1.25in,clip,keepaspectratio]{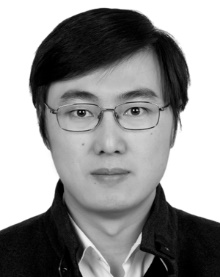}}]{Lianru Gao} (M'12--SM'18) received the B.S. degree in civil engineering from Tsinghua University, Beijing, China, in 2002, the Ph.D. degree in cartography and geographic information system from Institute of Remote Sensing Applications, Chinese Academy of Sciences (CAS), Beijing, China, in 2007.

He is currently a Professor with the Key Laboratory of Digital Earth Science, Aerospace Information Research Institute, CAS. He also has been a visiting scholar at the University of Extremadura, Cáceres, Spain, in 2014, and at the Mississippi State University (MSU), Starkville, USA, in 2016. His research focuses on hyperspectral image processing and information extraction. In last ten years, he was the PI of 10 scientific research projects at national and ministerial levels, including projects by the National Natural Science Foundation of China (2010-2012, 2016-2019, 2018-2020), and by the Key Research Program of the CAS (2013-2015). He has published more than 160 peer-reviewed papers, and there are more than 80 journal papers included by SCI. He was coauthor of an academic book entitled ``Hyperspectral Image Classification And Target Detection''. He obtained 28 National Invention Patents in China. He was awarded the Outstanding Science and Technology Achievement Prize of the CAS in 2016, and was supported by the China National Science Fund for Excellent Young Scholars in 2017, and won the Second Prize of The State Scientific and Technological Progress Award in 2018. He received the recognition of the Best Reviewers of the IEEE Journal of Selected Topics in Applied Earth Observations and Remote Sensing in 2015, and the Best Reviewers of the IEEE Transactions on Geoscience and Remote Sensing in 2017.
\end{IEEEbiography}

\vskip -2\baselineskip plus -1fil

\begin{IEEEbiography}[{\includegraphics[width=1in,height=1.25in,clip,keepaspectratio]{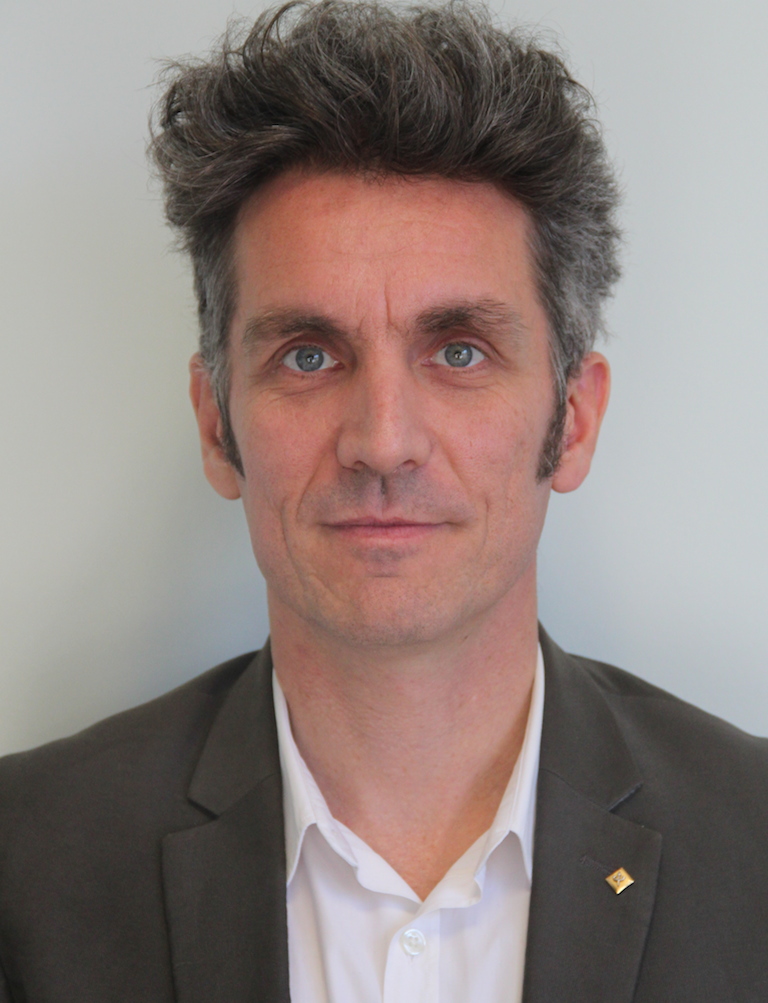}}]{Jocelyn Chanussot}
(M'04--SM'04--F'12) received the M.Sc. degree in electrical engineering from the Grenoble Institute of Technology (Grenoble INP), Grenoble, France, in 1995, and the Ph.D. degree from the Université de Savoie, Annecy, France, in 1998. Since 1999, he has been with Grenoble INP, where he is currently a Professor of signal and image processing. His research interests include image analysis, hyperspectral remote sensing, data fusion, machine learning and artificial intelligence. He has been a visiting scholar at Stanford University (USA), KTH (Sweden) and NUS (Singapore). Since 2013, he is an Adjunct Professor of the University of Iceland. In 2015-2017, he was a visiting professor at the University of California, Los Angeles (UCLA). He holds the AXA chair in remote sensing and is an Adjunct professor at the Chinese Academy of Sciences, Aerospace Information research Institute, Beijing.

Dr. Chanussot is the founding President of IEEE Geoscience and Remote Sensing French chapter (2007-2010) which received the 2010 IEEE GRS-S Chapter Excellence Award. He has received multiple outstanding paper awards. He was the Vice-President of the IEEE Geoscience and Remote Sensing Society, in charge of meetings and symposia (2017-2019). He was the General Chair of the first IEEE GRSS Workshop on Hyperspectral Image and Signal Processing, Evolution in Remote sensing (WHISPERS). He was the Chair (2009-2011) and  Cochair of the GRS Data Fusion Technical Committee (2005-2008). He was a member of the Machine Learning for Signal Processing Technical Committee of the IEEE Signal Processing Society (2006-2008) and the Program Chair of the IEEE International Workshop on Machine Learning for Signal Processing (2009). He is an Associate Editor for the IEEE Transactions on Geoscience and Remote Sensing, the IEEE Transactions on Image Processing and the Proceedings of the IEEE. He was the Editor-in-Chief of the IEEE Journal of Selected Topics in Applied Earth Observations and Remote Sensing (2011-2015). In 2014 he served as a Guest Editor for the IEEE Signal Processing Magazine. He is a Fellow of the IEEE, a member of the Institut Universitaire de France (2012-2017) and a Highly Cited Researcher (Clarivate Analytics/Thomson Reuters).

\end{IEEEbiography}

\vskip -2\baselineskip plus -1fil

\begin{IEEEbiography}[{\includegraphics[width=1in,height=1.25in,clip,keepaspectratio]{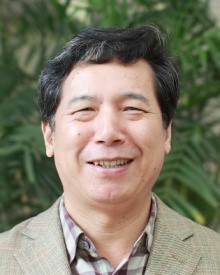}}]{Liangpei Zhang}
(M'06-SM'08-F'19) received the B.S. degree in physics from Hunan Normal University, Changsha, China, in 1982, the M.S. degree in optics from the Xi’an Institute of Optics and Precision Mechanics, Chinese Academy of Sciences, Xi’an, China, in 1988, and the Ph.D. degree in photogrammetry and remote sensing from Wuhan University, Wuhan, China, in 1998. \par
He is a ``Chang-Jiang Schola'' chair professor appointed by the ministry of education of China in state key laboratory of information engineering in surveying, mapping, and remote sensing (LIESMARS), Wuhan University. He was a principal scientist for the China state key basic research project (2011-2016) appointed by the ministry of national science and technology of China to lead the remote sensing program in China. He has published more than 700 research papers and five books. He is the Institute for Scientific Information (ISI) highly cited author. He is the holder of 30 patents. His research interests include hyperspectral remote sensing, high-resolution remote sensing, image processing, and artificial intelligence. \par
Dr. Zhang is a Fellow of Institute of Electrical and Electronic Engineers (IEEE) and the Institution of Engineering and Technology (IET). He was a recipient of the 2010 best paper Boeing award, the 2013 best paper ERDAS award from the American society of photogrammetry and remote sensing (ASPRS) and 2016 best paper theoretical innovation award from the international society for optics and photonics (SPIE). His research teams won the top three prizes of the IEEE GRSS 2014 Data Fusion Contest, and his students have been selected as the winners or finalists of the IEEE International Geoscience and Remote Sensing Symposium (IGARSS) student paper contest in recent years. He also serves as an associate editor or editor of more than ten international journals. Dr. Zhang is currently serving as an associate editor of the IEEE TRANSACTIONS ON GEOSCIENCE AND REMOTE SENSING. Dr. Zhang is the founding chair of IEEE Geoscience and Remote Sensing Society (GRSS) Wuhan Chapter. 
\end{IEEEbiography}

\vskip -2\baselineskip plus -1fil

\begin{IEEEbiography}[{\includegraphics[width=1in,height=1.25in,clip,keepaspectratio]{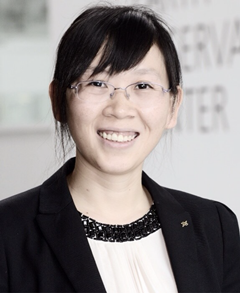}}]{Xiao Xiang Zhu}(M'12--SM'14--F'20) received the Master (M.Sc.) degree, her doctor of engineering (Dr.-Ing.) degree and her “Habilitation” in the field of signal processing from Technical University of Munich (TUM), Munich, Germany, in 2008, 2011 and 2013, respectively.
\par
She is currently the Professor for Data Science in Earth Observation (former: Signal Processing in Earth Observation) at Technical University of Munich (TUM) and the Head of the Department ``EO Data Science'' at the Remote Sensing Technology Institute, German Aerospace Center (DLR). Since 2019, Zhu is a co-coordinator of the Munich Data Science Research School (www.mu-ds.de). Since 2019 She also heads the Helmholtz Artificial Intelligence -- Research Field ``Aeronautics, Space and Transport". Since May 2020, she is the director of the international future AI lab "AI4EO -- Artificial Intelligence for Earth Observation: Reasoning, Uncertainties, Ethics and Beyond", Munich, Germany. Since October 2020, she also serves as a co-director of the Munich Data Science Institute (MDSI), TUM. Prof. Zhu was a guest scientist or visiting professor at the Italian National Research Council (CNR-IREA), Naples, Italy, Fudan University, Shanghai, China, the University  of Tokyo, Tokyo, Japan and University of California, Los Angeles, United States in 2009, 2014, 2015 and 2016, respectively. Her main research interests are remote sensing and Earth observation, signal processing, machine learning and data science, with a special application focus on global urban mapping.

Dr. Zhu is a member of young academy (Junge Akademie/Junges Kolleg) at the Berlin-Brandenburg Academy of Sciences and Humanities and the German National  Academy of Sciences Leopoldina and the Bavarian Academy of Sciences and Humanities. She is a Fellow of IEEE, serves as an associate Editor of IEEE Transactions on Geoscience and Remote Sensing and as the area editor responsible for special issues of IEEE Signal Processing Magazine.
\end{IEEEbiography}

\end{document}